\documentclass{article}

\PassOptionsToPackage{numbers, compress}{natbib}
\usepackage[preprint]{neurips_2020}

\usepackage[utf8]{inputenc} % allow utf-8 input
\usepackage[T1]{fontenc}    % use 8-bit T1 fonts
\usepackage{hyperref}       % hyperlinks
\usepackage{url}            % simple URL typesetting
\usepackage{booktabs}       % professional-quality tables
\usepackage{nicefrac}       % compact symbols for 1/2, etc.
\usepackage{microtype}      % microtypography
\usepackage{subfig}         % subplots
\usepackage[usenames,dvipsnames,table]{xcolor}

\hypersetup{%
  colorlinks=true,
  linkcolor=NavyBlue,
  citecolor=ForestGreen,
  filecolor=magenta,
  urlcolor=blue,
}

\usepackage{amsmath,amssymb,amsfonts,mathrsfs,mathtools}

\usepackage[amsmath,thmmarks]{ntheorem}

\usepackage{lipsum}

\usepackage[inline,shortlabels]{enumitem}

\usepackage[nameinlink]{cleveref}

\usepackage{array}

\usepackage{makecell}
\usepackage{colortbl}
\colorlet{col1}{blue!10}
\colorlet{col2}{blue!5}
\newcommand\hfilll{\hspace{0pt plus 1filll}}

\usepackage{tabularx}

\usepackage[labelfont=bf]{caption}
\usepackage{siunitx}

\usepackage{multirow}

\newtheorem{theorem}{Theorem}

\newtheorem{lemma}[theorem]{Lemma}
\newtheorem{proposition}[theorem]{Proposition}

\theoremstyle{nonumberplain}
\theorembodyfont{\normalfont}
\theoremsymbol{\ensuremath{\square}}
\newtheorem{proof}{Proof}

\newcommand{\R}{\mathbb{R}}

\DeclarePairedDelimiter\abs{\lvert}{\rvert}
\DeclarePairedDelimiter\norm{\lVert}{\rVert}

\renewcommand{\epsilon}{\ensuremath\varepsilon}

\renewcommand{\phi}{\ensuremath{\varphi}}

\let\vec\mathbf

\newcommand*{\deq}{\vcentcolon=}

\DeclareMathOperator{\Tr}{\mathop{\mathrm{Tr}}}

\DeclareMathOperator{\spn}{\mathop{\mathrm{span}}}

\DeclareMathOperator{\diag}{\mathop{\mathrm{diag}}}

\DeclareMathOperator{\eye}{\mathop{\mathrm{Id}}}

\DeclareMathOperator{\Arg}{\mathop{\mathrm{Arg}}}

\DeclareMathOperator{\Riem}{\mathrm{R}}

\DeclareMathOperator{\Ric}{\mathrm{Ric}}

\DeclareMathOperator{\Diff}{\textrm{D}}

\DeclareMathOperator{\Ker}{\mathop{ker}}

\DeclareMathOperator{\KL}{\mathop{\mathrm{D_{KL}}}}

\newcommand{\leqs}{\leqslant}

\newcommand{\geqs}{\geqslant}

\newcommand{\Sym}{\mathcal{S}}

\newcommand{\Skew}{\mathcal{S}_{skew}}

\newcommand{\Spd}{\mathcal{S}^{\texttt{++}}}

\newcommand{\Sspd}{\mathcal{S}_{\ast}^{\texttt{++}}}

\newcommand{\St}{V}

\newcommand{\Gr}{Gr}

\newcommand{\RP}{\mathbf{RP}}

\newcommand{\Ortho}{O}

\newcommand{\So}{SO}

\newcommand{\Hyp}{\mathbb{H}}

\newcommand{\Sph}{\mathbb{S}}

\newcommand{\tagarray}{%
\mbox{}\refstepcounter{equation}%
$(\theequation)$%
}

\makeatletter
\newcommand{\printfnsymbol}[1]{%
  \textsuperscript{\@fnsymbol{#1}}%
}
\makeatother

\title{Computationally Tractable Riemannian Manifolds for Graph Embeddings}
\author{%
  Calin Cruceru\printfnsymbol{2}
  \quad
  Gary B{\'e}cigneul\printfnsymbol{2}\printfnsymbol{3}
  \quad
  Octavian-Eugen Ganea\printfnsymbol{2}\printfnsymbol{3} \\
  \printfnsymbol{2}Department of Computer Science, ETH Z\"urich, Switzerland \\
  \printfnsymbol{3}Computer Science and Artificial Intelligence Lab, MIT, USA \\
  \texttt{ccruceru@inf.ethz.ch}, \enskip \texttt{\{garyb,oct\}@mit.edu}
}

\begin{document}

\maketitle

\begin{abstract}
  Representing graphs as sets of node embeddings in certain curved Riemannian manifolds has recently
gained momentum in machine learning due to their desirable geometric inductive biases, e.g.,
hierarchical structures benefit from hyperbolic geometry. However, going beyond embedding spaces of
constant sectional curvature, while potentially more representationally powerful, proves to be
challenging as one can easily lose the appeal of computationally tractable tools such as geodesic
distances or Riemannian gradients. Here, we explore computationally efficient matrix manifolds,
showcasing how to learn and optimize graph embeddings in these Riemannian spaces. Empirically, we
demonstrate consistent improvements over Euclidean geometry while often outperforming hyperbolic and
elliptical embeddings based on various metrics that capture different graph properties. Our results
serve as new evidence for the benefits of non-Euclidean embeddings in machine learning pipelines.

\end{abstract}

\section{Introduction}
Before representation learning started gravitating around deep
representations~\cite{bengio2009learning} in the last decade, a line of research that sparked
interest in the early 2000s was based on the so called manifold
hypothesis~\cite{bengio2013representation}. According to it, real-world data given in their raw
format (e.g., pixels of images) lie on a low-dimensional manifold embedded in the input space. At
that time, most manifold learning algorithms were based on locally linear approximations to points
on the sought manifold (e.g., LLE~\cite{roweis2000nonlinear}, Isomap~\cite{tenenbaum2000global}) or
on spectral methods (e.g., MDS~\cite{hofmann1995multidimensional}, graph Laplacian
eigenmaps~\cite{belkin2002laplacian}).

Back to recent years, two trends are apparent:
\begin{enumerate*}[(i)]
  \item the use of graph-structured data and their direct processing by machine learning
    algorithms~\cite{bruna2014spectral,henaff2015deep,defferrard2016convolutional,grover2016node2vec},
    and
  \item the resurgence of the manifold hypothesis, but with a different flavor -- being explicit
    about the assumed manifold and, perhaps, the inductive bias that it entails: hyperbolic
    spaces~\cite{nickel2017poincare,nickel2018learning,ganea2018hyperbolic}, spherical
    spaces~\cite{wilson2014spherical}, and Cartesian products of
    them~\cite{gu2018learning,tifrea2018poincare,Skopek2020Mixed-curvature}.
\end{enumerate*}
While for the first two the choice can be a priori justified -- e.g., complex networks are
intimately related to hyperbolic geometry~\cite{krioukov2010hyperbolic} -- the last one, originally
through the work of Gu et al.~\cite{gu2018learning}, is motivated through the presumed flexibility
coming from its varying curvature. Our work takes that hypothesis further by exploring the
representation properties of several \emph{irreducible spaces}\footnote{Not expressible as Cartesian
products of other manifolds, be they model spaces, as in~\cite{gu2018learning}, or yet others.} of
non-constant sectional curvature. We use, in particular, Riemannian manifolds where points are
represented as specific types of matrices and which are at the sweet spot between semantic richness
and tractability.

\begin{figure}
  \centering

  \subfloat[``web-edu''\label{fig:web-edu}]{{\includegraphics[scale=0.12]{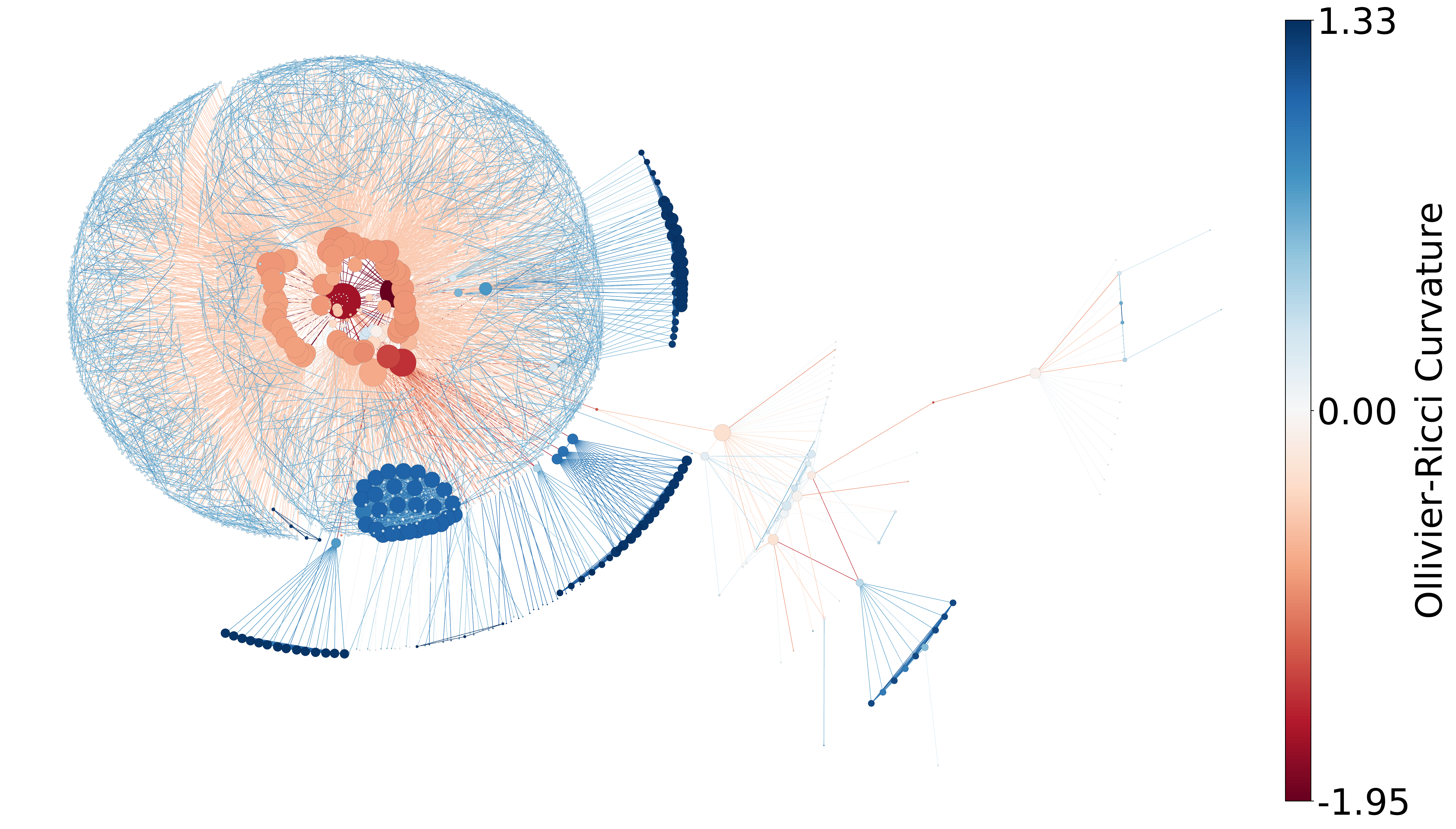} }}
  \subfloat[``road-minnesota''\label{fig:road-minnesota}]{{\includegraphics[scale=0.12]{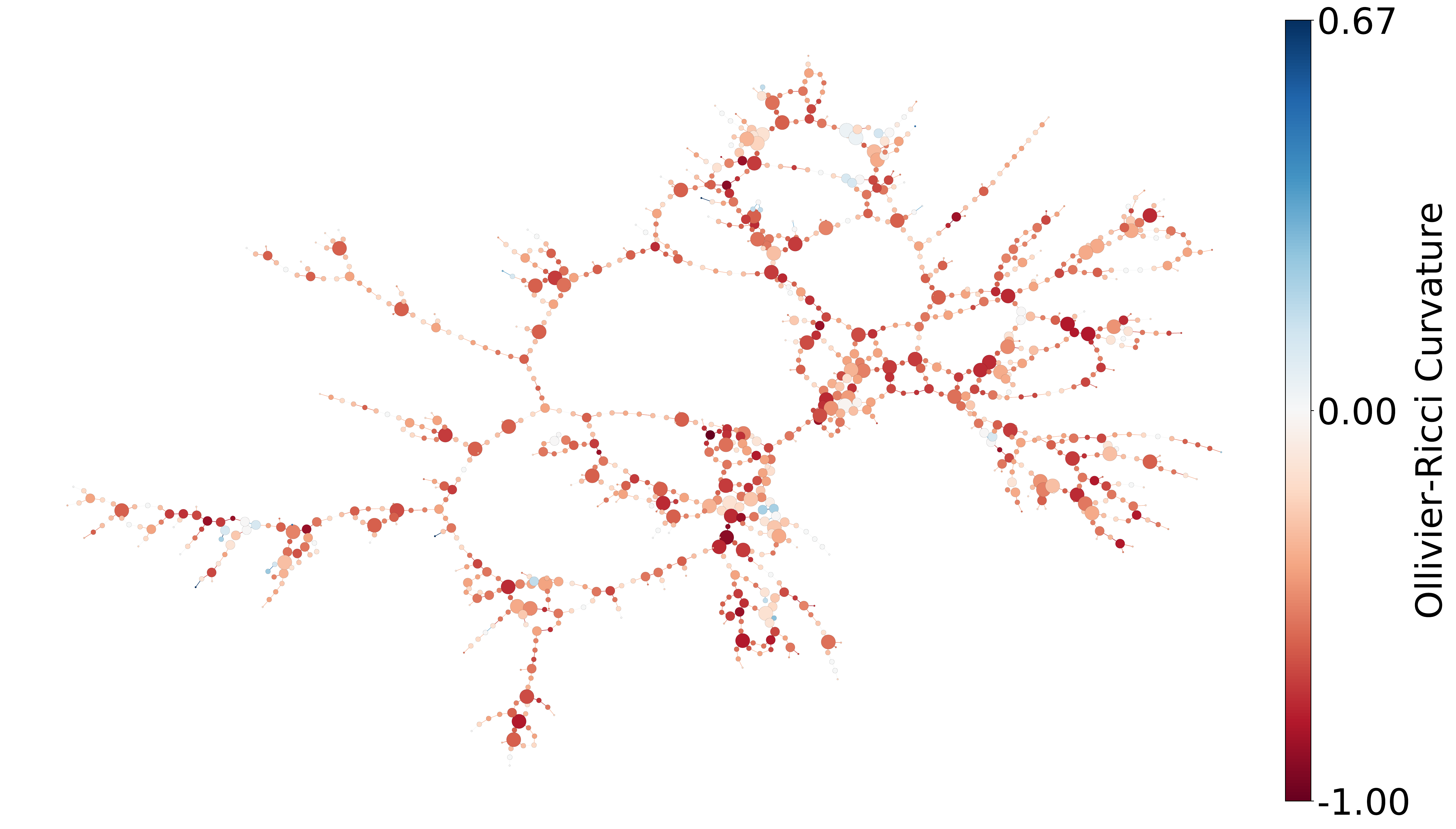} }}

  \caption{Two graphs used in our experiments: a web network from the .edu
  domain~\cite{gleich2004fast} and a road network in Minnesota~\cite{nr}. The plots include the
  Ollivier-Ricci curvatures of edges and nodes. We refer the reader to~\Cref{sec:geom-prop-graphs}
  for details. More such visualizations are included in~\Cref{sec:input-graphs}.}
  \label{fig:example-graphs}
\end{figure}

With no additional qualifiers, graph embedding is a vaguely specified intermediary step used as part
of systems solving a wide range of graph analytics
problems~\cite{nie2017unsupervised,wang2017community,wei2017cross,zhou2017scalable}.  What they all
have in common is the representation of certain parts of a graph as points in a continuous space.
% The desired mathematical properties depend on the problem setting. Classically,
% the Euclidean space has been ubiquitous due to its interpretability and structure: inner product,
% metric, and, very conveniently for compositional models, linearity.
% As part of the emerging field of geometric deep learning~\cite{bronstein2017geometric}, though,
% neural networks have been generalized to non-linear manifolds such as hyperbolic
% space~\cite{ganea2018hypnn}, elliptical spaces~\cite{davidson2018hyperspherical,xu2018spherical},
% and even several matrix manifolds~\cite{dong2017deep,huang2017riemannian,zhang2018grassmannian}.
As a particular instance of that general task, here we embed nodes of graphs with structural
information only (i.e., undirected and without node or edge labels), as the ones shown
in~\Cref{fig:example-graphs}, in novel curved spaces, by leveraging the closed-form expressions of
the corresponding Riemannian distance between embedding points; the resulting geodesic distances
enter a differentiable objective function which ``compares'' them to the ground-truth metric given
through the node-to-node graph distances. We focus on the representation capabilities of the
considered matrix manifolds relative to the previously studied spaces by monitoring graph
reconstruction metrics. We note that preserving graph structure is essential to downstream tasks
such as link prediction~\cite{trouillon2016complex,zhang2018link} and node
classification~\cite{li2016discriminative,wang2017community,tu2016max}.

Our \textbf{main contributions} include
\begin{enumerate*}[label=\textbf{(\roman*)}]
  \item the introduction of two families of matrix manifolds for graph embedding purposes: the
    non-positively curved spaces of symmetric positive definite (SPD) matrices, and the compact,
    non-negatively curved Grassmann manifolds;
  \item reviving Stochastic Neighbor Embedding (SNE)~\cite{hinton2003stochastic} in the context of
    Riemannian optimization as a way to unify, on the one hand, the loss functions based on the
    reconstruction likelihood of local graph neighborhoods and, on the other hand, the global,
    all-pairs stress functions used for global metric recovery;
  \item a generalization of the usual ranking-based metric to quantify reconstruction fidelity
    beyond immediate neighbors;
  \item a comprehensive experimental comparison of the introduced manifolds against the baselines in
    terms of their graph reconstruction capabilities, focusing on the impact of curvature.
    % Our results show that many graphs benefit from their novel geometric properties.
\end{enumerate*}

\paragraph{Related Work.}
% While we mentioned earlier the prior work that we build upon, let us also point out several
% tangential research directions.
Our work is inspired by the emerging field of geometric deep learning
(GDL)~\cite{bronstein2017geometric} through its use of geometry. That being said, our motivation and
approach are different. In GDL, deep networks transform data in a geometry-aware way, usually as
part of larger discriminative or generative models: e.g., graph neural
networks~\cite{bruna2014spectral,henaff2015deep}, hyperbolic neural
networks~\cite{ganea2018hypnn,mathieu2019continuous,gulcehre2018hyperbolic}, hyperspherical neural
networks~\cite{defferrard2016convolutional,xu2018spherical,davidson2018hyperspherical}, and
others~\cite{bachmann2019constant,Skopek2020Mixed-curvature}. We, on the other hand, embed graph nodes in a
simpler, transductive setting, employing Riemannian
optimization~\cite{bonnabel2013stochastic,becigneul2018riemannian} to directly obtain the
corresponding embeddings. The broader aim to which we contribute is that of understanding the role
played by the space curvature in graph representation learning. In this sense, works such as those
of Sarkar~\cite{sarkar2011low} and Krioukov et al.~\cite{krioukov2010hyperbolic}, who formally
describe the connections between certain types of graphs (i.e., trees and complex networks,
respectively) and hyperbolic geometry, inspire us: ultimately, we seek similar results for new
classes of graphs and embedding spaces. This work, mostly an empirical one, is a first step in that
direction for two families of matrix manifolds. It is similar in spirit to Gu et
al.~\cite{gu2018learning} who empirically show that Cartesian products of model spaces can provide a
good inductive bias in some cases. Finally, the manifolds themselves are not new in the machine
learning literature: recent computer vision applications take into account the intrinsic geometry of
SPD matrices~\cite{dong2017deep,huang2017riemannian} and Grassmann
subspaces~\cite{huang2018building,zhang2018grassmannian} when building discriminative models. It is
the study of the implications of their curvature for graph representations that is novel.
% Finally, our use of the SNE objective, to the best of our knowledge the first in the context of
% Riemannian optimization, differs from works such as GraphTSNE~\cite{leow2019graphtsne}, where it
% is used to visualize graphs in Euclidean space.

\section{Preliminaries \& Background}\label{sec:background}
% We introduce in this section basic facts needed to develop our embedding learning framework.

\paragraph{Notation.}
Let $G = (X, E, w)$ be an undirected graph, with $X$ the set of nodes, $E$ the set of edges, and $w
: E \to \R_+$ the edge-weighting function. Let $m = \abs{X}$. We denote by $d_G(x_i,x_j)$ the
shortest path distance between nodes $x_i,x_j \in X$, induced by $w$. The node embeddings
are\footnote{We use $i \in [m]$ as a short-hand for $i \in \{1,2,\ldots,m\}$.} $Y = \{y_i\}_{i \in
[m]} \subset \mathcal{M}$ and the geodesic distance function is $d_\mathcal{M}(y_i,y_j)$, with
$\mathcal{M}$ -- the embedding space -- a Riemannian manifold.  $\mathcal{N}(x_i)$ denotes the set
of neighbors of node $x_i$.

\vspace*{-0.2cm}
\paragraph{Riemannian Geometry.}
A comprehensive account of the fundamental concepts from Riemannian geometry is included
in~\Cref{sec:diff-geom}. Informally, an $n$-dimensional manifold $\mathcal{M}$ is a space that
locally resembles $\R^n$. Each point $x \in \mathcal{M}$ has attached a tangent space $T_x
\mathcal{M}$ -- a vector space that can be thought of as a first-order local approximation of
$\mathcal{M}$ around $x$. The Riemannian metric $\langle \cdot, \cdot \rangle_x$ is a collection of
inner products on these tangent spaces that vary smoothly with $x$. It makes possible measuring
geodesic distances, angles, and curvatures. The different notions of \emph{curvature} quantify the
ways in which a surface is locally curved around a point. The exponential map is a function $\exp_x
: T_x \mathcal{M} \to \mathcal{M}$ that can be seen as ``folding'' or projecting the tangent space
onto the manifold. Its inverse is called the logarithm map, $\log_x(\cdot)$.

\vspace*{-0.2cm}
\paragraph{Learning Framework.}
The embeddings are learned in the framework used in recent prior
work~\cite{nickel2017poincare,gu2018learning} in which a loss function $\mathcal{L}$ depending on
the embedding points solely via the Riemannian distances between them is minimized using stochastic
Riemannian optimization~\cite{bonnabel2013stochastic,becigneul2018riemannian}. In this respect, the
following general property is useful~\cite{lee2006riemannian}: for any point $x$ on a Riemannian
manifold $\mathcal{M}$ and any $y$ in a neighborhood of $x$, we have $\nabla_x^R\ d^2(x, y)  =
-2\log_x(y)$.\footnote{$\nabla_x^R$ denotes the Riemannian gradient at $x$.
See~\Cref{sec:diff-geom}.} Hence, as long as $\mathcal{L}$ is differentiable with respect to the
(squared) distances, it will also be differentiable with respect to the embedding points. The
specifics of $\mathcal{L}$ are deferred to~\Cref{sec:decoupling}.

\vspace*{-0.2cm}
\paragraph{Model Spaces \& Cartesian Products.}
The model spaces of Riemannian geometry are manifolds with constant sectional curvature $K$:
\begin{enumerate*}[(i)]
  \item Euclidean space ($K = 0$),
  \item hyperbolic space ($K < 0$), and
  \item elliptical space ($K > 0$).
\end{enumerate*}
We summarize the Riemannian geometric tools of the last two in~\Cref{sec:hyp-sph}. They are used as
baselines in our experiments. We also recall that given a set of manifolds
$\{\mathcal{M}_i\}_{i=1}^k$, the product manifold $\mathcal{M} = \bigtimes_{i=1}^k \mathcal{M}_i$
has non-constant sectional curvature and can be used for graph embedding purposes as long as each
factor has efficient closed-form formulas for the quantities of interest~\cite{gu2018learning}.

\vspace*{-0.2cm}
\paragraph{Measuring Curvature around Embeddings.}
Curvature properties are central to our work since they set apart the matrix manifolds discussed
in~\Cref{sec:matrix-manifolds}. Recall that any manifold locally resembles Euclidean space. Hence,
several ways of quantifying \emph{the actual} space curvature between embeddings have been proposed;
see~\Cref{sec:geom-prop-graphs} for an overview. One which we find more convenient for analysis and
presentation purposes, because it yields bounded and easily interpretable values, is based on sums
of angles in geodesic triangles formed by triples $x,y,z \in \mathcal{M}$,
\begin{equation}
  k_{\theta}(x, y, z) = \theta_{x,y} + \theta_{x,z} + \theta_{y,z},\enskip\text{with}\
  \theta_{x_1,x_2} = \cos^{-1} \frac{\langle u_1, u_2 \rangle_{x_3}}{\norm*{u_1}_{x_3} \norm*{u_2}_{x_3}},\
  u_{\{1,2\}} = \log_{x_3}(x_{\{1,2\}}).
\end{equation}
It takes values in the intervals $[0, \pi]$ and $[\pi, 3 \pi]$, in hyperbolic and elliptical spaces,
respectively. In practice, we look at empirical distributions of $\overline{k}_\theta = (k_\theta -
\pi) / 2 \pi$, with values in $[-0.5, 0]$ and $[0, 1]$, respectively, obtained by sampling triples
$(x,y,z)$ from an embedding set $\{y_i\}_{i=1}^k$.

\section{Matrix Manifolds for Graph Embeddings}\label{sec:matrix-manifolds}
We propose two families of matrix manifolds that lend themselves to computationally tractable
Riemannian optimization in our graph embedding framework.\footnote{Counterexamples: low-rank
manifold, (compact) Stiefel manifold; they lack closed-form distance functions.} They cover negative
and positive curvature ranges, respectively, resembling the relationship between hyperbolic and
hyperspherical spaces. Their properties are summarized in~\Cref{tab:diff-geom-summary}. Details and
proofs are included in~\Cref{sec:manifolds-details}.

\begin{table}
  \caption{
    Summary of Riemannian geometric tools for the
    SPD~\cite{bhatia2009positive,bridson2013metric,jeuris2015riemannian} and
    Grassmann~\cite{edelman1998geometry,zhang2018grassmannian} manifolds.
    Notation:
      $A, B$ -- manifold points;
      $P, Q$ -- tangent space points;
      $P'$ -- ambient space point;
      $\exp(A)$\, /\, $\log(A)$ -- matrix exponential / logarithm.
  }
  \label{tab:diff-geom-summary}

  \small
  \centering
  %% We use a double break because, for some reason, \makecell leaves too little vertical space. But
  %% then there's too much vertical space, so we make the array stretch from booktabs less than 1.0
  \renewcommand{\arraystretch}{0.4}

  \begin{tabular}{@{}lll@{}}
    \toprule
    \thead{\bfseries Property} & \thead{\textbf{SPD} $\Spd(n)$}  & \thead{\textbf{Grassmann} $\Gr(k,n)$} \\
    \midrule

    \\
    Dimension, $\dim(\mathcal{M})$ &
    $n (n + 1) / 2$ &
    $k (n - k)$
    \\

    \\
    Tangent space, $T_A\, \mathcal{M}$ &
    $\{A \in \R^{n \times n} : A = A^\top\}$ &
    $\{ P \in \R^{n \times k} : A^\top P = 0 \}$
    \\

    \\
    Projection, $\pi_A(P')$ &
    $(P' + P'^\top)/2$ &
    $(\eye_n - A A^\top) P'$
    \\

    \\
    Riem.\ metric, $\langle P, Q \rangle_A$ &
    $\Tr A^{-1} P A^{-1} Q$ \hfilll\tagarray\label{eq:spd-metric} &
    $\Tr P^\top Q$ \hfilll\tagarray\label{eq:grass-metric}
    \\

    \\
    Riem.\ gradient, $\nabla_A^R$ &
    $A\, \pi_A(\nabla_A^E)\, A$ &
    $\pi_A(\nabla_A^E)$
    \\

    \\
    Geodesic, $\gamma_{A;P}(t)$ &
    $A \exp(t A^{-1} P)$ &
    $\begin{bmatrix} AV & U\end{bmatrix} \begin{bmatrix} \cos(t \Sigma) & \sin(t \Sigma)\end{bmatrix} V^\top$
      \, with $P = U \Sigma V^\top$
    \\

    \\
    Retraction, $R_A(P)$ &
    $A + P + \frac{1}{2} P A^{-1} P$ &
    $U V^\top$\, with $A + P = U \Sigma V$
    \\

    \\
    Log map, $\log_A(B)$ &
    $A \log(A^{-1} B)$ &
    $U \Sigma V^\top$\, with $\begin{bmatrix}
        A^\top B \\
        \big(\eye_n - A A^\top\big) B
      \end{bmatrix} = \begin{bmatrix}
        V \cos(\Sigma) V^\top \\
        U \sin(\Sigma) V^\top
      \end{bmatrix}$
    \\

    \\
    Riem.\ dist., $d(A,B)$ &
    $\norm*{\log(A^{-1} B)}_F$ \hfilll\tagarray\label{eq:spd-dist} &
    $\sqrt{\sum_{i=1}^k \theta_i^2}$
      \, with $A^\top B = U \diag\big(\cos(\theta_i)\big) V^\top$ \hfilll\tagarray\label{eq:grass-dist}
    \\

    \bottomrule
  \end{tabular}
\end{table}

\subsection{Non-positive Curvature: SPD Manifold}\label{sec:nonpos}
The space of $n \times n$ real symmetric positive-definite matrices,
$\Spd(n) \deq \{A \in \Sym(n) : \langle x, A x \rangle > 0\ \textrm{for all}\ x \ne 0\}$,
is an $\frac{n (n+1)}{2}$-dimensional differentiable manifold -- an embedded submanifold of
$\Sym(n)$, the space of $n \times n$ symmetric matrices. Its tangent space can be identified with
$\Sym(n)$.

\vspace*{-0.2cm}
\paragraph{Riemannian Structure.}
The most common Riemannian metric endowed to $\Spd(n)$ is $\langle P, Q \rangle_A = \Tr A^{-1} P
A^{-1} Q$. Also called the \emph{canonical} metric, it is motivated as being invariant to congruence
transformations $\Gamma_X(A) = X^\top A X$, with $X$ an $n \times n$ invertible
matrix~\cite{pennec2006riemannian}. The induced distance function is\footnote{We use $\lambda_i(X)$
to denote the $i$th eigenvalue of $X$ when the order is not important.} $d(A, B) =
\sqrt{\sum_{i=1}^n \log^2\big(\lambda_i(A^{-1} B)\big)}$. It can be interpreted as measuring how
well $A$ and $B$ can be simultaneously reduced to the identity matrix~\cite{chossat2009hyperbolic}.

% An interpretation of the eigenvalues $\lambda_i(A^{-1} B)$ can be obtained by recalling that for any
% $A, B \in \Spd(n)$, there exist an invertible matrix $X$ and a diagonal matrix $D$ such that $X^\top
% A X = \eye_n$ and $X^\top B X = D$. Thus, the distance can be seen as measuring how well $A$ and $B$
% can be simultaneously reduced to the identity matrix~\cite{chossat2009hyperbolic}.
% See~\Cref{sec:manifolds-details} for proofs and details.

\vspace*{-0.2cm}
\paragraph{Properties.}
The canonical SPD manifold has non-positive sectional curvature
everywhere~\cite{bhatia2009positive}. It is also a high-rank symmetric
space~\cite{lang2012fundamentals}. The high-rank property tells us that there are \emph{at least
planes} of the tangent space on which the sectional curvature vanishes. Contrast this with the
hyperbolic space which is also a symmetric space but where the only (intrinsic) flats are the
geodesics. Moreover, only one degree of freedom can be factored out of the manifold $\Spd(n)$: it is
isometric to $\Sspd(n) \times \R$, with $\Sspd(n) \deq \{A \in \Spd(n) : \det(A) = 1\}$, an
irreducible manifold ~\cite{dolcetti2018differential}. Hence, $\Spd$ achieves a mix of flat and
negatively-curved areas that cannot be obtained via other Riemannian Cartesian products.

\vspace*{-0.2cm}
\paragraph{Alternative Metrics.}
There are several other metrics that one can endow the SPD manifold with, yielding different
geometric properties: the \emph{log-Euclidean} metric~\citep[see, e.g.,][]{arsigny2006log} induces a
flat manifold, while the \emph{Bures-Wasserstein} metric from quantum information
theory~\cite{bhatia2019bures} leads to a non-negatively curved manifold. The latter has been
leveraged in~\cite{muzellec2018generalizing} to embed graph nodes as elliptical distributions.
Finally, a popular alternative to the (squared) canonical distance is the \emph{symmetric Stein
divergence},
$S(A, B) \deq \log \det\big(\frac{A + B}{2}\big) - \frac{1}{2} \log \det(A B)$.
It has been thoroughly studied in~\cite{sra2012new,sra2015conic} who prove that $\sqrt{S}$ is indeed
a metric and that $S(A, B)$ shares many properties of the Riemannian distance
function~\eqref{eq:spd-dist} such as congruence and inversion invariances as well as geodesic
convexity in each argument. It is particularly appealing for backpropagation-based training due to
its computationally efficient gradients (see below). Hence, we experiment with it too when matching
graph metrics.

\vspace*{-0.2cm}
\paragraph{Computational Aspects.}
We compute gradients via automatic differentiation~\cite{paszke2017automatic}. Nonetheless, notice
that if $A = U D U^\top$ is the eigendecomposition of a symmetric matrix with distinct eigenvalues
and $\mathcal{L}$ is some loss function that depends on $A$ only via $D$, then $\frac{\partial
\mathcal{L}}{\partial A} = U \frac{\partial \mathcal{L}}{\partial D}
U^\top$~\cite{giles2008extended}.  Computing geodesic distances requires the eigenvalues of $A^{-1}
B$, though, which may not be symmetric. We overcome that by using the matrix $A^{-1/2} B A^{-1/2}$
instead which is SPD and has the same spectrum. Moreover, for the $2 \times 2$ and $3 \times 3$
cases, we use closed-form eigenvalue formulas to speed up our implementation.\footnote{
This could be done in theory for $n \leqs 4$ -- a consequence of the Abel-Ruffini theorem from
algebra. However, for $n=4$ the formulas are outperformed by numerical eigenvalue algorithms.
}
For the Stein divergence, the gradients can be computed in closed form as $\nabla_A S(A, B) =
\frac{1}{2} (A + B)^{-1} - \frac{1}{2} A^{-1}$. We additionally note that many of the required
matrix operations can be efficiently computed via Cholesky decompositions.
See~\Cref{sec:manifolds-details} for details.

\subsection{Non-negative Curvature: Grassmann Manifold}\label{sec:nonneg}
The orthogonal group $\Ortho(n)$ is the set of $n \times n$ real orthogonal matrices. It is a
special case of the compact Stiefel manifold $\St(k,n) \deq \{ A \in \R^{n \times k} : A^\top A =
\eye_k \}$, i.e., the set of $n \times k$ ``tall-skinny'' matrices with orthonormal columns, for $k
\leqs n$. The Grassmannian is defined as the space of $k$-dimensional linear subspaces of $\R^n$. It
is related to the Stiefel manifold in that every orthonormal $k$-frame in $\R^n$ spans a
$k$-dimensional subspace of the $n$-dimensional Euclidean space. Similarly, every such subspace
admits infinitely many orthonormal bases. This suggests the identification of the Grassmann manifold
$\Gr(k,n)$ with the quotient space $\St(k,n) / \Ortho(k)$ (more about quotient manifolds
in~\Cref{sec:diff-geom}).  In other words, an $n \times k$ orthonormal matrix $A \in \St(k,n)$
represents the equivalence class $[A] = \{ A Q_k : Q_k \in \Ortho(k) \} \cong \spn(A)$, which is a
single point on $\Gr(k,n)$.

\vspace*{-0.2cm}
\paragraph{Riemannian Structure.}
The canonical Riemannian metric of $\Gr(k, n)$ is simply the Frobenius inner
product~\eqref{eq:grass-metric}. We refer to~\cite{edelman1998geometry} for details on how it arises
from its quotient geometry. The closed form formula for the Riemannian distance, shown
in~\eqref{eq:grass-dist}, depends on the set $\{\theta_i\}_{i=1}^k$ of so-called principal angles
between two subspaces. They can be interpreted as the minimal angles between all possible bases of
the two subspaces~\cite{zhang2018grassmannian}.

\vspace*{-0.2cm}
\paragraph{Properties.}
The Grassmann manifold $\Gr(k,n)$ is a compact, non-negatively curved manifold. As shown
in~\cite{wong1968sectional}, its sectional curvatures at $A \in \Gr(k,n)$ satisfy $K_A(P, Q) = 1$
(for $k=1,\ n>2$) and $0 \leqs K_A(P, Q) \leqs 2$ (for $k > 1,\ n > k$), for all $P, Q \in T_A\,
\Gr(k, n)$. Contrast this with the constant positive curvature of the sphere which can be made
arbitrarily large by making $R \to 0$.

% \paragraph{Alternative Metrics.}
% Several alternatives to the arc-length metric~\eqref{eq:grass-dist} have been proposed, all
% expressible in terms of the principle angles -- see~\citep[][Section 4.3]{edelman1998geometry} for
% an overview. A popular one is the so-called projection norm, $d_p(A,B) = \norm*{A A^\top - B
% B^\top}_F$. It corresponds to embedding $\Gr(k,n)$ in $\R^{n \times n}$ and then using the ambient
% space metric -- analogous to taking Euclidean distances between points on a sphere, which ignores
% its geometry.

\vspace*{-0.2cm}
\paragraph{Computational Aspects.}
Computing a geodesic distance requires the SVD decomposition of an $k \times k$, matrix which can be
significantly smaller than the manifold dimension $k (n - k)$. For $k=2$, we use closed-form
solutions for singular values. See~\Cref{sec:manifolds-details} for details. Otherwise, we employ
standard numerical algorithms. For the gradients, a result analogous to the one for eigenvalues from
earlier (\Cref{sec:nonpos},~\cite{giles2008extended}) makes automatic differentiation
straight-forward.

\section{Decoupling Learning and Evaluation}\label{sec:decoupling}
Recall that our goal is to preserve the graph structure given through its node-to-node shortest
paths by minimizing a loss which encourages similar \emph{relative}\footnote{
  An embedding satisfying
  $d_\mathcal{M}(y_i,y_j) = \alpha\, d_G(x_i,x_j)$ (for all $i,j \in [m]$),
  for $\alpha > 0$, should be perfect.
} geodesic distances between node embeddings. Recent related work broadly uses local or global loss
functions that focus on either close neighborhood information or all-pairs interactions,
respectively. The methods that fall under the former emphasize correct placement of immediate
neighbors, such as the one used in~\cite{nickel2017poincare} for unweighted graphs:\footnote{
  We overload the sets $E$ and $\mathcal{N}(x_i)$ with index notation, assuming an arbitrary but
  fixed order over nodes.
}
\begin{equation}\label{eq:nk-loss}
  \mathcal{L}_{\textrm{neigh}}(Y) = -\sum_{(i,j) \in E} \log \frac{\exp\big(-d_\mathcal{M}(y_i,y_j)\big)}{\sum_{k \in \mathcal{N}(i)} \exp\big(-d_\mathcal{M}(y_i,y_k)\big)}.
\end{equation}
Those that fall under the latter, on the other hand, compare distances directly via loss functions
inspired by generalized MDS~\cite{bronstein2006generalized}, e.g.,\footnote{
  Note that $\mathcal{L}_{\textrm{stress}}$ focuses mostly on distant nodes while
  $\mathcal{L}_{\textrm{distortion}}$ yields larger values when close ones are misrepresented. The
  latter is one of several objectives used in~\cite{gu2018learning} (as per their code and private
  correspondence).
}
\begin{equation}\label{eq:gmds-gu}
  \mathcal{L}_{\textrm{stress}}(Y) = \sum_{i < j} \Big( d_G(x_i,x_j) - d_{\mathcal{M}}(y_i,y_j) \Big)^2
  \enskip\text{or}\enskip
  \mathcal{L}_{\textrm{distortion}}(Y)  = \sum_{i < j} \abs*{\frac{d_\mathcal{M}^2(y_i,y_j)}{d_G^2(x_i,x_j)} - 1}.
\end{equation}
The two types of objectives yield embeddings with different properties. It is thus not surprising
that each one of them has been coupled in prior work with a preferred metric quantifying
reconstruction fidelity.  The likelihood-based one is evaluated via the popular rank-based mean
average precision (mAP), while the global, stress-like ones yield best scores when measured by the
average distortion (AD) of the reference metric.
% The likelihood-based one is evaluated via the rank-based \emph{mean average precision} $\mathrm{mAP}
% = \frac{1}{m} \sum_i \frac{1}{\mathcal{N}(i)} \sum_{j \in \mathcal{N}(i)} \frac{\mathcal{N}(i) \cap
% \mathcal{B}(j; i)}{\mathcal{B}(j; i)}$, with $\mathcal{B}(j;i) = \{ y_k \in \mathcal{M} :
% d_\mathcal{M}(y_i,y_k) \leqs d_\mathcal{M}(y_i,y_j) \}$, while the global, stress-like ones yield
% best scores when measured by the \emph{average distortion} of the reference metric $D_{\textrm{avg}}
% = \frac{2}{m (m-1)} \sum_{i < j} \frac{\abs*{d_\mathcal{M}(y_i,y_j) - d_G(x_i,x_j)}}{d_G(x_i,x_j)}$.
See, e.g.,~\cite{de2018representation} for their definitions.

% \paragraph{Our Proposal.}
To decouple learning and evaluation, as well as to get both fairer and more informative comparisons
between embeddings spaces, we propose to optimize another loss function that allows
\emph{explicitly} moving in a continuous way on the representation scale ranging from ``local
neighborhoods patching,'' as encouraged by~\eqref{eq:nk-loss}, to the global topology matching, as
measured by those from~\eqref{eq:gmds-gu}. In the same spirit, we propose a more fine-grained
ranking metric that makes the trade-off clearer.

\paragraph{RSNE -- Unifying Two Disparate Regimes.}
% \paragraph{Riemannian Stochastic Neighbor Embedding.}
We advocate training embeddings via a version of the celebrated Stochastic Neighbor Embedding
(SNE)~\cite{hinton2003stochastic} adapted to the Riemannian setting. SNE works by attaching to each
node a distribution defined over all other nodes and based on the distance to them. This is done for
both the input graph distances, yielding the ground truth distribution, and for the embedding
distances, yielding the model distribution. That is, with $j \ne i$, we have
\begin{equation}\label{eq:sne-orig}
  p_{ij} \deq p(x_j \mid x_i) = \frac{\exp\big(-d_G^2(x_i, x_j) / T\big)}{Z_{p_i}}
  \enskip\text{and}\enskip
  q_{ij} \deq q(y_j \mid y_i) = \frac{\exp\big(-d_{\mathcal{M}}^2(y_i, y_j)\big)}{Z_{q_i}},
\end{equation}
where $Z_{p_i}$ and $Z_{q_i}$ are the normalizing constants and $T$ is the input scale
parameter. The original SNE formulation uses $\mathcal{M} = \R^n$. In this case, the probabilities
are proportional to an isotropic Gaussian $\mathcal{N}(y_j \mid y_i, T)$. As defined above, it
is our (natural) generalization to Riemannian manifolds -- RSNE. The embeddings are then learned by
minimizing the sum of Kullback-Leibler (KL) divergences between $p_i \deq p(\cdot \mid x_i)$ and
$q_i \deq q(\cdot \mid y_i)$:
$\mathcal{L}_{\textrm{SNE}}(Y) \deq \sum_{i=1}^m \KL\big[p_i \parallel q_i \big]$.
For $T \to 0$, it is easy to show that it recovers the \emph{local neighborhood} regime
from~\eqref{eq:nk-loss}.
% The connection to the \emph{local neighborhood} regime from~\eqref{eq:nk-loss} is stated next.
% \begin{lemma}\label{th:sne-limit}
%   For $T \to 0$, minimizing the SNE loss from above is equivalent to maximizing the sum of the
%   following per-node terms
%   \[
%     \frac{1}{\abs{\mathcal{N}(i)}} \sum_{j \in \mathcal{N}(i)} \log \frac{\exp(-d_\mathcal{M}^2(y_i,
%       y_j))}{\sum_{k \ne i} \exp(-d_\mathcal{M}^2(y_i, y_k))}.
%   \]
% \end{lemma}
% \begin{proof}
%   The result follows directly from the definition of the KL divergence, $\KL\big[p_i \parallel
%   q_i\big] = -\sum_{j \ne i} p_{ij} \log q_{ij} + \mathrm{const}$, and the limit of the
%   distributions defined in~\eqref{eq:sne-orig},
%   \[
%     \lim_{T \to 0} p_{ij} = \frac{1}{\abs{\mathcal{N}(i)}} \sum_{k \in \mathcal{N}(i)} \delta(x_k - x_j).
%   \]
%   The Euclidean intuition is that of a Gaussian becoming ``infinitely peaked'' around $x_i$, so its
%   nearest neighbors will have ``infinitely more'' mass assigned to them than the others.
% \end{proof}
% Interestingly, it has been remarked that feeding squared distances to the objective function
% improves training stability in certain cases because they are \emph{continuously}
% differentiable~\cite{de2018representation}. In this regard, \Cref{th:sne-limit} serves as a more
% principled justification for doing that in~\eqref{eq:nk-loss}.
% Finally, we point out that a connection to an MDS-like loss function is mentioned
% in~\citep[][Section~6]{hinton2003stochastic}, in the regime $T \to \infty$, but we have not
% been able to make sense out of it.
For a large $T$, the SNE objective tends towards placing equal emphasis on the relative
distances between all pairs of points, thus behaving similar to the MDS-like loss
functions~\eqref{eq:gmds-gu}~\citep[][Section~6]{hinton2003stochastic}. What we have gained is that
the temperature parameter $T$ acts as a knob for controlling the optimization goal.

\paragraph{F1@k -- Generalizing Ranking Fidelity.}
We generalize the ranking fidelity metric in the spirit of mAP@k (e.g.,~\cite{gu2018learning}) for
nodes that are $k$ hops away from a source node, with $k > 1$.  Recall that the motivation stems,
for one, from the limitation of mean average precision to immediate neighbors, and, at the other
side of the spectrum, from the sensitivity to absolute values of non-ranking metrics such as the
average distortion. For an unweighted\footnote{
  We have mostly used unweighted graphs in our experiments, so here we restrict the treatment as
  such.
}
graph $G$, we denote by $L_G(u; k)$ the set of nodes that are exactly $k$ hops away from a source
node $u$ (i.e., ``on layer $k$''), and by $\mathcal{B}_G(v; u)$ the set of nodes that are closer to
node $u$ than another node $v$. Then, for an embedding $f : G \to \mathcal{M}$, the \emph{precision}
and \emph{recall} of a node $v$ in the shortest-path tree rooted at $u$, with $u \ne v$, are given
by
\begin{equation}
  P(v; u) \deq \frac{\abs{\mathcal{B}_G(v; u) \cap \mathcal{B}_{\mathcal{M}}(f(v); f(u))}}{\abs{\mathcal{B}_{\mathcal{M}}(f(v); f(u))}}
  \enskip\text{and}\enskip
  R(v; u) \deq \frac{\abs{\mathcal{B}_G(v; u) \cap \mathcal{B}_{\mathcal{M}}(f(v); f(u))}}{\abs{\mathcal{B}_G(v; u)}}.
\end{equation}
They follow the conventional definitions. For instance, the numerator is the number of true
positives: the nodes that appear before $v$ in the shortest-path tree rooted at $u$ and, at the same
time, are embedded closer to $u$ than $v$ is. The definition of the F1 score of $(u, v)$, denoted by
$F_1(v; u)$, follows naturally as the harmonic mean of precision and recall. Then, the F1@k metric
is obtained by averaging the F1 scores of all nodes that are on layer $k \geqs 1$, across all
shortest-path trees. That is, $F_1(k) \deq \frac{1}{c(k)} \sum_{u \in G}\ \sum_{v \in L_G(u; k)} F_1(v;
u)$, with $c(k) = \sum_{u \in G} \abs{L_G(u; k)}$. This draws a curve $\{(k, F_1(k))\}_{k \in [d(G)]}$,
where $d(G)$ is the diameter of the graph.

\section{Experiments}\label{sec:experiments}
We restrict our experiments here to evaluating the graph reconstruction capabilities of the proposed
matrix manifolds relative to the constant curvature baseline spaces.  A thorough analysis via
properties of nearest-neighbor graphs constructed from manifold random samples, inspired
by~\cite{krioukov2010hyperbolic}, is included in~\Cref{sec:man-rand-graphs}. It shows that the two
matrix manifolds lead to distinctive network structures.\footnote{Our code is accessible at
\url{https://github.com/dalab/matrix-manifolds}.}

\vspace*{-0.2cm}
\paragraph{Training Details \& Evaluation.}
We start by computing all-pairs shortest-paths in all input graphs, performing max-scaling, and
serializing them to disk. Then, for each manifold, we optimize a set of embeddings for several
combinations of optimization settings and loss functions, including both the newly proposed
Riemannian SNE, for several values of $T$, and the ones used in prior work (\Cref{sec:decoupling}).
Finally, because we are ultimately interested in the representation power of each embedding space,
we report the best F1@1, area under the F1@k curve (AUC), and average distortion (AD), across those
repetitions. This is in line with the experimental framework from prior
works~\cite{nickel2017poincare,gu2018learning} with the added benefit of treating the objective
function as a nuisance. More details are included in~\Cref{sec:exp-setup}.

\begin{figure}
  \centering
  \subfloat{{\includegraphics[scale=0.271]{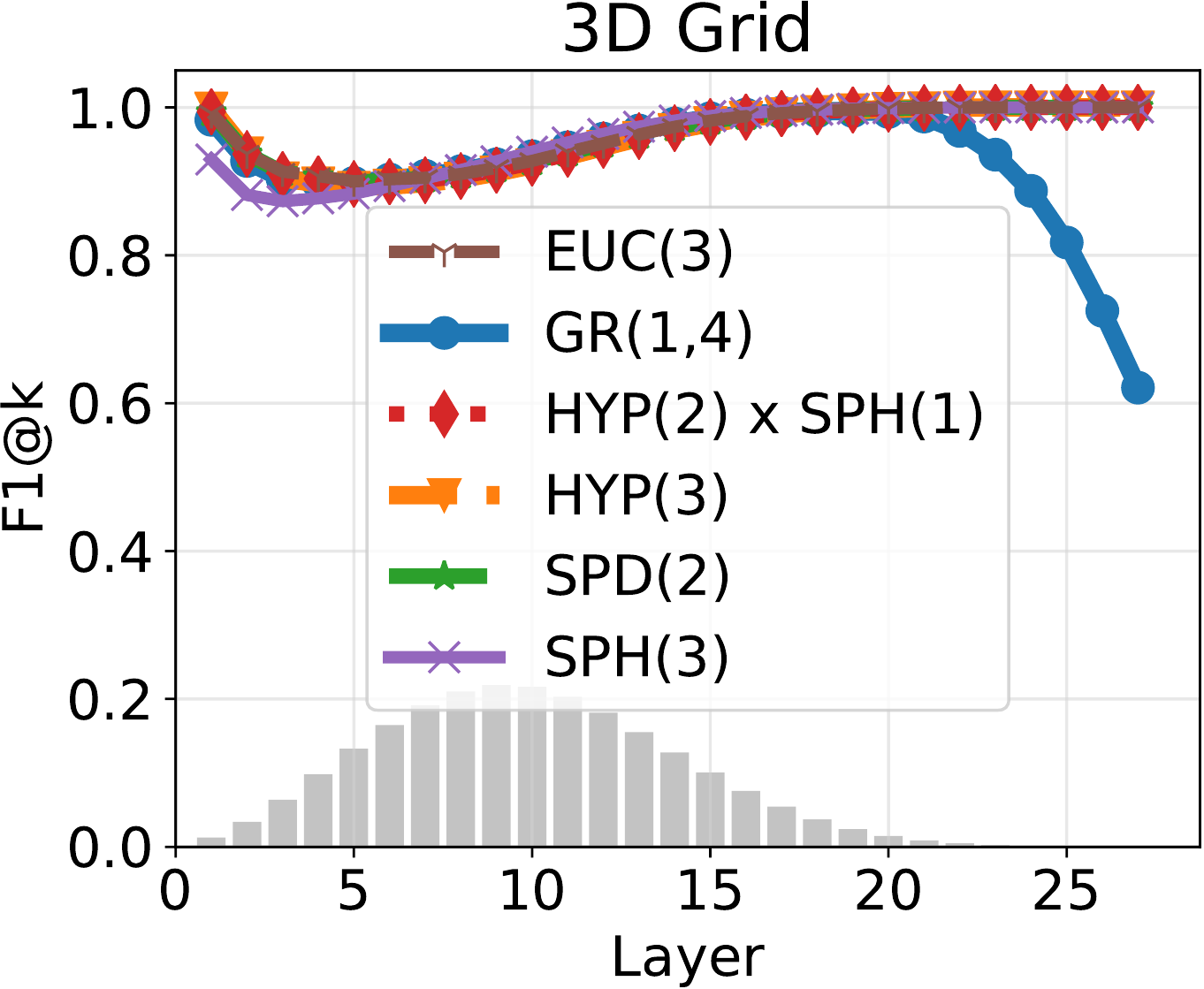} }}
  \subfloat{{\includegraphics[scale=0.271]{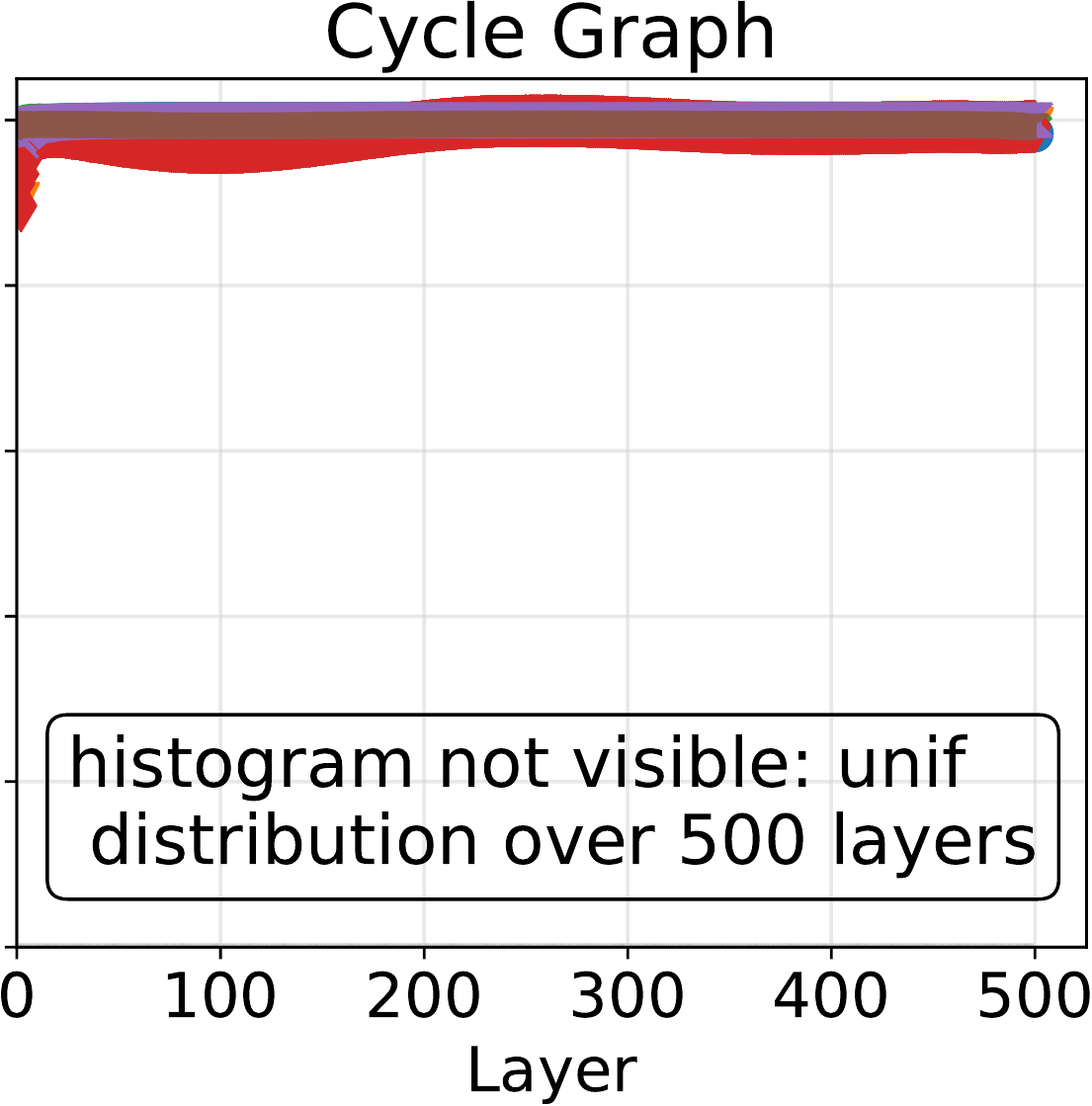} }}
  \subfloat{{\includegraphics[scale=0.271]{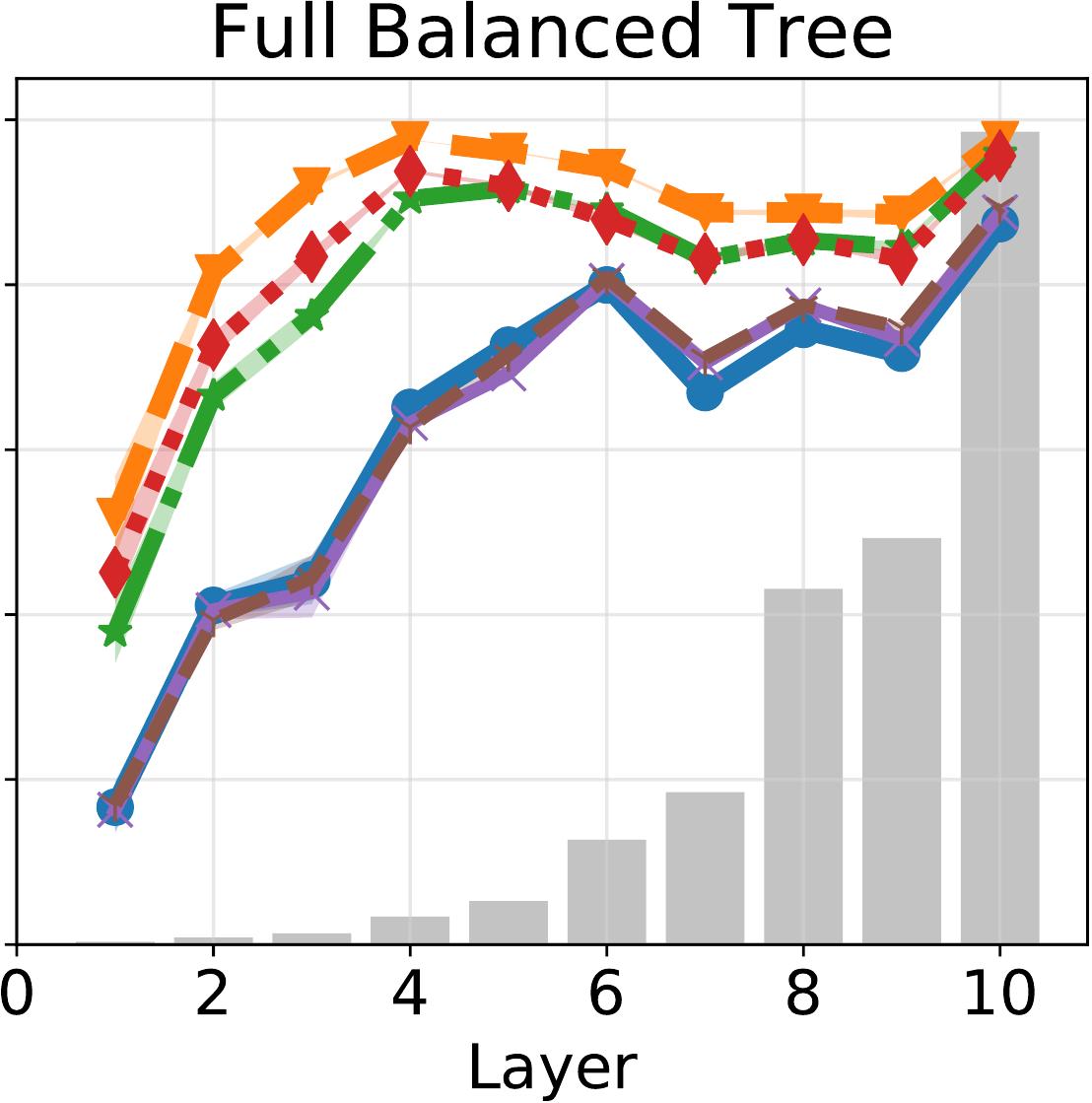} }}
  \subfloat{{\includegraphics[scale=0.271]{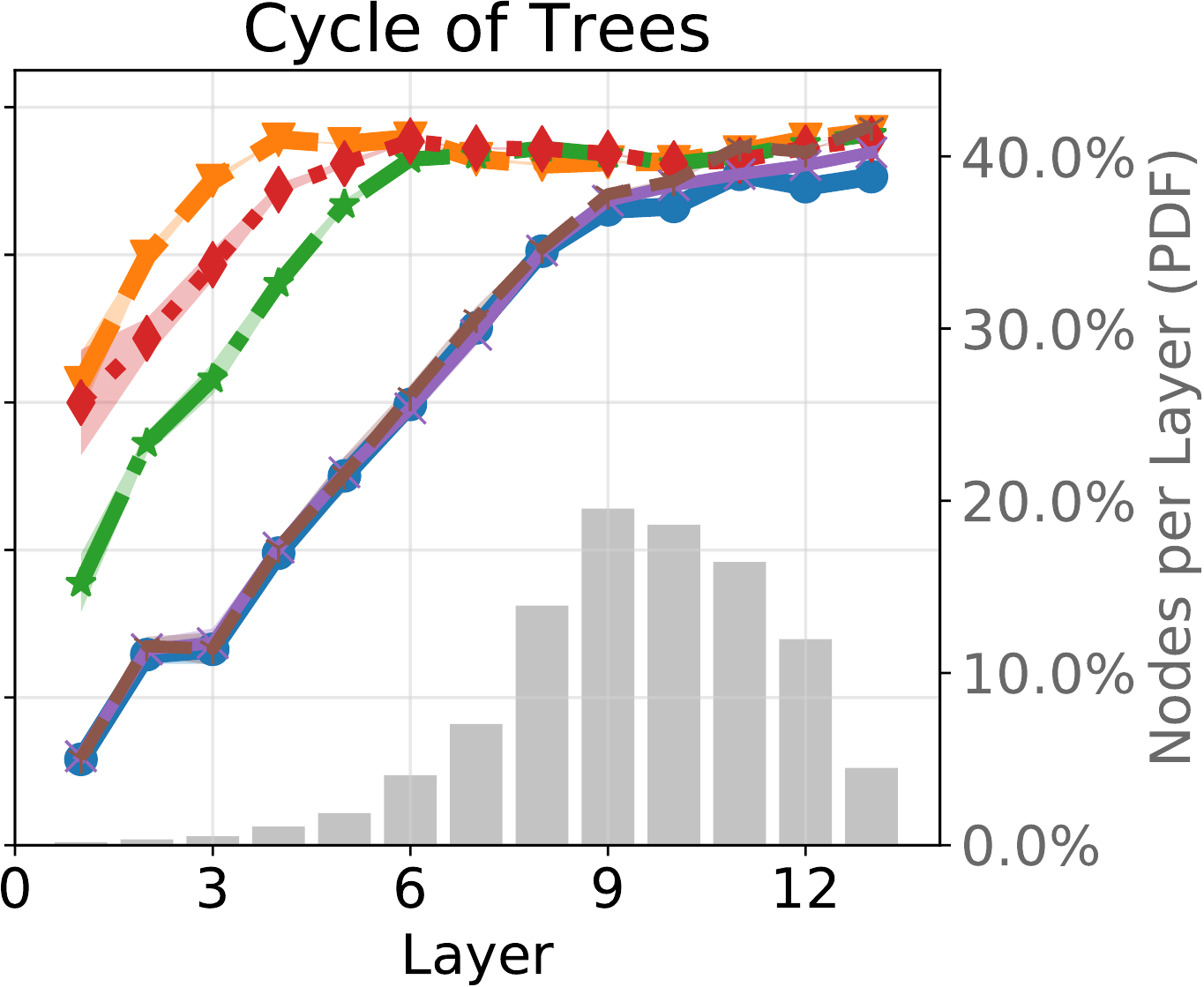} }}
  \caption{F1@k curves (left $y$-axis) and PMFs of node pairs per hop count (right $y$-axis) for
  several synthetic graphs. The objective was RSNE at high temperature $T$.}
  \label{fig:synthetic-graphs}
\end{figure}

\vspace*{-0.2cm}
\paragraph{Synthetic Graphs.}
We begin by showcasing the F1@k metric for several generated graphs in~\Cref{fig:synthetic-graphs}.
On the $10 \times 10 \times 10$ grid and the $1000$-nodes cycle all manifolds perform well. This is
because every Riemannian manifold generalizes Euclidean space and Euclidean geometry suffices for
grids and cycles (e.g., a cycle looks locally like a line). The more discriminative ones are the two
other graphs -- a full balanced tree (branching factor $r=4$ and depth $h=5$) and a cycle of 10
trees ($r=3$ and $h=4$). The best performing embeddings involve a hyperbolic component while the SPD
ones rank between those and the non-negatively curved ones (which are indistinguishable). The
results confirm our expectations: (more) negative curvature is useful when embedding trees. Finally,
notice that the high-temperature RSNE regime used here encourages the recovery of the global
structure (high AUC F1@k) more than the local neighborhoods (low individual F1@k values for small
$k$).

\begin{figure}[!t]
\begin{minipage}[t]{0.41\linewidth}
  \captionof{table}{
    The results for ``$\Spd$ vs.\ $\Hyp$''. Best / second-best results are highlighted. The ``Stein
    manifold'' is SPD trained with the Stein divergence (see text). The F1@1 and AUC metrics are
    multiplied by 100.
  }
  \label{tab:noncomp-recon-results}
  \vspace*{0.1in}

  \small
  \renewcommand{\arraystretch}{1.0}

  \begin{tabularx}{\textwidth}{@{}cclccc@{}}
    \toprule
    $G$ & $n$ & $\mathcal{M}$ & \textbf{F1@1} & \textbf{AUC} & \textbf{AD} \\

    \midrule
    \multirow{8}{*}{\normalsize\bfseries \rotatebox[origin=c]{90}{facebook}} &
    \multirow{4}{*}{3}
    & Euc & 70.28 & 95.27 & 0.193\\
    && Hyp & 71.08 & 95.46 & 0.173\\
    && SPD & 71.09 & 95.26 & 0.170\\
    && Stein & \cellcolor{col1}75.91 & 95.59 & \cellcolor{col1}0.114 \\
    \cmidrule{2-6}
    & \multirow{4}{*}{6}
    & Euc & 79.60 & 96.41 & 0.090\\
    && Hyp & \cellcolor{col2}81.83 & 96.53 & 0.089\\
    && SPD & 79.52 & 96.37 & 0.090\\
    && Stein & \cellcolor{col1}83.95 & 96.74 & \cellcolor{col1}0.061 \\

    \midrule
    \multirow{8}{*}{\normalsize\bfseries \rotatebox[origin=c]{90}{web-edu}} &
    \multirow{4}{*}{3}
    & Euc & 29.18 & 87.14 & 0.245\\
    && Hyp & \cellcolor{col1}55.60 & \cellcolor{col1}92.10 & 0.245\\
    && SPD & 29.02 & 88.54 & 0.246\\
    && Stein & \cellcolor{col2}48.28 & \cellcolor{col2}90.87 & \cellcolor{col1}0.084\\
    \cmidrule{2-6}
    & \multirow{4}{*}{6}
    & Euc & 49.31 & 91.19 & 0.143\\
    && Hyp & \cellcolor{col1}66.23 & \cellcolor{col2}95.78 & 0.143\\
    && SPD & 42.16 & 91.90 & 0.142\\
    && Stein & \cellcolor{col2}62.81 & \cellcolor{col1}96.51 & \cellcolor{col1}0.043\\

    \midrule
    \multirow{8}{*}{\normalsize\bfseries \rotatebox[origin=c]{90}{bio-diseasome}} &
    \multirow{4}{*}{3}
    & Euc & 83.78 & 91.21 & 0.145\\
    && Hyp & \cellcolor{col1}86.21 & \cellcolor{col1}95.72 & 0.137\\
    && SPD & 83.99 & 91.32 & 0.140\\
    && Stein & \cellcolor{col1}86.70 & \cellcolor{col2}94.54 & \cellcolor{col1}0.105\\
    \cmidrule{2-6}
    & \multirow{4}{*}{6}
    & Euc & 93.48 & 95.84 & 0.073\\
    && Hyp & \cellcolor{col1}96.50 & \cellcolor{col1}98.42 & 0.071\\
    && SPD & 93.83 & 95.93 & 0.072\\
    && Stein & \cellcolor{col2}94.86 & \cellcolor{col2}97.64 & \cellcolor{col1}0.066\\

    \midrule
    \multirow{8}{*}{\normalsize\bfseries \rotatebox[origin=c]{90}{power}} &
    \multirow{4}{*}{3}
    & Euc & 49.34 & 87.84 & 0.119 \\
    && Hyp & \cellcolor{col1}60.18 & \cellcolor{col1}91.28 & \cellcolor{col1}0.068 \\
    && SPD & 52.48 & \cellcolor{col2}90.17 & 0.121 \\
    && Stein & \cellcolor{col2}54.06 & \cellcolor{col2}90.16 & \cellcolor{col2}0.076 \\
    \cmidrule{2-6}
    & \multirow{4}{*}{6}
    & Euc & 63.62 & 92.09 & 0.061 \\
    && Hyp & \cellcolor{col1}75.02 & \cellcolor{col1}94.34 & 0.060 \\
    && SPD & 67.69 & 91.76 & 0.062 \\
    && Stein & \cellcolor{col2}70.70 & \cellcolor{col2}93.32 & \cellcolor{col1}0.049 \\

    \bottomrule
  \end{tabularx}
\end{minipage}\hfill
\begin{minipage}[t]{0.55\linewidth}
  \begin{minipage}[t]{\linewidth}
    \captionof{table}{
      The results for ``$\Gr$ vs.\ $\Sph$''. The dataset ``cat-cortex''~\cite{scannell1995analysis}
      is a dissimilarity matrix, lacking graph connectivity information, so F1@k cannot be computed.
    }
    \label{tab:compact-recon-results}
    \vspace*{0.1in}

    \small
    \centering
    \renewcommand{\arraystretch}{1.0}

    \begin{tabular}{@{}cclcccc@{}}
      \toprule
      $G$ & $n$ & $\mathcal{M}$ & \textbf{F1@1} & \textbf{AUC} & \textbf{AD} \\

      \midrule
      \multirow{8}{*}{\normalsize\bfseries \rotatebox[origin=c]{90}{road-minnesota}} &
      \multirow{2}{*}{2}
      & Sphere & \cellcolor{col1}82.19 & 94.02 & 0.085\\
      && $\Gr(1,3)$ & 78.91 & 94.02 & 0.085\\
      \cmidrule{2-6}
      & \multirow{2}{*}{3}
      & Sphere & 89.55 & 95.89 & 0.059\\
      && $\Gr(1,4)$ & \cellcolor{col2}90.02 & 95.88 & 0.058\\
      \cmidrule{2-6}
      & \multirow{3}{*}{4}
      & Sphere & 93.65 & 96.66 & 0.049\\
      && $\Gr(1,5)$ & 93.89 & 96.67 & 0.049\\
      && $\Gr(2,4)$ & \cellcolor{col2}94.01 & 96.66 & 0.049\\

      \midrule
      \multirow{8}{*}{\normalsize\bfseries \rotatebox[origin=c]{90}{cat-cortex}} &
      \multirow{2}{*}{2}
      & Sphere & {-} & {-} & 0.255\\
      && $\Gr(1,3)$ & {-} & {-} & \cellcolor{col2}0.234\\
      \cmidrule{2-6}
      & \multirow{2}{*}{3}
      & Sphere & {-} & {-} & 0.195\\
      && $\Gr(1,4)$ & {-} & {-} & \cellcolor{col1}0.168\\
      \cmidrule{2-6}
      & \multirow{3}{*}{4}
      & Sphere & {-} & {-} & 0.156\\
      && $\Gr(1,5)$ & {-} & {-} & \cellcolor{col2}0.139\\
      && $\Gr(2,4)$ & {-} & {-} & \cellcolor{col1}0.129\\

      \bottomrule
    \end{tabular}
  \end{minipage}
  \vspace*{0.5cm}

  \begin{minipage}[t]{\linewidth}
    \centering
    \includegraphics[scale=0.24]{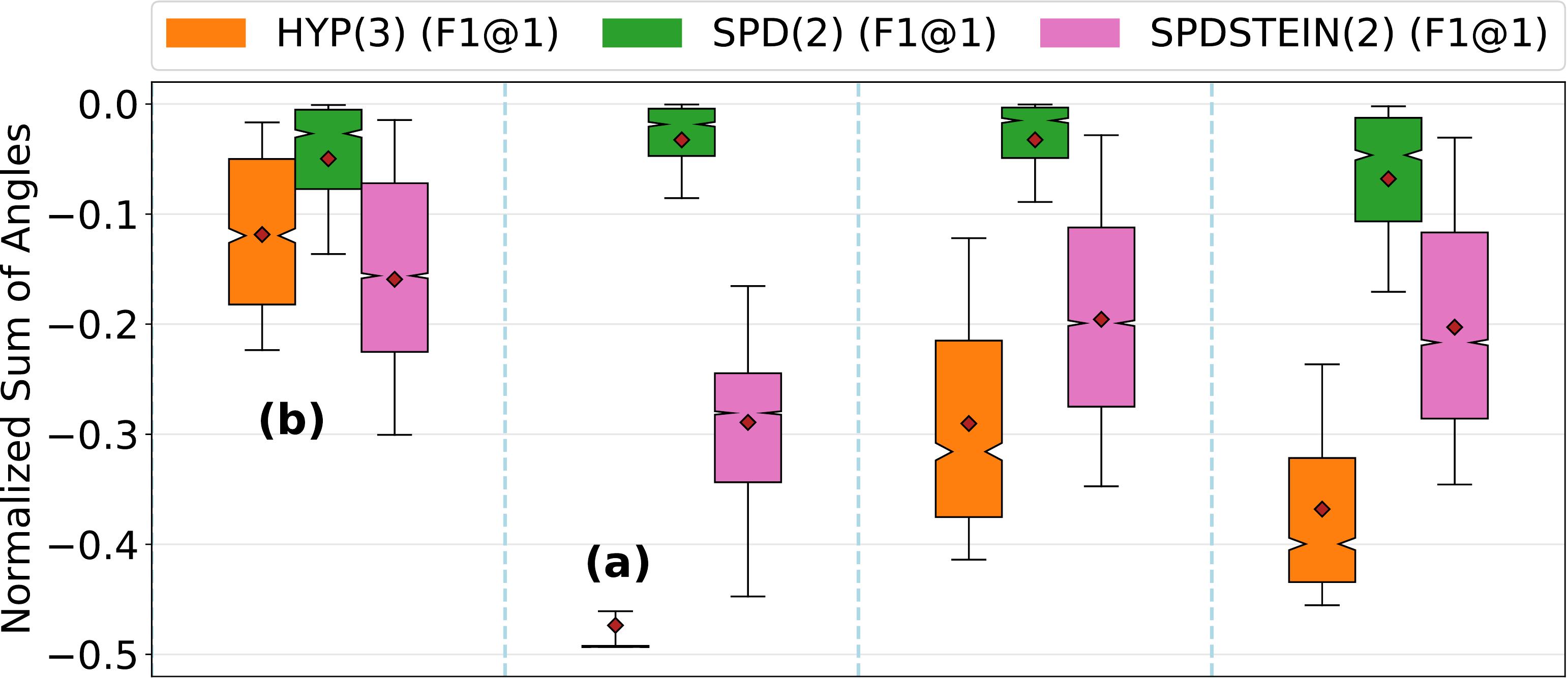}
    \includegraphics[scale=0.24]{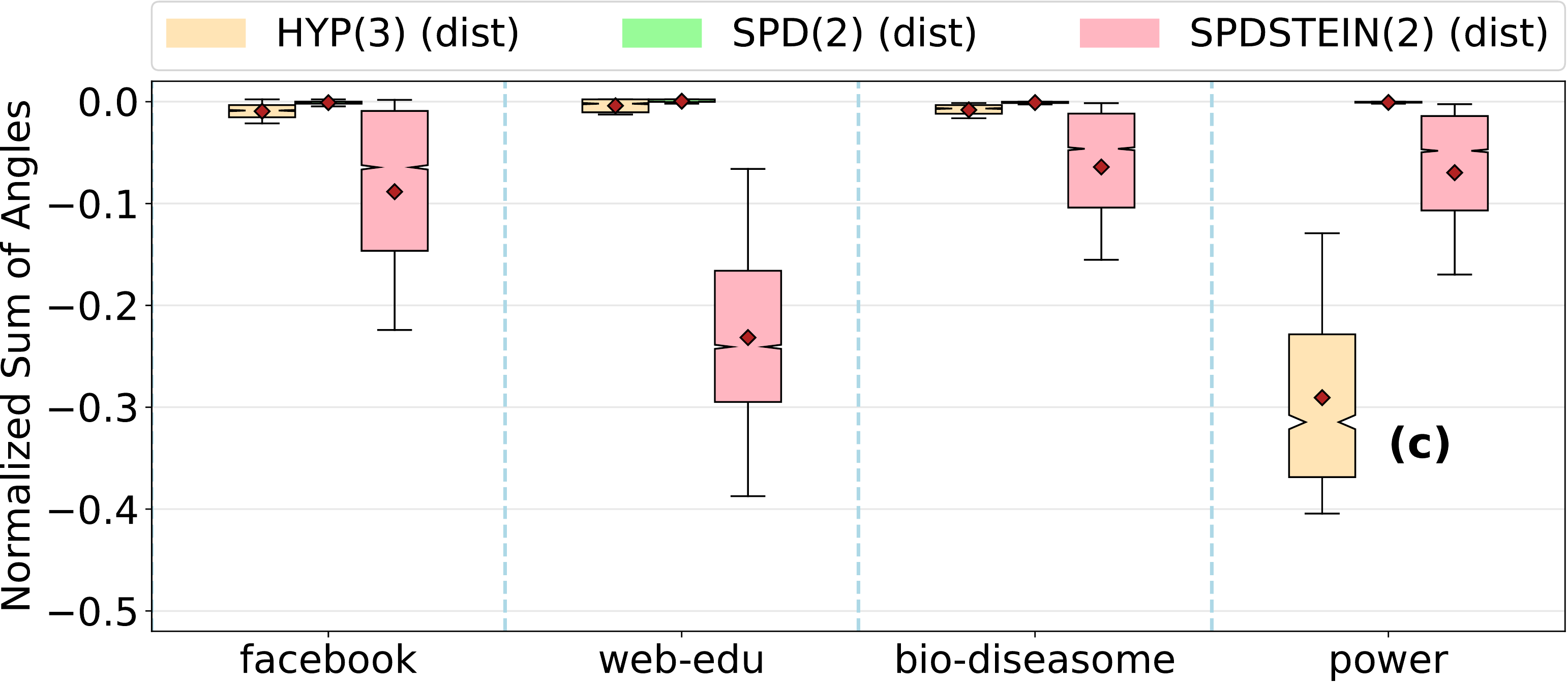}

    \captionof{figure}{
      Distributions of (normalized) sum-of-angles in geodesic triangles formed by the learned
      embeddings that yield the best F1@1 metrics (up) and the best average distortion metrics
      (down), for all datasets from~\Cref{tab:noncomp-recon-results}, for $n=3$. 10000 triples are
      sampled.
    }
    \label{fig:noncomp-angles}
  \end{minipage}
\end{minipage}
\end{figure}

\vspace*{-0.2cm}
\paragraph{Non-positive Curvature.}
We compare the Euclidean, hyperbolic, and SPD spaces on several real datasets
in~\Cref{tab:noncomp-recon-results}. For the SPD manifold, we experiment with both the canonical
distance function and the (related) S-divergence as model metrics. When performing Riemannian
optimization, we use the same canonical Riemannian tools (as per~\Cref{tab:diff-geom-summary}).
More details about the graphs and an analysis of their geometric properties are attached
in~\Cref{sec:input-graphs}. Two of the ones shown here are plotted in~\Cref{fig:example-graphs}.
Extended results are included in~\Cref{sec:extended-results}. First of all, we see that the
(partial) negative curvature of the SPD and hyperbolic manifolds is beneficial: they outperform the
flat Euclidean embeddings in almost all scenarios. This can be explained by the apparent scale-free
nature of the input graphs~\cite{krioukov2010hyperbolic}. Second, we see that especially when using
the S-divergence, which we attribute to the better-behaved optimization task thanks to its geodesic
convexity and stable gradients (see~\Cref{sec:nonpos}), the SPD embeddings achieve significant
improvements on the average distortion metric and are competitive and sometimes better on the
ranking metrics.

\vspace*{-0.2cm}
\paragraph{How Do the Embeddings Curve?}
Since any manifold locally resembles Euclidean space, it is a priori unclear to what extent its
theoretical curvature is leveraged by the embeddings. To shed light on that, we employ the analysis
technique based on sum-of-angles in geodesic triangles from~\Cref{sec:background}. We recognize
in~\Cref{fig:noncomp-angles} a remarkably consistent pattern: the better performing embeddings (as
per~\Cref{tab:noncomp-recon-results}) yield more negatively-curved triangles. Notice, for instance,
the collapsed box plot corresponding to the ``web-edu'' hyperbolic embedding \textbf{(a)}, i.e.,
almost all triangles sampled have sum-of-angles close to 0 . This is explained by its obvious
tree-like structure (\Cref{fig:web-edu}). Similarly, the SPD-Stein embedding of ``facebook''
outperforms the hyperbolic one in terms of F1@1 and that reflects in the slightly more stretched box
plot \textbf{(b)}. Moreover, the pattern applies to the best average-distortion embeddings, where
the SPD-Stein embeddings are the only ones that make non-negligible use of negative curvature and,
hence, perform better -- the only exception is the ``power'' graph \textbf{(c)}, for which
indeed~\Cref{tab:noncomp-recon-results} confirms that the hyperbolic embeddings are slightly better.
% Thus, this analysis suggests that curvature around the learned embeddings, estimated via
% sum-of-angles distributions, is a predictor of graph embedding quality measured by both
% ranking-based metrics and average distortion.

\vspace*{-0.2cm}
\paragraph{Compact Embeddings.}
We embed several graphs with traits associated with positive curvature in Grassmann manifolds and
compare them to spherical embeddings. \Cref{tab:compact-recon-results} shows that the former yields
non-negligibly lower average distortion on the ``cat-cortex'' dissimilarity dataset and that the two
are on-par on the ``road-minnesota'' graph (displayed in~\Cref{fig:road-minnesota} -- notice its
particular structure, characterized by cycles and low node degrees). More such results are included
in~\Cref{sec:extended-results}. As a general pattern, we find learning compact embeddings to be
optimization-unfriendly.

\section{Conclusion \& Future Work}
We proposed to use the SPD and Grassmann manifolds for learning representations of graphs and showed
that they are competitive against previously considered constant-curvature spaces on the graph
reconstruction task, consistently and significantly outperforming them in some cases. Our results
suggest that their geometry can accommodate certain graphs with better precision and less distortion
than other embedding spaces. We thoroughly described their properties, emphasizing those that set
them apart, and worked out the practically challenging aspects. Moreover, we advocate the Riemannian
SNE objective for learning embeddings as a way to unify two different families of loss functions
used in recent related works. It allows practitioners to explicitly tune the desired optimization
goal by adjusting the temperature parameter. Finally, we defined the F1@k metric as a more general
way of quantifying ranking fidelity.

Our work is related to (and further motivates) some fundamental research questions. How does the
curvature of a Riemannian manifold influence the types of metrics that it can represent? Can we
design a theoretical framework that connects them to discrete metric spaces represented by graphs?
How would a faithful embedding influence downstream tasks, such as node classification or link
prediction? These are some of the questions we are excited about and plan to pursue in future work.

\section*{Broader Impact}
Our research deals with a deeply technical question: how can we leverage geometry to better
represent graphs? It is also part of a nascent area of machine learning that tries to improve
existing methods by bringing forward geometry tools and theories which have been known in
mathematics for a (relatively) long time. That being said, should it later materialize into
practically useful applications, its impact can be significant. That is due to the ubiquity of
graphs as models of data and the possibility for our research to improve graph-based models. To give
several examples, our research can have a broader impact in network analysis (in particular, social
networks), working with biological data (e.g., proteins, molecules), or learning from knowledge
graphs. The impact of our work can be beneficial, for instance, when having more faithful graph
models can lead to improved recommender systems or drug discovery. However, we emphasize that
current models might suffer from various other modes of failure and, thus, are not capable of
replacing human expertise and intervention.

%% ONLY IF ACCEPTED
% \input{src/sections/software}
\section*{Acknowledgements}
We would like to thank Andreas Bloch for suggesting the curvature quantification approach based on
the sum of angles in geodesic triangles. We are grateful to Prof.\ Thomas Hofmann for making this
collaboration possible. We thank the anonymous reviewers for helping us improve this work.

Gary B\'{e}cigneul is funded by the Max Planck ETH Center for Learning Systems.

\bibliography{matrix-manifolds}
\bibliographystyle{plain}

\clearpage
\appendix
\section{Overview of Differential \& Riemannian Geometry}\label{sec:diff-geom}
In this section, we introduce the foundational concepts from differential geometry, a discipline
that has arisen from the study of differentiable functions on curves and surfaces, thus generalizing
calculus. Then, we go a step further, into the more specific Riemannian geometry which enables the
abstract definitions of lengths, angles, and curvatures. We base this section
on~\cite{carmo1992riemannian,absil2009optimization} and point the reader to them for a more thorough
treatment of the subject.

%%%
\paragraph{Differentiable Manifolds.}
Informally, an $n$-dimensional manifold is a set $\mathcal{M}$ which locally resembles
$n$-dimensional Euclidean space. To be formal, one first introduces the notions of charts and
atlases. A bijection $\phi$ from a subset $\mathcal{U} \in \mathcal{M}$ onto an open subset of
$\R^n$ is called an $n$-dimensional chart of the set $\mathcal{M}$, denoted by $(\mathcal{U},
\phi)$. It enables the study of points $x \in \mathcal{M}$ via their coordinates $\phi(x) \in \R^n$.
A differentiable atlas $\mathcal{A}$ of $\mathcal{M}$ is a collection of charts
$(\mathcal{U}_\alpha, \phi_\alpha)$ of the set $\mathcal{M}$ such that
\begin{enumerate}[(i)]
  \item $\bigcup_\alpha \mathcal{U}_\alpha = \mathcal{M}$
  \item For any pair $\alpha, \beta$ with $\mathcal{U}_\alpha \cap \mathcal{U}_\beta \ne \emptyset$,
    the sets $\phi_\alpha(\mathcal{U}_\alpha \cap \mathcal{U}_\beta)$ and
    $\phi_\beta(\mathcal{U}_\alpha \cap \mathcal{U}_\beta)$ are open sets in $\R^n$ and the change
    of coordinates $\phi_\beta \circ \phi_\alpha^{-1}$ is smooth.
\end{enumerate}
Two atlases $\mathcal{A}_1$ and $\mathcal{A}_2$ are equivalent if they generate the same
maximal atlas. The maximal atlas $\mathcal{A}^+$ is the set of all charts $(\mathcal{U}, \phi)$ such
that $\mathcal{A} \cup \{(\mathcal{U}, \phi)\}$ is also an atlas. It is also called a differentiable
structure on $\mathcal{M}$. With that, an $n$-dimensional differentiable manifold is a couple
$(\mathcal{M}, \mathcal{A}^+)$, with $\mathcal{M}$ a set and $\mathcal{A}^+$ a maximal atlas of
$\mathcal{M}$ into $\R^n$. In more formal treatments, $\mathcal{A}^+$ is also constrained to induce
a well-behaved topology on $\mathcal{M}$.

%%%
\paragraph{Embedded Submanifolds and Quotient Manifolds.}
How is a differentiable structure for a set of interest usually constructed? From the definition
above it is clear that it is something one endows the set with. That being said, in most useful
cases it is not explicitly chosen or constructed. Instead, a rather recursive approach is taken:
manifolds are obtained by considering either subsets or quotients (see last subsection paragraph) of
other manifolds, thus inheriting a ``natural'' differentiable structure. Where does this recursion
end? It mainly ends when one reaches a vector space, which is trivially a manifold via the global
chart $\phi : \R^{n \times k} \to \R^{n k}$, with $X \mapsto \mathrm{vec}(X)$. That is the case for
the matrix manifolds considered in~\Cref{sec:matrix-manifolds} too.

What makes the aforementioned construction approach (almost) assumption-free is the following
essential property: if $\mathcal{M}$ is a manifold and $\mathcal{N}$ is a subset of the set
$\mathcal{M}$ (respectively, a quotient $\mathcal{M}/{\sim}$), then there is at most one
differentiable structure that agrees with the subset topology (respectively, with the quotient
projection).\footnote{We say almost assumption-free because we still assume that this agreement is
desirable.} The resulting manifolds are called \emph{embedded submanifolds} and \emph{quotient
manifolds}, respectively. Sufficient conditions for their existence (hence, uniqueness) are known
too and do apply for our manifolds. For instance, the \emph{submersion theorem} says that for a
smooth function $F : \mathcal{M}_1 \to \mathcal{M}_2$ with
$\dim(\mathcal{M}_1) = d_1 > d_2 = \dim(\mathcal{M}_2)$
and a point $y \in \mathcal{M}_2$ such that $F$ has full rank\footnote{That is, the Jacobian
$\frac{\partial F(x)}{\partial x} \in \R^{d_2 \times d_1}$ has rank $d_2$ irrespective of the chosen
charts.} for all $x \in F^{-1}(y)$, its preimage $F^{-1}(y)$ is a closed embedded submanifold of
$\mathcal{M}_1$ and $\dim(F^{-1}(y)) = d_1 - d_2$.

The quotient of a set $\mathcal{M}$ by an equivalence relation $\sim$ is defined as
$\mathcal{M}/{\sim} \deq \big\{ [x] : x \in \mathcal{M} \big\}$,
with $[x] \deq \{ y \in \mathcal{M} : y \sim x \}$ -- the equivalence class of $x$. The function
$\pi : \mathcal{M} \to \mathcal{M}/{\sim}$, given by $x \mapsto [x]$, is the canonical projection.
The simplest example of a quotient manifold is the real projective space, $\RP(n-1)$. It is the set
of lines through the origin in $\R^n$. With the notation $\R_\ast^n = \R^n \setminus \{0\}$, the
real projective space can be identified with the quotient $\R_\ast^n/{\sim}$ given by the
equivalence relation
$x \sim y \iff \exists t \in \R_\ast : y = t x$.

%%%
\paragraph{Tangent Spaces.}
To do even basic calculus on a manifold, one has to properly define the derivatives of manifold
curves, $\gamma : \R \to \mathcal{M}$, as well as the directional derivatives of smooth real-valued
functions defined on the manifold, $f : \mathcal{M} \to \R$. The usual definitions,
\begin{equation}\label{eq:usual-derivs}
  \gamma'(t) \deq \lim_{\tau \to 0} \frac{\gamma(t + \tau) - \gamma(t)}{\tau}
  \enskip\text{and}\enskip
  \Diff f(x) [\eta] \deq \lim_{t \to 0} \frac{f(x + t \eta) - f(x)}{t},
\end{equation}
are invalid as such because addition does not make sense on general manifolds. However, notice that
$f \circ \gamma : t \mapsto f(\gamma(t))$ is differentiable in the usual sense. With that in mind,
let $\mathfrak{T}_x(\mathcal{M})$ denote the set of smooth real-valued functions defined on a
neighborhood of $x$. Then, the mapping $\dot{\gamma}(0)$ from $\mathfrak{T}_x(\mathcal{M})$ to $\R$
defined by
\begin{equation}\label{eq:curv-deriv}
  \dot{\gamma}(0) f \deq \left.\frac{\mathrm{d} f(\gamma(t))}{\mathrm{d}t}\right\vert_{t=0}
\end{equation}
is called the tangent vector to the curve $\gamma$ at $t = 0$. The equivalence class
\begin{equation}
  [\dot{\gamma}(0)] \deq \{ \gamma_1 : \R \to \mathcal{M} : \dot{\gamma}_1(0) f = \dot{\gamma}(0) f,\ \forall f \in \mathfrak{T}_x \}
\end{equation}
is a \emph{tangent vector} at $x \in \mathcal{M}$. The set of such equivalence classes forms the
\emph{tangent space} $T_x \mathcal{M}$. It is immediate from the linearity of the differentiation
operator from~\eqref{eq:curv-deriv} that $T_x \mathcal{M}$ inherits a linear structure, thus forming
a vector space.

The abstract definition from above recovers the classical definition of directional derivative
from~\eqref{eq:usual-derivs} in the following sense: if $\mathcal{M}$ is an embedded submanifold of
a vector space $\mathcal{E}$ and $\overline{f}$ is the extension of $f$ in a neighborhood of
$\gamma(0) \in \mathcal{E}$, then
\begin{equation}
  \dot{\gamma}(0) f = \Diff \overline{f}(\gamma(0)) [\gamma'(0)],
\end{equation}
so there is a natural identification of the mapping $\dot{\gamma}(0)$ with the vector $\gamma'(0)$.

For quotient manifolds $\mathcal{M}/{\sim}$, the tangent space splits into two complementary linear
subspaces called the \emph{vertical space} $\mathcal{V}_x$ and the \emph{horizontal space}
$\mathcal{H}_x$. Intuitively, the vectors in the former point in tangent directions which, if we
were to ``follow'' for an infinitesimal step, we would get another element of $[x]$. Thus, only the
horizontal tangent vectors make us ``move'' on the quotient manifold.

%%%
\paragraph{Riemannian Metrics.}
They are inner products $\langle \cdot, \cdot \rangle_x$, sometimes denoted by $g_x(\cdot, \cdot)$,
attached to each tangent space $T_x \mathcal{M}$. They give a notion of length via $\norm*{\xi_x}_x
\deq \sqrt{\langle \xi_x, \xi_x \rangle_x}$, for all $\xi_x \in T_x \mathcal{M}$. A Riemannian
metric is an additional structure added to a differentiable manifold $(\mathcal{M}, \mathcal{A}^+)$,
yielding the Riemannian manifold $(\mathcal{M}, \mathcal{A}^+, g_x)$. However, as it was the case
for the differentiable structure, for Riemannian submanifolds and Riemannian quotient manifolds it
is inherited from the ``parent'' manifold in a natural way -- see~\cite{absil2009optimization} for
several examples.

The Riemannian metric enables measuring the length of a curve $\gamma : [a, b] \to \mathcal{M}$,
\begin{equation}
  L(\gamma) = \int_a^b \sqrt{\langle \dot{\gamma}(t), \dot{\gamma}(t) \rangle_{\gamma(t)}} \mathrm{d}t,
\end{equation}
which, in turn, yields the \emph{Riemannian distance} function,
\begin{equation}
  d_\mathcal{M} : \mathcal{M} \times \mathcal{M} \to \R,\enskip d_\mathcal{M}(x, y) = \inf_\gamma L(\gamma),
\end{equation}
that is, the shortest path between two points on the manifold. The infimum is taken over all curves
$\gamma : [a, b] \to \mathcal{M}$ with $\gamma(a) = x$ and $\gamma(b) = y$. Note that in general it
is only defined locally because $\mathcal{M}$ might have several connected components or it might
not be geodesically complete (see next paragraphs). \emph{Deriving a closed-form expression of it is
paramount for graph embedding purposes}.

The \emph{Riemannian gradient} of a smooth function $f : \mathcal{M} \to \R$ at $x$, denoted by
$\nabla^R f(x)$ is defined as the unique element of $T_x \mathcal{M}$ that satisfies
\begin{equation}
  \langle \nabla^R f(x), \xi \rangle_x = \Diff f(x) [\xi],\enskip \forall \xi \in T_x \mathcal{M}.
\end{equation}

%%%
\paragraph{Retractions.}
Up until this point, we have not introduced the concept of ``moving in the direction of a tangent
vector'', although we have appealed to this intuition. This is formally achieved via retractions. At
a point $x \in \mathcal{M}$, the retraction $R_x$ is a map from $T_x \mathcal{M}$ to $\mathcal{M}$
satisfying local rigidity conditions: $R_x(0) = x$ and $\Diff R_x(0) = \eye_n$. For embedded
submanifolds, the two-step approach consisting of (i) taking a step along $\xi$ in the ambient
space, and (ii) projecting back onto the manifold, defines a valid retraction. In quotient
manifolds, the retractions of the base space that move an \emph{entire} equivalence class to
\emph{another} equivalence class induce retractions on the quotient space.

%%%
\paragraph{Riemannian Connections.}
Let us first briefly introduce and motivate the need for an additional structure attached to
differentiable manifolds, called \emph{affine connections}. They are functions
\begin{equation}
  \nabla : \mathfrak{X} \times \mathfrak{X} \to \mathfrak{X},\enskip (\xi, \zeta) \to \nabla_\xi \zeta
\end{equation}
where $\mathfrak{X}$ is the set of vector fields on $\mathcal{M}$, i.e., functions assigning to each
point $x \in \mathcal{M}$ a tangent vector $\xi_x \in T_x \mathcal{M}$. They satisfy several
properties that represent the generalization of the \emph{directional derivative of a vector field}
in Euclidean space. Affine connections are needed, for instance, to generalize second-order
optimization algorithms, such as Newton's method, to functions defined on manifolds.

The \emph{Riemannian connection}, also known as the Levi-Civita connection, is the \emph{unique}
affine connection that, besides the properties referred to above, satisfies two others, one of which
depends on the Riemannian metric -- see~\cite{carmo1992riemannian} for details. It is the affine
connection implicitly assumed when working with Riemannian manifolds.

%%%
\paragraph{Geodesics, Exponential Map, Logarithm Map, Parallel Transport.}
They are all concepts from Riemannian geometry, defined in terms of the Riemannian connection. A
\emph{geodesic} is a curve with zero acceleration,
\begin{equation}
  \nabla_{\dot{\gamma}(t)} \dot{\gamma}(t) = 0.
\end{equation}
Geodesics are the generalization of straight lines from Euclidean space. They are locally
distance-minimizing and parameterized by arc-length.  Thus, for every $\xi \in T_x \mathcal{M}$,
there exists a unique geodesic $\gamma(t; x, \xi)$ such that $\gamma(0) = x$ and $\dot{\gamma}(0) =
\xi$.

The \emph{exponential map} is the function
\begin{equation}
  \exp_x : \widehat{U} \subseteq T_x \mathcal{M} \to \mathcal{M},\enskip \xi \mapsto \exp_x(\xi) \deq \gamma(1; x, \xi),
\end{equation}
where $\widehat{U}$ is a neighborhood of $0$. The manifold is said to be \emph{geodesically
complete} if the exponential map is defined on the entire tangent space, i.e., $\widehat{U} = T_x
\mathcal{M}$. It can be shown that $\exp_x(\cdot)$ is a retraction and satisfies the following
useful property
\begin{equation}
  d_{\mathcal{M}}\big(x,\ \exp_x(\xi)\big) = \norm*{\xi}_x,\enskip\text{for all}\ \xi \in \widehat{U}.
\end{equation}

The exponential map defines a diffeomorphism between a neighborhood of $0 \in T_x \mathcal{M}$ onto
a neighborhood of $x \in \mathcal{M}$. If we follow a geodesic $\gamma(t; x, \xi)$ from $t = 0$ to
infinity, it can happen that it is minimizing only up to $t_0 < \infty$. If that is the case, the
point $y = \gamma(t_0; x, \xi)$ is called a cut point. The set of all such points, gathered across
all geodesics starting at $x$, is called the \emph{cut locus} of $\mathcal{M}$ at $x$,
$\mathcal{C}(x) \subset \mathcal{M}$. It can be proven that the cut locus has finite
measure~\cite{pennec2006intrinsic}. The maximal domain where the exponential map is a diffeomorphism
is given by its preimage on $\mathcal{M} \setminus \mathcal{C}(x)$. Hence, the inverse is called the
\emph{logarithm map},
\begin{equation}
  \log_x : \mathcal{M} \setminus \mathcal{C}(x) \to T_x \mathcal{M}.
\end{equation}

The \emph{parallel transport} of a vector $\xi_x \in T_x \mathcal{M}$ along a curve $\gamma : I \to
\mathcal{M}$ is the unique vector field $\xi$ that points along the curve to tangent vectors,
satisfying
\begin{equation}
  \nabla_{\dot{\gamma}(t)} \xi\big(\gamma(t)\big) = 0\enskip\text{and}\enskip \xi(x) = \xi_x.
\end{equation}

%%%
\paragraph{Curvature.}
The \emph{Riemann curvature tensor} is a tensor field that assigns a tensor to each point of a
Riemannian manifold. Each such tensor measures the extent to which the manifold \emph{is not}
locally isometric to Euclidean space. It is defined in terms of the Levi-Civita connection. For each
pair of tangent vectors $u, v \in T_x \mathcal{M}$, $\Riem_x(u, v)$ is a linear transformation on
the tangent space. The vector $w' = \Riem_x(u, v)\, w$ quantifies the failure of parallel transport
to bring $w$ back to its original position when following a quadrilateral determined by $-tu, -tv,
tu, tv$, with $t \to 0$. This failure is caused by curvature and it is also known as the
\emph{infinitesimal non-holonomy} of the manifold.

The \emph{sectional curvature} is defined for a fixed point $x \in \mathcal{M}$ and two tangent
vectors $u, v \in T_x \mathcal{M}$ as
\begin{equation}\label{eq:seccurv}
  K_x(u, v) = \frac{\langle \Riem_x(u, v) v, u \rangle_x}{\langle u, u \rangle_x \langle v, v \rangle_x - \langle u, v \rangle_x^2}.
\end{equation}
It measures how far apart two geodesics emanating from $x$ diverge. If it is positive, the two
geodesics will eventually converge. It is the most common curvature characterization that we use to
compare, from a theoretical perspective, the manifolds discussed in this work.

The \emph{Ricci tensor} $\Ric(w, v)$ is defined as the trace of the linear map $T_x \mathcal{M} \to
T_x \mathcal{M}$ given by $u \mapsto \Riem(u, v) w$. It is fully determined by specifying the
scalars $\Ric(u, u)$ for all unit vectors $u \in T_x \mathcal{M}$, which is known simply as
\emph{the Ricci curvature}.  It is equal to the average sectional curvature across all planes
containing $u$ times $(n - 1)$.  Intuitively, it measures how the volume of a geodesic cone in
direction $u$ compares to that of an Euclidean cone.

Finally, the \emph{scalar curvature} (or Ricci scalar) is the most coarse-grained notion of
curvature at a point on a Riemannian manifold. It is the trace of the Ricci tensor, or,
equivalently, $n (n - 1)$ times the average of all sectional curvatures. Note that a space of
non-constant sectional curvature can have constant Ricci scalar. This is true, in particular, for
homogeneous spaces (e.g., all manifolds studied in this work; see~\Cref{sec:manifolds-details}).

% \clearpage
\section{Differential Geometry Tools for Hyperbolic and Elliptical Spaces}\label{sec:hyp-sph}

\begin{table}
  \caption{
    Summary of Riemannian geometric tools for hyperbolic and spherical spaces.
    Notation:
      $x,y \in \mathcal{M}$;
      $u, v \in T_x \mathcal{M}$;
      $\langle x, y \rangle_L$ -- Lorentz product.
  }
  \label{tab:hyp-sph-tools}

  \small
  \centering
  \renewcommand{\arraystretch}{0.4}

  \begin{tabular}{@{}llll@{}}
    \toprule
    \thead{\bfseries Property} & \thead{\bfseries Expr.} & \thead{\bfseries Hyperbolic $\Hyp(n)$} &
      \thead{\bfseries Elliptical $\Sph(n)$} \\
    \midrule

    \\
    Representation & $\Hyp(n)$ / $\Sph(n)$ &
    $\{ x \in \R^{n + 1} : \langle x, x \rangle_L = -1, x_0 > 0 \}$ &
    $\{ x \in \R^{n + 1} : \norm*{x}_2 = 1 \}$
    \\

    \\
    Tangent space & $T_x \mathcal{M}$ &
    $\{ u \in \R^{n + 1} : \langle u, x \rangle_L = 0 \}$ &
    $\{ u \in \R^{n + 1} : x^\top u = 0 \}$
    \\

    \\
    Projection & $\pi_x(u)$ &
    $u + \langle u, x \rangle_L x$ &
    $\big(\eye_{n + 1} - x x^\top\big) u$
    \\

    \\
    Riem.\ metric & $\langle u, v \rangle_x$ &
    $\langle u, v \rangle_L$ &
    $\langle u, v \rangle$
    \\

    \\
    Riem.\ gradient & $\nabla_A^R$ &
    $\diag(-1,1,\ldots,1) \pi_x(\nabla_x^E)$ &
    $\pi_x(\nabla_x^E)$
    \\

    \\
    Geodesics & $\gamma_{x;y}(t)$ &
    $x \cosh(t) + y \sinh(t)$ &
    $x \cos(t) + y \cos(t)$
    \\

    \\
    Retraction & $R_x(u)$ &
    Not used &
    Not used
    \\

    \\
    Log map & $\log_x(y)$ &
    \makecell[l]{
      $\frac{\cosh^{-1}(\alpha)}{\sqrt{\alpha^2 - 1}} (y - \alpha x)$ \\
      \quad with $\alpha = -\langle x, y \rangle_L$
    } &
    \makecell[l]{
      $\log_x(y) = \frac{\cos^{-1}(\langle x, y \rangle)}{\norm{u'}_x} u'$ \\
      \quad with $u' = \pi_x(y - x)$
    }
    \\

    \\
    Riem.\ distance & $d(x, y)$ &
    $\cosh^{-1}(-\langle x, y \rangle_L)$ &
    $\cos^{-1}(\langle x, y \rangle)$
    \\

    \\
    Parallel transport & $\mathcal{T}_{x,y}(u)$ &
    \makecell[l]{
      $u + \frac{\langle y - \alpha x, u \rangle_L}{\alpha + 1} (x + y)$ \\
      \quad with $\alpha = -\langle x, y \rangle_L$
    } &
    $\pi_b(u)$
    \\

    \\
    Characterizations &&
    \makecell[l]{Constant negative curvature\\ Isotropic} &
    \makecell[l]{Constant positive curvature\\ Isotropic}
    \\

    \bottomrule
  \end{tabular}
\end{table}

\begin{figure}
  \centering
  \subfloat[Hyperboloid]{{\includegraphics[scale=0.3]{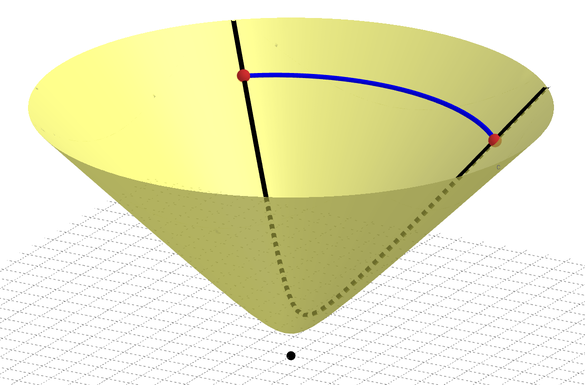}}}
  \hspace*{1cm}
  \subfloat[Sphere]{{\includegraphics[scale=0.3]{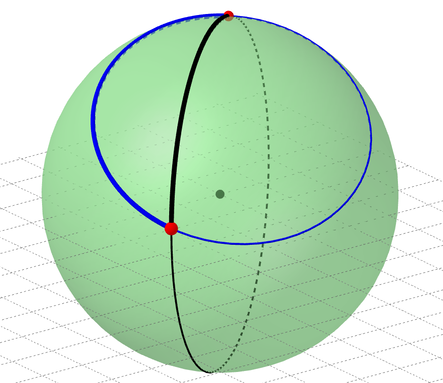} }}

  \caption{The hyperboloid model of hyperbolic geometry (left) and the spherical model of elliptical
  geometry (right). The black curves are geodesics between two points while the blue ones are
  arbitrary paths connecting them (not geodesics). Note that the ambient space of the hyperboloid is
  the Minkowski space, hence our Euclidean intuition does not apply, as it does for the sphere.}
  \label{fig:hyp-sph}
\end{figure}

We include in~\Cref{tab:hyp-sph-tools} the differential geometry tools for the hyperbolic and
elliptical spaces. We use the \emph{hyperboloid model} for the former and the \emph{hyperspherical
model} for the latter, depicted in~\Cref{fig:hyp-sph}. Some prior work prefers working with the
Poincar\'e ball model and/or the stereographic projection. They have both advantages and
disadvantages.

For instance, our choice yields simple formulas for certain quantities of interest, such as
exponential and logarithm maps. They are also more numerically stable. In fact, it is claimed
in~\cite{nickel2018learning} that numerical stability together with its impact on optimization are
the only explanations for the Lorentz model outperforming the prior experiments in the
(theoretically equivalent) Poincar\'e ball~\cite{nickel2017poincare}.

On the other hand, the just mentioned alternative models have the strong advantage that the
corresponding metric tensors are \emph{conformal}. This means that they are proportional to the
Riemannian metric of Euclidean space,
\begin{equation}
  g_{\Hyp} = \bigg(\frac{2}{1 - \norm{x}_2^2}\bigg)^2 g^E
  \enskip\text{and}\enskip
  g_{\Sph} = \bigg(\frac{2}{1 + \norm{x}_2^2}\bigg)^2 g^E.
\end{equation}
Notice the syntactic similarity between them. Furthermore, the effect of the denominators in the
\emph{conformal factors} reinforce the intuition we have about the two spaces: distances around far
away points are \emph{increasingly larger} in the hyperbolic space and \emph{increasingly smaller}
in the elliptical space.

Let us point out that both hyperbolic and elliptical spaces are \emph{isotropic}.  Informally,
isotropy means ``uniformity in all directions.'' Note that this is a stronger property than the
homogeneity of the matrix manifolds discussed in~\Cref{sec:manifolds-details} which means that the
space ``looks the same around each point.''

% \clearpage
\section{Geometric Properties of Graphs}\label{sec:geom-prop-graphs}
Graphs and manifolds, while different mathematical abstractions, share many similar properties
through Laplace operators, heat kernels, and random walks. Another example is the deep connection
between trees and the hyperbolic plane: any tree can be embedded in $\Hyp(2)$ with arbitrarily small
distortion~\cite{sarkar2011low}. On a similar note, complex networks arise naturally from hyperbolic
geometry~\cite{krioukov2010hyperbolic}. With these insights in mind, in this section we review some
continuous geometric properties that have been adapted to arbitrary weighted graphs. See
also~\cite{ni2015ricci}.

%%%
\paragraph{Gromov Hyperbolicity.}
Also known as $\delta$-hyperbolicity \cite{gromov1987hyperbolic}, it quantifies with a single number
the hyperbolicity of a given metric space: the smaller $\delta$ is, the more hyperbolic-like or
negatively-curved the space is. The definition that makes it easier to picture it is via the
\emph{slim triangles property}: a metric space\footnote{Recall that a Riemannian manifold with its
induced distance function is a metric space only if it is connected.} $(M, d_M)$ is
$\delta$-hyperbolic if all geodesic triangles are $\delta$-slim. Three points $x, y, w \in M$ form a
$\delta$-slim triangle if any point on the geodesic segment between any two of them is within
distance $\delta$ from the other two geodesics (i.e., ``sides'' of the geodesic triangle).

For discrete metric spaces such as graphs, an equivalent definition using the so-called ``4-points
condition'' can be used to devise algorithms that look at quadruples of points. Both exact and
approximating algorithms exist that run fast enough to analyze graphs with tens of thousands of
nodes within minutes~\cite{fournier2015computing,cohen2015computing}. In practice we look
at histograms of $\delta$ instead of the worst-case value.

%%%
\paragraph{Ollivier-Ricci Curvature.}
Ollivier et al.~\cite{ollivier2009ricci} generalized the \emph{Ricci curvature} to metric spaces
$(M, d_M)$ equipped with a family of probability measures $\{m_x(\cdot)\}_{x \in M}$. It is defined
in a way that mimics the interpretation of Ricci curvature on Riemannian manifolds: it is the
average distance between two small balls taken relative to the distance between their centers. The
difference is that now the former is given by the Wasserstein distance (i.e., Earth mover's
distance) between the corresponding probability measures,
\begin{equation}
  \overline{\Ric}_1(x, y) \deq 1 - \frac{W(m_x, m_y)}{d_M(x, y)},
\end{equation}
with $W(\mu_1,\mu_2) \deq \inf_{\xi} \int_x \int_y d(x, y) d\xi(x, y)$ and $\xi(x, y)$ -- a join
distribution with marginals $\mu_1$ and $\mu_2$. This definition was then specialized
in~\cite{lin2011ricci} for graphs by making $m_x$ assign a probability mass of $\alpha \in [0, 1)$
to itself (node $x$) and spread the remaining $1 - \alpha$ uniformly across its neighbors. They
refer to $\overline{\Ric}_{\alpha}(x, y) \deq \overline{\Ric}_1(x, y)$ as the $\alpha$-Ricci
curvature and define instead
\begin{equation}
  \overline{\Ric}_2(x, y) \deq \lim_{\alpha \to 1} \frac{\overline{\Ric}_{\alpha}(x, y)}{1 - \alpha}
\end{equation}
to be \emph{the Ricci curvature of edge $(x, y)$ in the graph}. In practice, we approximate the
limit via a large $\alpha$, e.g., $\alpha = 0.999$.

Notice that in contrast to $\delta$-hyperbolicity, the Ricci curvature characterizes the space only
locally. It yields the curvatures one would expect in several cases: negative curvatures for trees
(except for the edges connecting the leaves) and positive for complete graphs and hypercubes; but it
does not capture the curvature of a cycle with more than 5 nodes because it locally looks like a
straight line.

\begin{figure}[t]
  \centering
  \includegraphics[scale=0.22]{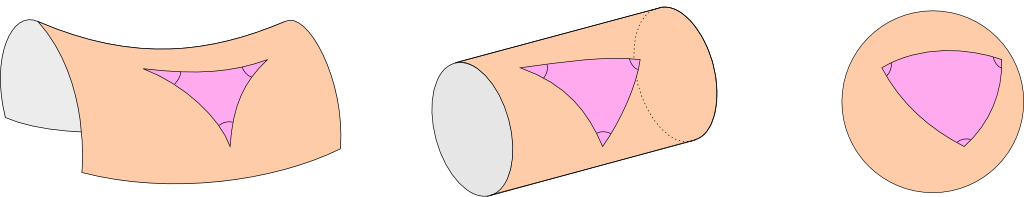}
  \caption{Geodesic triangles in negatively curved, flat, and positively curved spaces,
  respectively. Source: \url{www.science4all.org}}
  \label{fig:curvatures}
\end{figure}

%%%
\paragraph{Sectional Curvatures.}
A discrete analogue of sectional curvature for graphs is obtained as the \emph{deviation from the
parallelogram law} in Euclidean geometry~\cite{gu2018learning}. It uses the same intuition as
before: in non-positively curved spaces triangles are ``slimmer'' while in non-negatively curved
ones they are ``thicker'' (see~\Cref{fig:curvatures}).

For any Riemannian manifold $\mathcal{M}$, let $x, y, z \in \mathcal{M}$ form a geodesic triangle,
and let $m$ be the midpoint of the geodesic between $y$ and $z$. Then, the following quantity
\begin{equation}\label{eq:dev-from-par-law}
  k_{\mathcal{M}}(x, y, z) \deq d_{\mathcal{M}}(x, m)^2 + d_{\mathcal{M}}(y, z)^2 / 4 
    -\big(d_{\mathcal{M}}(x, y)^2 + d_{\mathcal{M}}(x, z)^2\big)/2
\end{equation}
has the same sign as the sectional curvatures in $\mathcal{M}$ and it is identically zero if
$\mathcal{M}$ is flat. For a graph $G = (V, E)$, an analogue is defined for each node $m$ and two
neighbors $y$ and $z$, as
\begin{equation}\label{eq:graph-seccurv-init}
  k_G(m; y, z; x) = \frac{1}{2 d_G(x, m)} k_G(x, y, z).
\end{equation}
With that, the sectional curvature of the graph $G$ at $m$ in ``directions'' $y$ and $z$ is defined as
the average of~\eqref{eq:graph-seccurv-init} across all $x \in G$,
\begin{equation}
  k_G(m; y, z) = \frac{1}{\abs{V} - 1} \sum_{x \ne m} k_G(m; y, z; x).
\end{equation}
It is also shown in~\citep[][AppxC.2]{gu2018learning} that this definition recovers the expected
signs for trees (negative or zero), cycles (positive or zero) and lines (zero).

% \clearpage
\section{Matrix Manifolds -- Details}\label{sec:manifolds-details}
In this section, we include several results that are useful for better understanding and working
with the matrix manifolds introduced in~\Cref{sec:matrix-manifolds}. They have been left out from
the main text due to space constraints. Furthermore, we describe the orthogonal group $\Ortho(n)$
which is used in our random manifold graphs analysis from~\Cref{sec:man-rand-graphs}. We do not use
it for graph embedding purposes in this work because its corresponding distance function does not
immediately lend itself to simple backpropagation-based training (in general). This is left for
future work.

\paragraph{Homogeneity.}
It is a common property of the matrix manifolds used in this work. Formally, it means that the
isometry group of $\mathcal{M}$ acts transitively: for any $A, B \in \mathcal{M}$ there is an
isometry that maps $A$ to $B$. In non-technical terms this says that $\mathcal{M}$ ``looks locally
the same'' around each point. A consequence of homogeneity is that in order to prove curvature
properties, it suffices to do so at a single point, e.g., the identity matrix for $\Spd(n)$.

\subsection{SPD Manifold}
The following theorem, proved here for completeness, states that $\Spd(n)$ is a differentiable
manifold.
\begin{theorem}
  The set $\Spd(n)$ of symmetric positive-definite matrices is an $\frac{n (n + 1)}{2}$-dimensional
  differentiable manifold.
\end{theorem}
\begin{proof}
  The set $\Sym(n)$ is an $\frac{n (n + 1)}{2}$-dimensional vector space. Any finite dimensional
  vector space is a differentiable manifold: fix a basis and use it as a global chart mapping points
  to the Euclidean space with the same dimension.

  The set $\Spd(n)$ is an open subset of $\Sym(n)$. This follows from $(A, v) \mapsto v^\top A v$
  being a continuous function. The fact that open subsets of (differentiable) manifolds are
  (differentiable) manifolds concludes the proof.
\end{proof}

The following result from linear algebra makes it easier to compute geodesic distances.
\begin{lemma}\label{th:same-eigs}
  Let $A, B \in \Spd(n)$. Then $A B$ and $A^{1/2} B A^{1/2}$ have the same eigenvalues.
\end{lemma}
\begin{proof}
  Let $A = A^{1/2} A^{1/2}$, where $A^{1/2}$ is the unique square root of $A$. Note that $A^{1/2}$
  is itself symmetric and positive-definite because every analytic function $f(A)$ is equivalent to
  $U \diag(\{f(\lambda_i)\}_i) U^\top$, where $A = U \diag(\{\lambda_i\}_i) U^\top$ is the
  eigenvalue decomposition of $A$, and $\lambda_i > 0$ for all $i$ from positive-definiteness. Then,
  we have
  \[
    A B = A^{1/2} A^{1/2} B = A^{1/2} (A^{1/2} B A^{1/2}) A^{-1/2},
  \]
  Therefore, $A B$ and $A^{1/2} B A^{1/2}$ are similar matrices so they have the same eigenvalues.
\end{proof}
This is useful from a computational perspective because $A^{-1/2} B A^{-1/2}$ is an SPD matrix while
$A^{-1} B$ may not even be symmetric. It also makes it easier to see the following equivalence, for
any matrix $A \in \Spd(n)$:
\begin{equation}
  \norm{\log(A)}_F = \norm*{U \diag\Big(\log(\lambda_i(A))\Big) U^T}_F
                   = \norm*{\diag\Big(\log(\lambda_i(A))\Big)}_F
                   = \sqrt{\sum_{i=1}^n \log^2\big(\lambda_i(A)\big)},
\end{equation}
where the first step decomposes the \emph{principle logarithm} of $A$ and the second one uses the
change of basis invariance of the Frobenius norm.

An even better behaved matrix that has the same spectrum and avoids matrix square roots altogether
is $L B L^\top$, where $L L^\top$ is the Cholesky decomposition of $A$. Note that for the Stein
divergence, the log-determinants can be computed in terms of $L$ as $\log \det(A) = 2 \sum_{i=1}^n
\log(L_{ii})$.

An additional challenge with eigenvalue computations is that most linear algebra libraries are
optimized for large matrices while our use-case involves milions of very small matrices.\footnote{To
illustrate, support for batched linear algebra operations in PyTorch is still work in progress at
the time of this writing (June 2020): \url{https://github.com/pytorch/pytorch/issues/7500}.} To
overcome that, for $2 \times 2$ and $3 \times\ 3$ matrices we use custom formulas that can be
derived explicitly:
\begin{itemize}
  \item For $A \in \Spd(2)$, we have
    \begin{equation}
      \lambda_k(A) = \frac{t}{2} \pm \sqrt{\bigg(\frac{t}{2}\bigg)^2 - d},
    \end{equation}
  with $t = \Tr A$ and $d = \det(A)$.

  \item For $A \in \Spd(3)$, we express it as an affine transformation of another matrix, i.e., $A =
    p B + q \eye_n$. Then we have $\lambda_k(A) = p \lambda_k(B) + q$. Concretely, if we let
    \[
      q = \frac{\Tr A}{3}
      \enskip\text{and}\enskip
      p = \sqrt{\frac{\Tr\ (A - q\eye_n)^2}{6}},
    \]
    then the eigenvalues of $B$ are
    \begin{equation}
      \lambda_k(B) = 2 \cos\bigg(\frac{1}{3} \arccos\Big(\frac{\det(B)}{2}\Big) + \frac{2 k \pi}{3}\bigg).
    \end{equation}
\end{itemize}

\subsection{Grassmann Manifold}
The following singular value formulas are useful in computing geodesic
distances~\eqref{eq:grass-dist} between points on $\Gr(2, n)$.
\begin{proposition}
  The singular values of $A = \begin{bmatrix}a & b \\ c & d\end{bmatrix}$ are
    \begin{gather*}
      \sigma_1 = \sqrt{\frac{S_1 + S_2}{2}}
      \enskip\text{and}\enskip
      \sigma_2 = \sqrt{\frac{S_1 - S_2}{2}},
    \end{gather*}
    with
    \[
      S_1 = a^2 + b^2 + c^2 + d^2
      \enskip\text{and}\enskip
      S_2 = \sqrt{(a^2 + b^2 - c^2 - d^2)^2 + 4 (ac + bd)^2}.
    \]
\end{proposition}

\subsection{Orthogonal Group}
It is defined as the set of $n \times n$ real orthogonal matrices,
\begin{equation}\label{eq:ortho-def}
  \Ortho(n) \deq \{ A \in \R^{n \times n} : A^\top A = A A^\top = \eye_n \}.
\end{equation}
We are interested in the geometry of $\Ortho(n)$ rather than its group properties. Formally, this
means that in what follows we describe its so-called principal homogeneous space, the special
Stiefel \emph{manifold} $\St(n, n)$, rather than the \emph{group} $\Ortho(n)$.

\begin{theorem}\label{th:ortho-man}
  $\Ortho(n)$ is an $\frac{n (n-1)}{2}$-dimensional differentiable manifold.
\end{theorem}
\begin{proof}
  Consider the map $F : \R^{n \times n} \rightarrow \Sym(n),\ F(A) = A^\top A - \eye_n$. It is clear
  that $\Ortho(n) = F^{-1}(\vec{0})$. The differential at $A$, $\Diff F(A) [B] = A^\top B + B^\top
  A$, is onto the space of symmetric matrices as a function of the direction $B \in \R^{n \times
  n}$, as shown by the fact that $\Diff F(A) \Big[\frac{1}{2} A P\Big] = P$ for all $P \in \Sym(n)$.
  Therefore, $F(\cdot)$ is full rank and by the submersion theorem (\Cref{sec:diff-geom}) the
  orthogonal group is a differentiable (sub-)manifold. Its dimension is $d = \dim(\R^{n \times n}) -
  \dim(\Sym(n)) = n (n - 1) / 2$.
\end{proof}

As a Riemannian manifold, $\Ortho(n)$ inherits the metric structure from the ambient space and thus
the Riemannian metric is simply the Frobenius inner product,
\begin{equation}\label{eq:ortho-metric}
  \langle P, Q \rangle_A = \Tr P^\top Q.
\end{equation}
for $A \in \Ortho(n)$ and $P, Q \in T_A \Ortho(n)$. To derive its tangent space, we use the
following general result.

\begin{proposition}[\cite{absil2009optimization}]
  If $\mathcal{M}$ is an embedded submanifold of a vector space $\mathcal{E}$, defined as a level
  set of a constant-rank function $F : \mathcal{E} \rightarrow \R^n$, we have
  \[
    T_x \mathcal{M} = \Ker (\Diff F(x)).
  \]
\end{proposition}
Differentiating $A^\top A = \eye_n$ yields $\dot{A}^\top A + A^\top \dot{A} = \vec{0}$ so the
tangent space at $A$ is
\begin{equation}
  T_A \Ortho(n) = \{ P \in \R^{n \times n} : A^\top P = X \in \Skew(n) \} = A\, \Skew(n),
\end{equation}
where $\Skew(n)$ is the space of $n \times n$ skew-symmetric matrices.

Many Riemannian quantities of interest are syntactically very similar to those for the canonical SPD
manifold (see~\Cref{tab:diff-geom-summary}). The connection is made precise by the following
property.
\begin{lemma}[\cite{dolcetti2018skew}]\label{th:ortho-spd}
  The metric \eqref{eq:ortho-metric} is the opposite of the affine-invariant
  metric~\eqref{eq:spd-metric} that the SPD manifold is endowed with.
\end{lemma}
\begin{proof}
  Let $A \in \Ortho(n)$ and $P_1, P_2 \in T_A \Ortho(n)$. Then,
  \begin{align*}
    \Tr A^{-1} P_1 A^{-1} P_2 &= \Tr A^\top A X_1 A^\top A X_2 = \Tr X_1 X_2 &&\text{($P_i = A X_i$)} \\
    &= -\Tr X_1^\top X_2 = -\Tr\ (A X_1)^\top (A X_2) &&\text{($X_i \in \Skew(n)$)} \\
    &= -\Tr P_1^\top P_2.
  \end{align*}
\end{proof}

An important characteristic of $\Ortho(n)$ is its compact Lie group structure which implies, via the
Hopf-Rinow theorem, that its components (see below) are geodesically complete: all geodesics $t
\mapsto A \exp(t A^\top P)$ are defined on the whole real line. Note, though, that they are
minimizing only up to some $t_0 \in \R$. This is another consequence of its compactness.

Moreover, $\Ortho(n)$ has two connected components so the expression for the geodesic between two
matrices $A, B \in \Ortho(n)$,
\begin{equation}
  \gamma_{A,B}(t) = A(A^\top B)^t,
\end{equation}
only makes sense if they belong to the same component. The same is true for the logarithm map,
\begin{equation}
  \log_A(B) = A \log(A^\top B).
\end{equation}
The two components contain the orthogonal matrices with determinant $1$ and $-1$, respectively. The
former is the so-called \emph{special orthogonal group},
\begin{equation}
  \So(n) \deq \{ A \in \Ortho(n): \det(A) = 1 \}.
\end{equation}
Restricting to one of them guarantees that the logarithm map is defined on the whole manifold except
for the cut locus (see~\Cref{sec:diff-geom}). This is true in general for compact connected Lie
groups endowed with bi-invariant exponential maps, but we can bring it down to a clearer matrix
property by looking at its expression,
\begin{equation}\label{eq:ortho-dist}
  d(A, B) = \norm{\log(A^\top B)}_F.
\end{equation}
Notice that $A^\top B$ is in $\So(n)$ whenever $A$ and $B$ belong to the same component,
\begin{equation}
  \det(A) = \det(B) = d \in \{-1,1\} \implies \det(A^\top B) = 1.
\end{equation}
Then, the claim is true due the surjectivity of the matrix exponential, as follows.
\begin{proposition}[\cite{cardoso2010exponentials}]
  For any matrix $A \in \So(n)$, there exists a matrix $X \in \Skew(n)$ such that $\exp(X) = A$, or,
  equivalently, $\log(A) = X$. Moreover, if $A$ has no negative eigenvalues, there is a unique such
  matrix with $\mathop{\mathrm{Im}} \lambda_i(X) \in (-\pi, \pi)$, called its \emph{principal
  logarithm}.
\end{proposition}

This property makes it easier to see that the cut locus at a point $A \in \So(n)$ consists of those
matrices $B \in \So(n)$ such that $A^\top B$ has eigenvalues equal to $-1$.\footnote{Such
eigenvalues must come in pairs since $\det(A^\top B) = 1$.} They can be thought of as analogous to
the antipodal points on the sphere. It also implies that the distance
function~\eqref{eq:ortho-dist}, expanded as
\begin{equation}
  d(A, B) = \norm{\log(A^\top B)}_F = \sqrt{\sum_{i=1}^n \Arg\big(\lambda_i(A^\top B)\big)^2},
\end{equation}
is well-defined for points in the same connected components. The $\Arg(\cdot)$ operator denotes the
complex argument. Notice again the similarity to the canonical SPD distance~\eqref{eq:spd-dist}.
However, the nicely behaved symmetric eigenvalue decomposition cannot be used anymore and the
various approximations to the matrix logarithm are too slow for our graph embedding purposes. That
is why we limit our graph reconstruction experiments with compact matrix manifolds to the Grassmann
manifold. Moreover, the small dimensional cases where we could compute the complex argument
``manually'' are isometric to other manifolds that we experiment with: $\Ortho(2) \cong \Sph(1)$ and
$\Ortho(3) \cong \Gr(1,4)$.

\clearpage
\section{Manifold Random Graphs}\label{sec:man-rand-graphs}
The idea of sampling points on the manifolds of interest and constructing nearest-neighbor graphs,
used in prior influential works such as~\cite{krioukov2010hyperbolic}, is motivated, in our case, by
the otherwise black-box nature of matrix embeddings. However, in~\Cref{sec:experiments} we do not
stop at reporting reconstruction metrics, but look into manifold properties around the embeddings.
But even with that, the \emph{discretization} of the embedding spaces is a natural first step in
developing an intuition for them.

The discretization is achieved as follows: several samples ($1000$ in our case) are generated
randomly on the target manifold and a graph is constructed by linking two nodes together
(corresponding to two sample points) when the geodesic distance between them is less than some
threshold. The details of the random generation is discussed in each of the following two
subsections. As mentioned in the introductory paragraph, it is the same approach employed
in~\cite{krioukov2010hyperbolic} who sample from the hyperbolic plane and obtain graphs that
resemble complex networks. In our experiments, we go beyond their technique and study the properties
of the generated graphs when varying the distance threshold.

\subsection{Compact Matrix Manifolds vs.\ Sphere}
We begin the analysis of random manifold graphs with the compact matrix manifolds compared against
the elliptical geometry. The sampling is performed uniformly on each of the compact spaces.

\begin{figure}[t]
  \centering
  \subfloat[node degrees; $n = 2$]{{\includegraphics[scale=0.30]{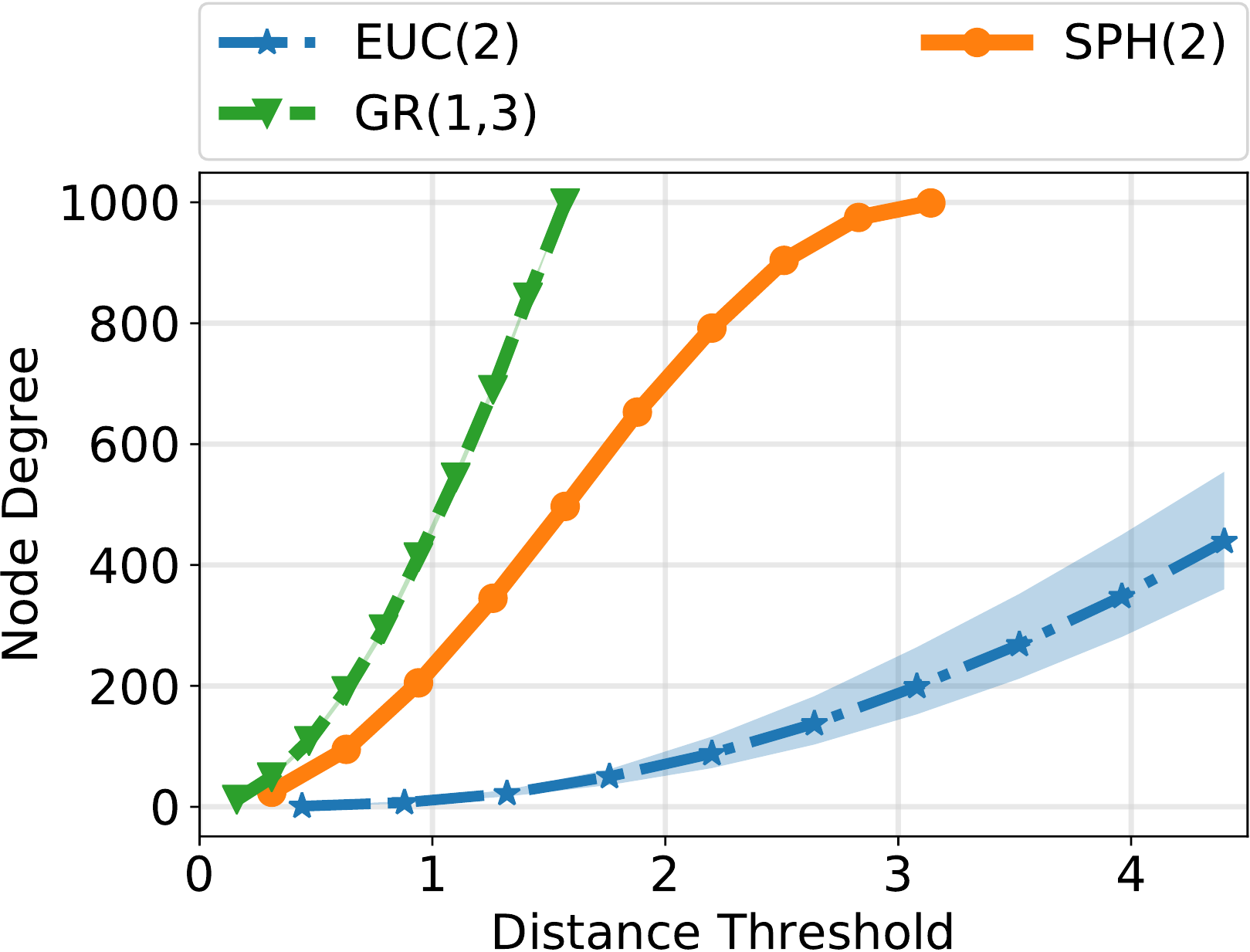} }}
  \subfloat[node degrees; $n = 3$\label{fig:comp-curv-deg3}]{{\includegraphics[scale=0.30]{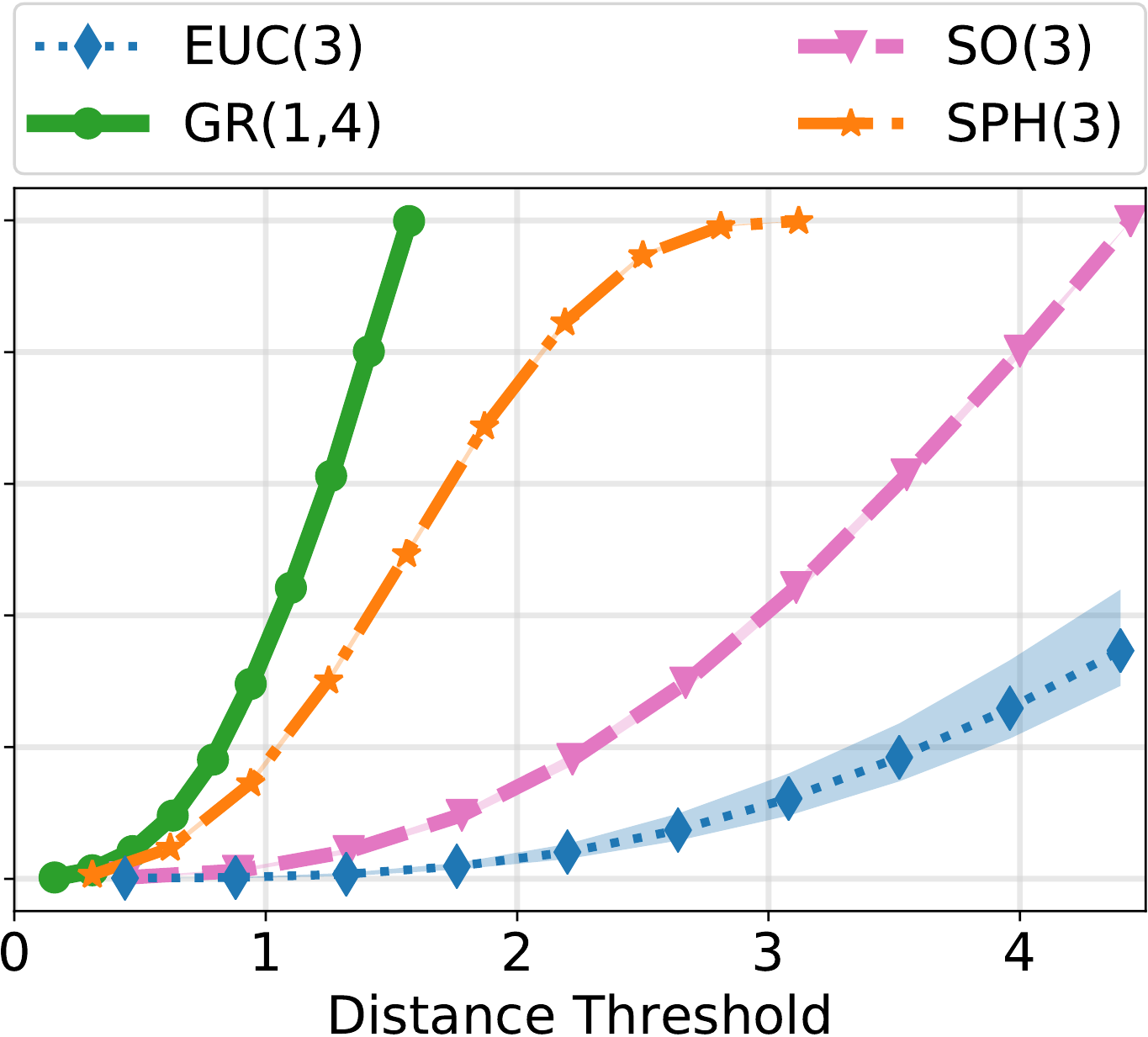} }}
  \subfloat[node degrees; $n = 4$]{{\includegraphics[scale=0.30]{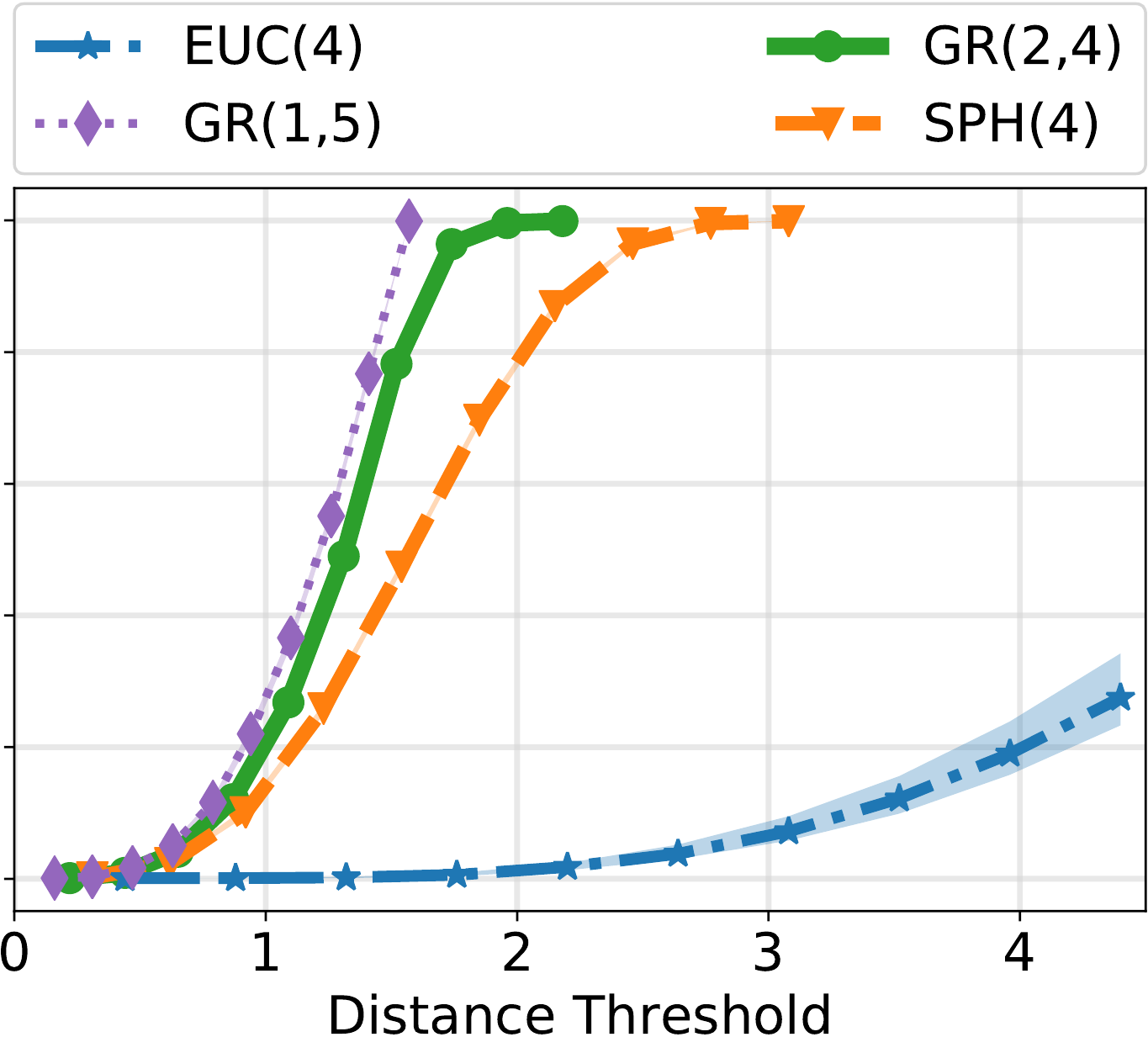} }}

  \subfloat[graph sec.\ curv.; $n = 2$]{{\includegraphics[scale=0.30]{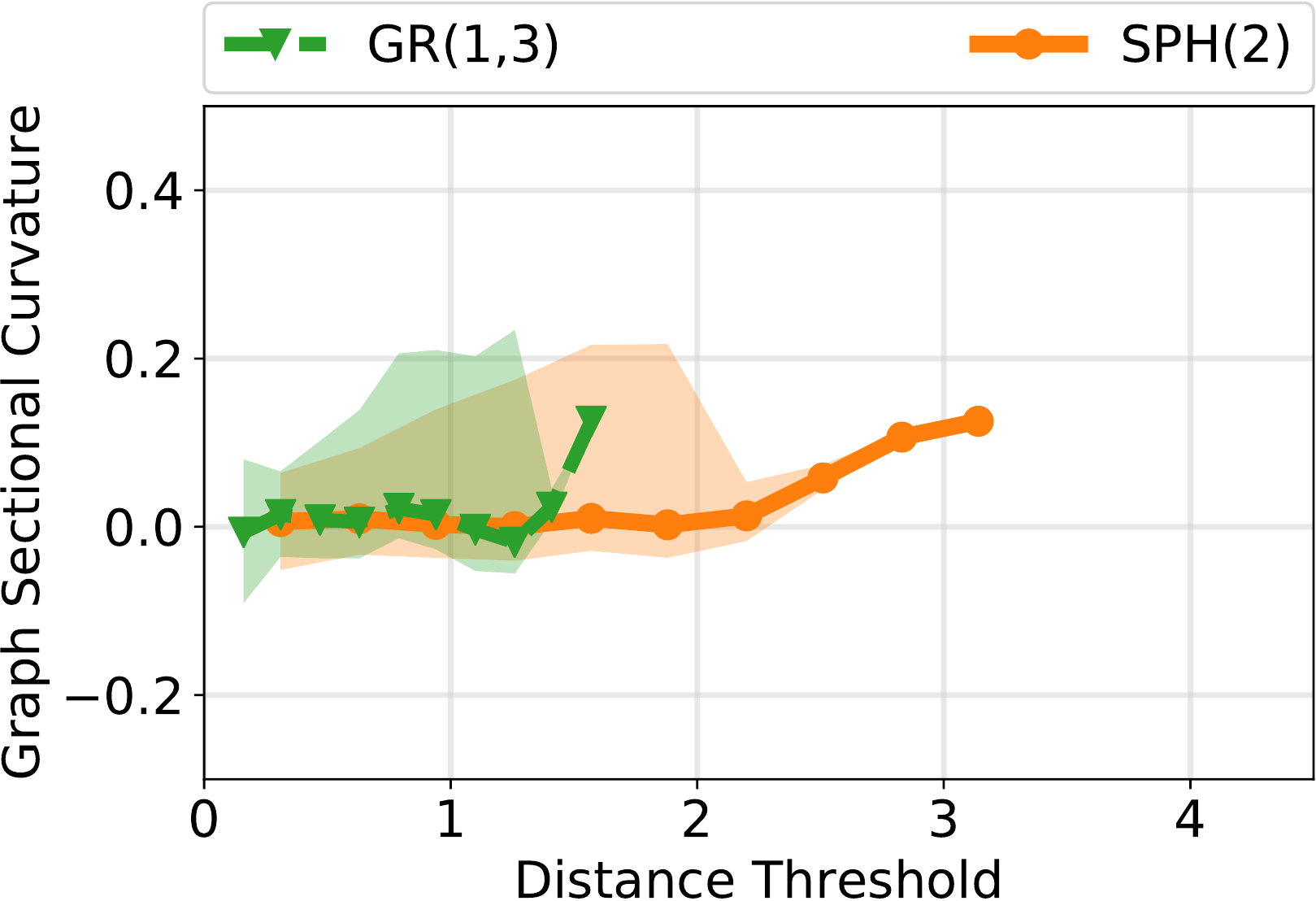} }}
  \subfloat[graph sec.\ curv.; $n = 3$]{{\includegraphics[scale=0.30]{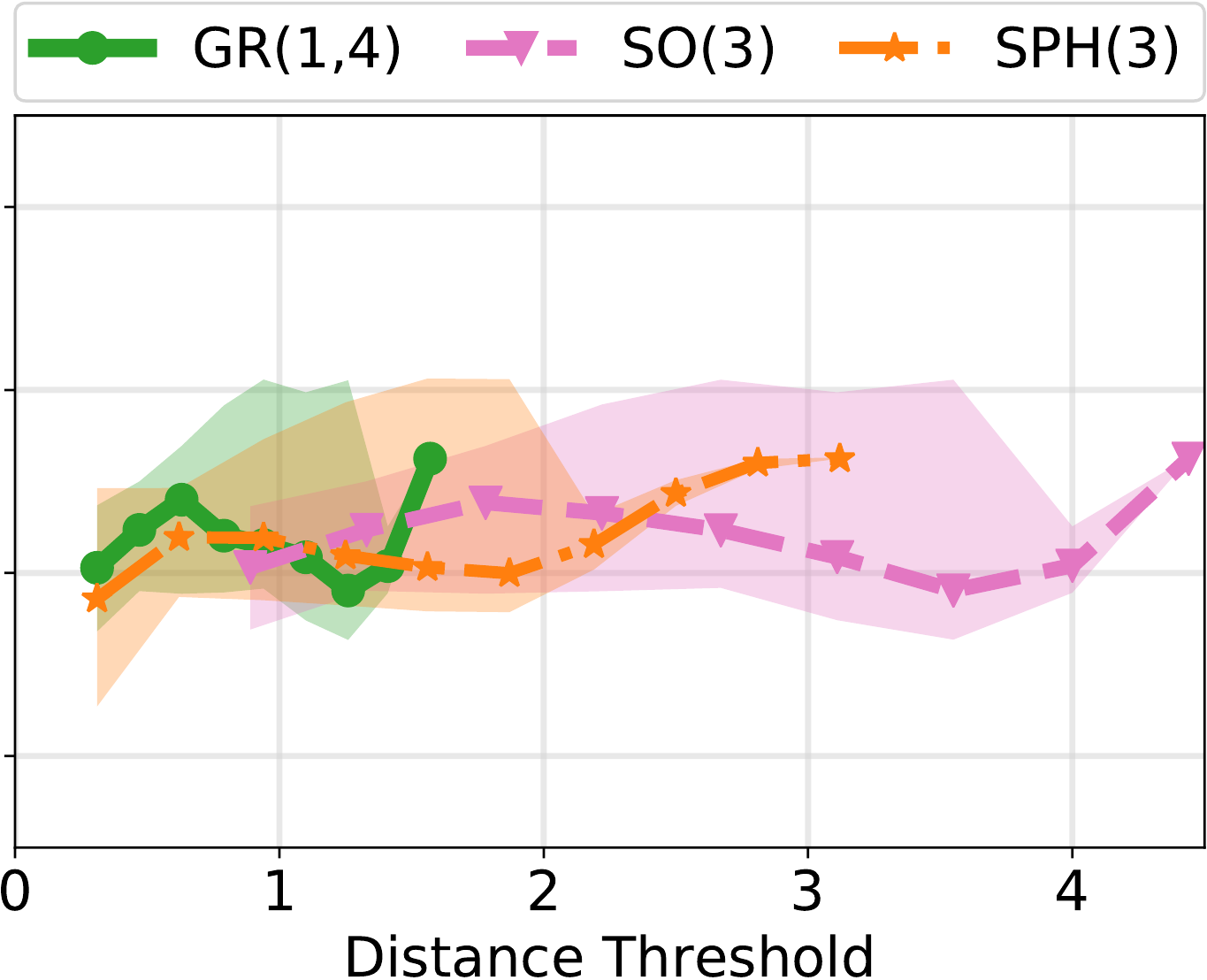} }}
  \subfloat[graph sec.\ curv.; $n = 4$\label{fig:comp-curv-est4}]{{\includegraphics[scale=0.30]{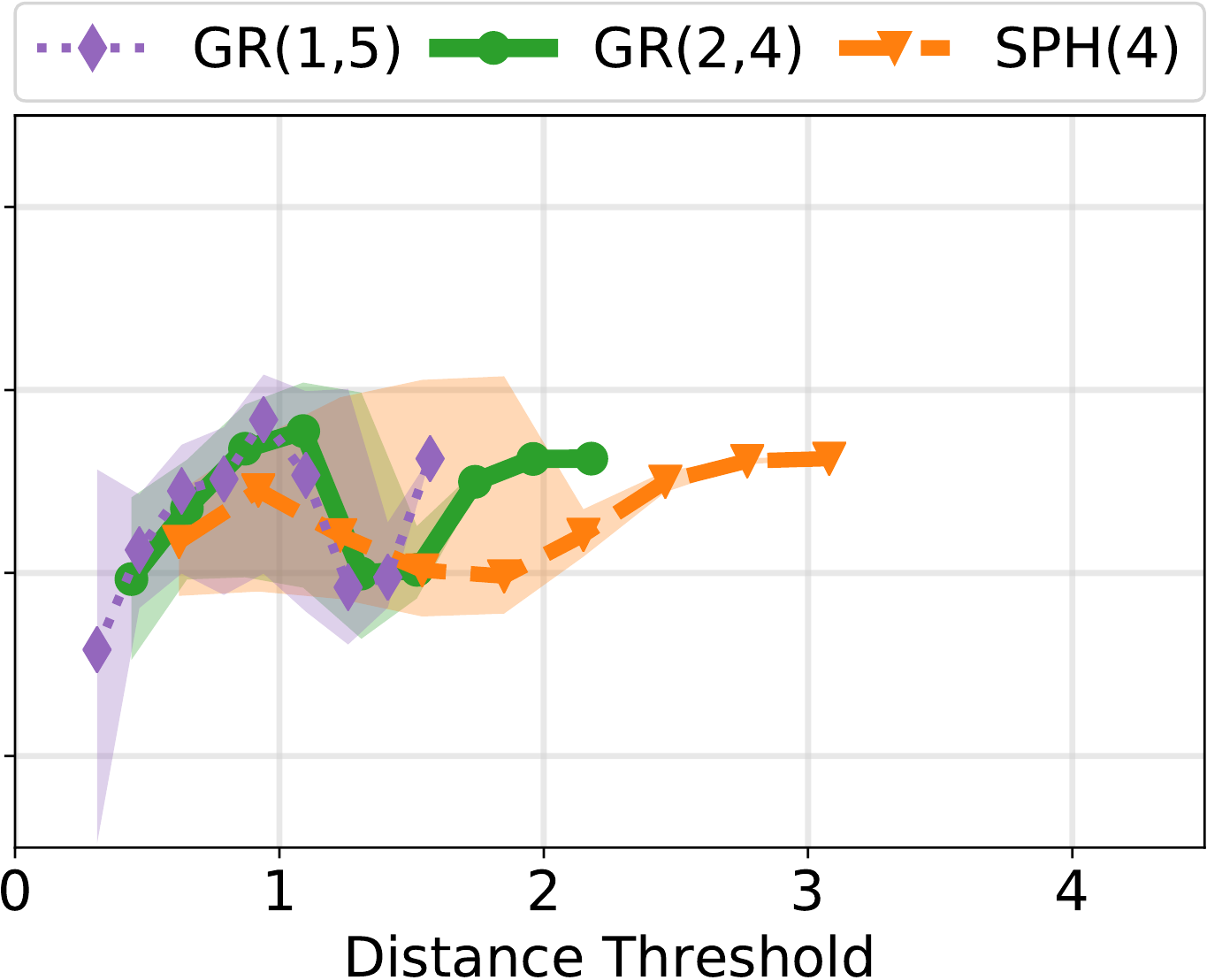} }}

  \subfloat[man.\ angles; $n = 2$]{{\includegraphics[scale=0.30]{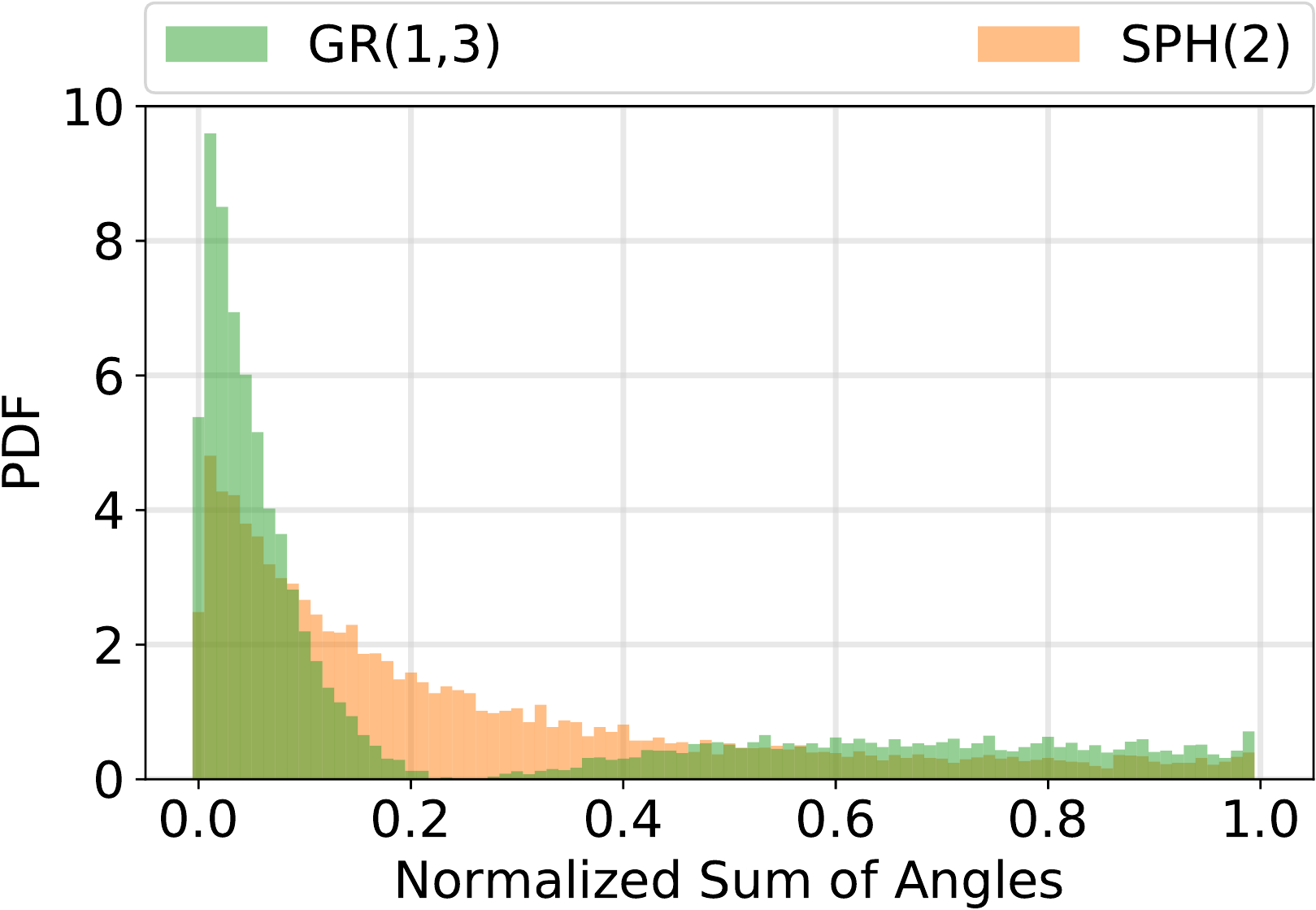} }}
  \subfloat[man.\ angles; $n = 3$\label{fig:comp-curv-ang3}]{{\includegraphics[scale=0.30]{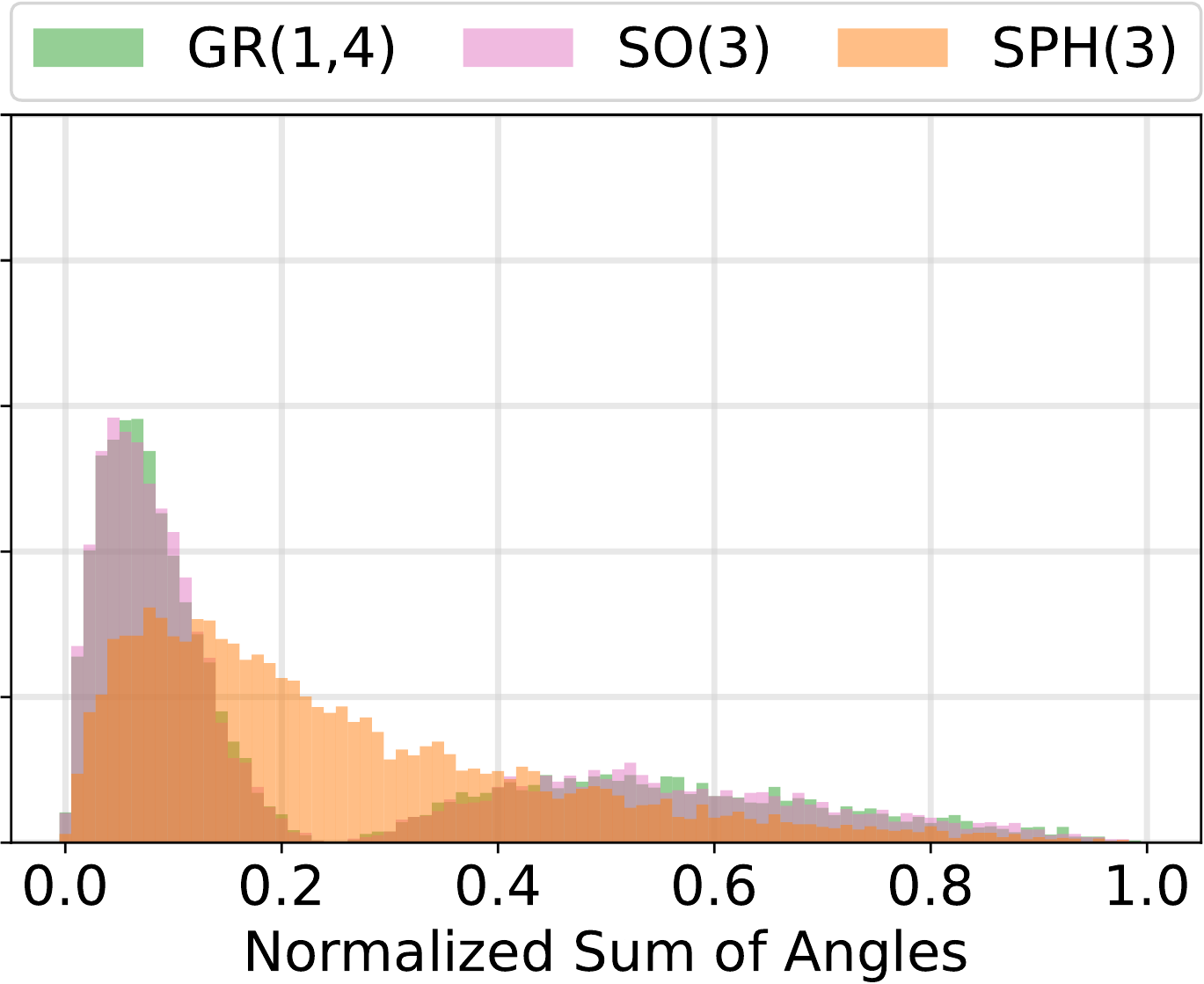} }}
  \subfloat[man.\ angles; $n = 4$]{{\includegraphics[scale=0.30]{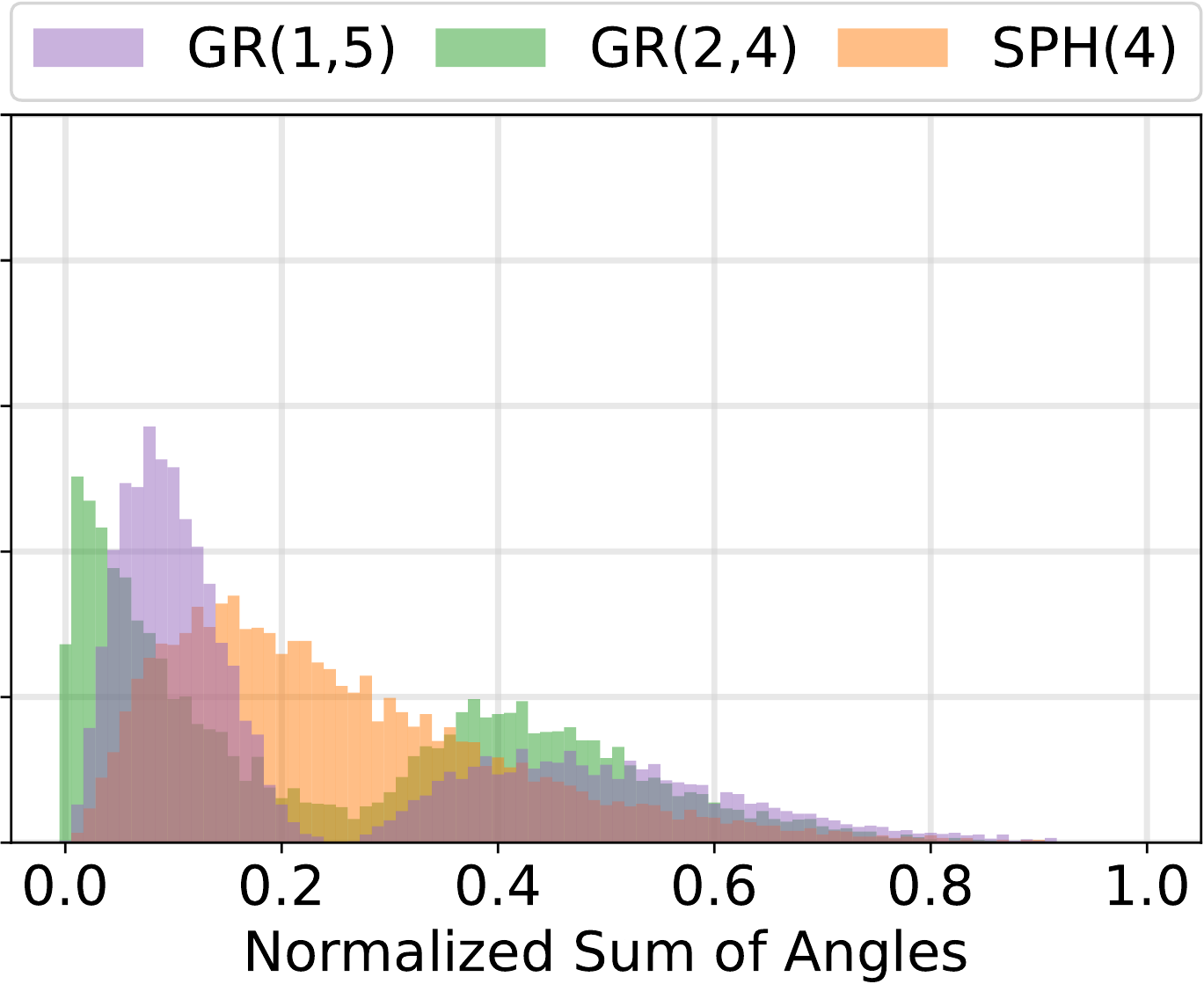} }}

  \caption[Analysis of random graphs sampled from the compact manifolds]{%
    Analysis of graphs sampled from the compact manifolds. The results are grouped by manifold
    dimension (one column each). The first row (a)-(c) shows the node distributions as the distance
    threshold used to link two nodes is varied. Samples from an Euclidean ball of radius $4.4$ are
    included for comparison. The shaded range, hardly noticeable for compact manifolds, is the IQR.
    The second row (d)-(f), with the same $x$-axis as the first one, shows the distributions of
    graph sectional curvatures obtained via the deviation from parallelogram law
    (see~\Cref{sec:geom-prop-graphs}). The third row (g)-(i) shows histograms of normalized angle
    sums obtained from manifold-sampled triangles. It does not depend on the graphs analyzed in the
    previous plots but only on the manifold samples.}
  \label{fig:comp-gen-results}
\end{figure}

The results are shown in~\Cref{fig:comp-gen-results}. The node degrees and the graph sectional
curvatures (computed via the ``deviation from parallelogram law''; see~\Cref{sec:geom-prop-graphs})
are on the first two rows. The markers are the median values and the shaded area corresponds to the
inter-quartile range (IQR). The last row shows normalized sum-of-angles histograms.  All of them are
repeated for three consecutive dimensions, organized column-wise. The distance thresholds used to
link nodes in the graph range from $d_{\textrm{max}} / 10$ to $d_{\textrm{max}}$, where
$d_{\textrm{max}}$ is the maximum distance between any two sample points, for each instance except
for the Euclidean baseline which uses the maximum distance across all manifolds (i.e., $\approx
4.4$ corresponding to $\So(3)$).

%%%
\paragraph{Degree Distributions.} From the degree distributions, we first notice that all compact
manifolds lead to graphs with higher degrees for the same distance threshold than the ones based on
uniform samples from Euclidean balls. This is not surprising given that points tend to be closer due
to positive curvature. Furthermore, the distributions are concentrated around the median -- the IQRs
are hardly visible. To get more manifold-specific, we see that Grassmannians lead to full cliques
faster than the spherical geometry, but the curves get closer and closer as the dimension is
increased. This is the same behavior that in Euclidean space is known as the ``curse of
dimensionality'', i.e., a small threshold change takes us from a disconnected graph to a
fully-connected one. For the compact manifolds, though, this can be noticed already at a very small
dimension: the degree distributions tend towards a step function. The only difference is the point
where that happens, which depends on the maximum distance on each manifold. For instance, that is
much earlier on the real projective space $\RP(n-1) \cong \Gr(1, n)$ than the special orthogonal
group $\So(n)$.

%%%
\paragraph{Graph Sectional Curvatures.} They confirm the similarity between the compact manifolds
that we have just hinted at: for each of the three dimensions, the curves look almost identical up
to a frequency change (i.e., ``stretching them''). We say ``almost'' because we can still see
certain differences between them (explained by the different geometries); for instance,
in~\Cref{fig:comp-curv-est4}, the graph curvatures corresponding to Grassmann manifolds are slightly
larger at distance threshold $\approx 1$. This frequency change seems to be intimately related to
the injectivity radii of the manifolds~\citep[see, e.g.,][]{tron2011average}. We also see that
distributions are mostly positive, matching their continuous analogues. A-priori, it is unclear if
manifold discretization will preserve them. Finally, the convergence point is $k_G(m; x, y) = 1/8$
-- the constant sectional curvature of a complete graph
(see~\cref{eq:dev-from-par-law,eq:graph-seccurv-init}).

%%%
\paragraph{Triangles Thickness.} The normalized sum-of-angles plots do not depend on the generated
graphs: the geodesic triangles are randomly selected from the manifold-sampled points.  As a sanity
check, we first point out that they are all positive.\footnote{To make it clear, note that in
contrast to the discrete sectional curvatures of random nearest-neighbor graphs, this is a property
of the manifold.} We observe that the Grassmann samples yield empirical distributions that look
bi-modal. At the same time, the elliptical ones result in normalized sum-of-angles that resemble
Poisson distributions, with the dimension playing the role of the parameter $\lambda$. We could not
justify these contrasting behaviors, but they show that the spaces curve differently. What we
\emph{can} justify is the perfect overlap of the distributions corresponding to $\So(3)$ and
$\Gr(1,4)$ in~\Cref{fig:comp-curv-ang3}: the two manifolds are isometric. The seemingly different
degree distributions from~\Cref{fig:comp-curv-deg3} should, in fact, be identical after a rescaling.
In other words, they have different volumes but curve in the same way.

\subsection{SPD vs.\ Hyperbolic vs.\ Euclidean}\label{sec:noncomp-gen}
In this subsection, using the same framework as for compact manifolds, we compare the non-positively
curved manifold of positive-definite matrices to hyperbolic and Euclidean spaces.

\begin{figure}
  \centering
  \subfloat[$\Hyp(3)$ -- uniform]{{\includegraphics[scale=0.25]{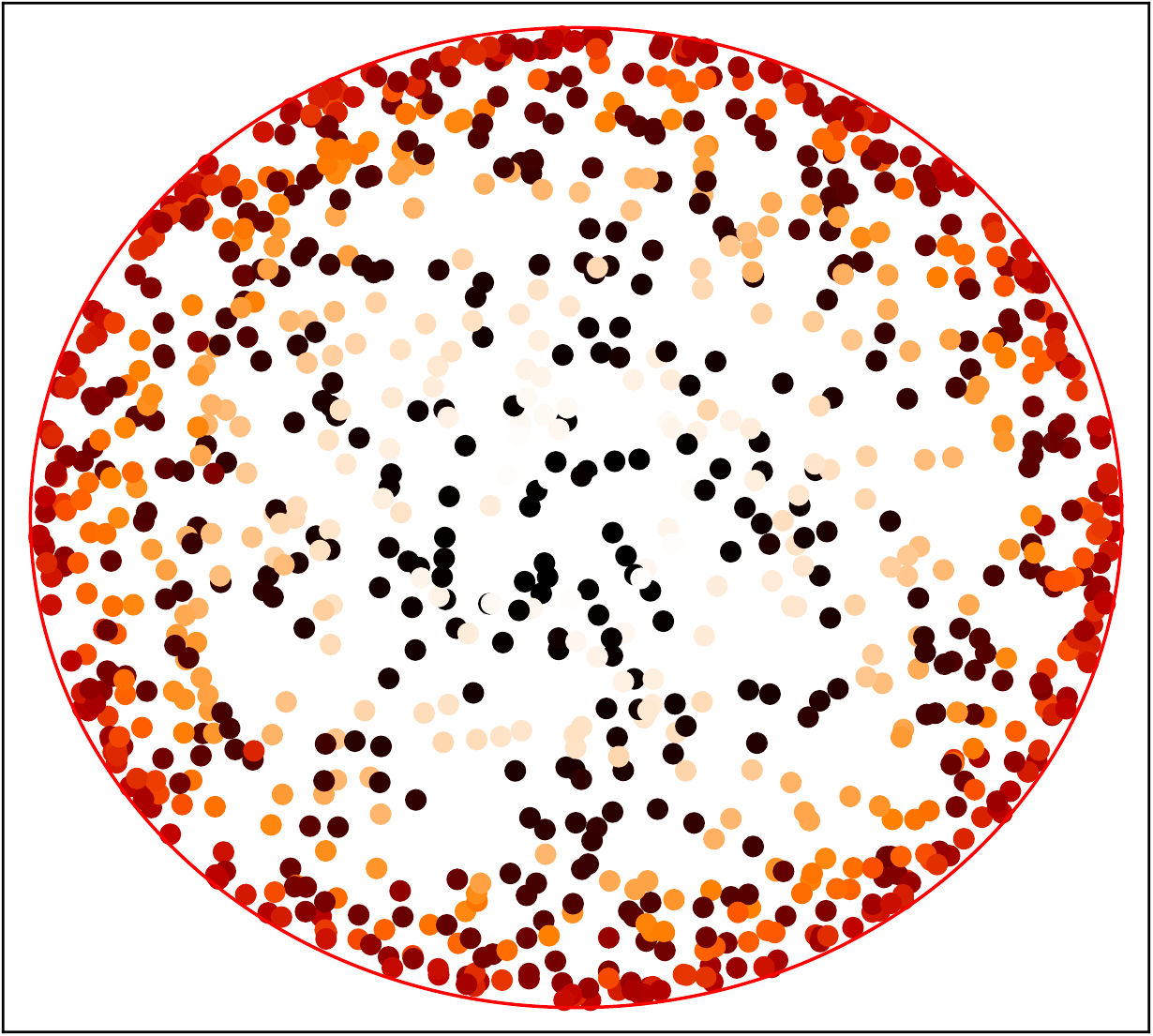} }}
  \subfloat[$\Hyp(3)$ -- exp map]{{\includegraphics[scale=0.25]{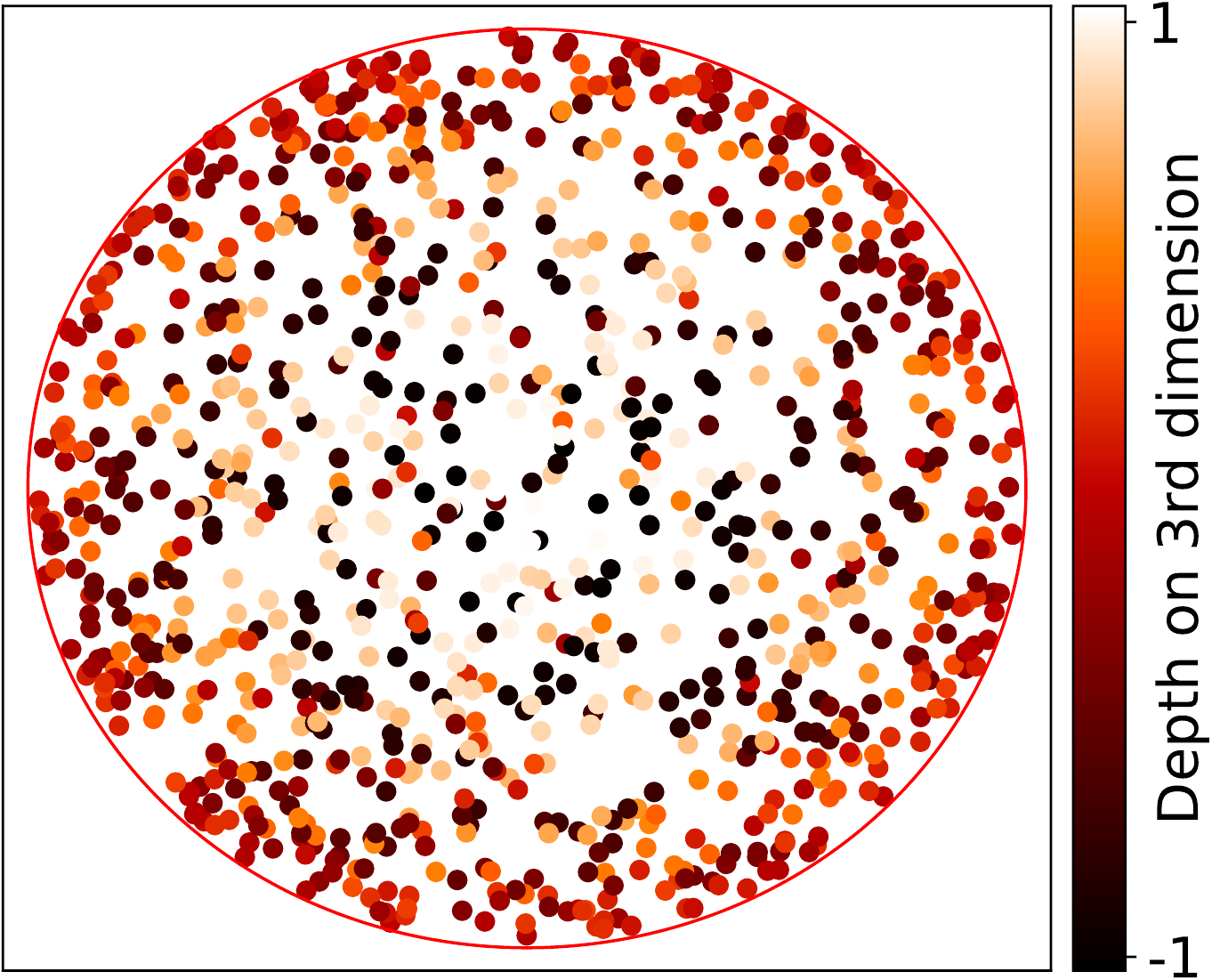} }}
  \hspace*{0.1cm}
  \subfloat[$\Sph(2)$ -- uniform]{{\includegraphics[scale=0.25]{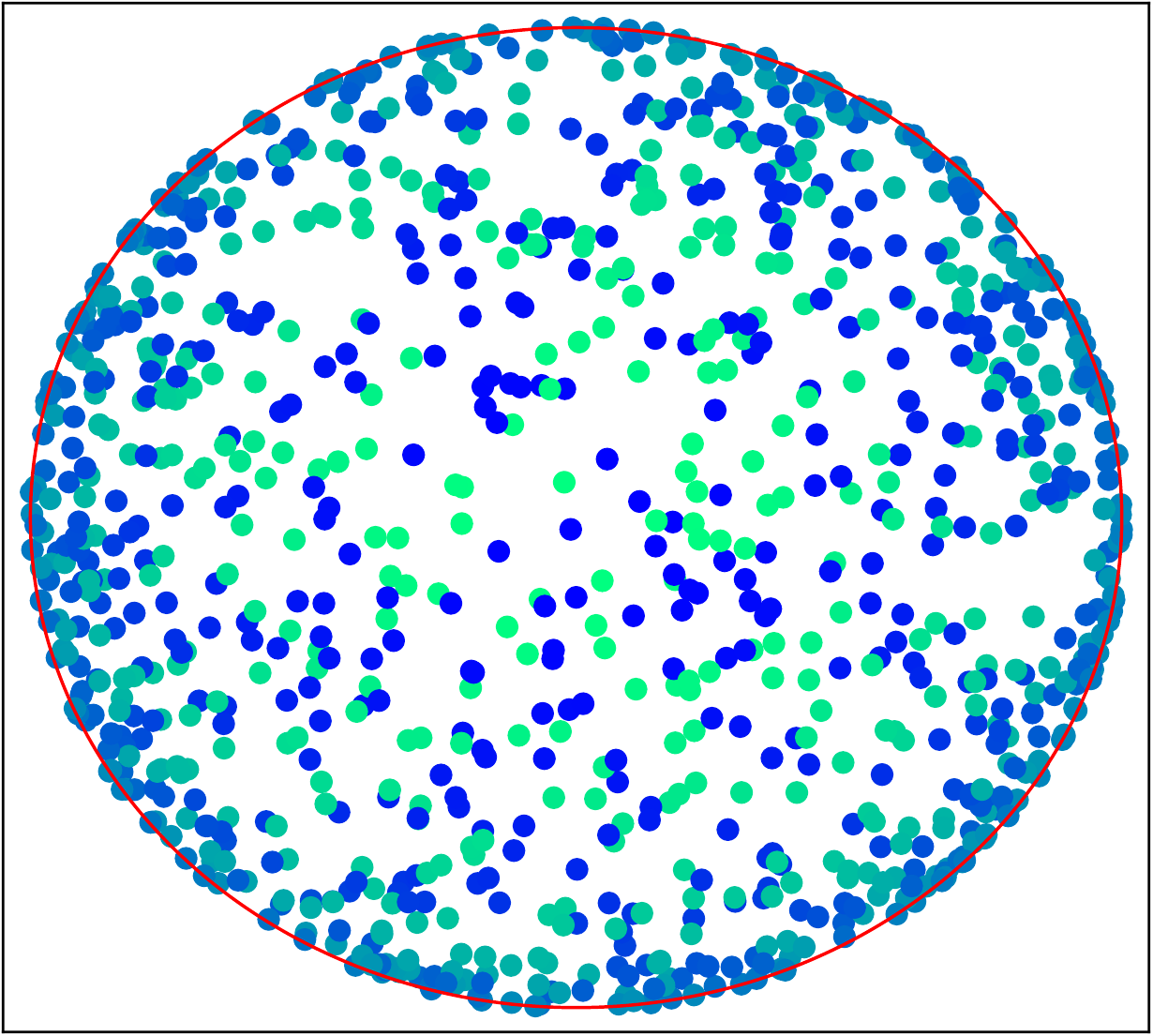} }}
  \subfloat[$\Sph(2)$ -- exp map]{{\includegraphics[scale=0.25]{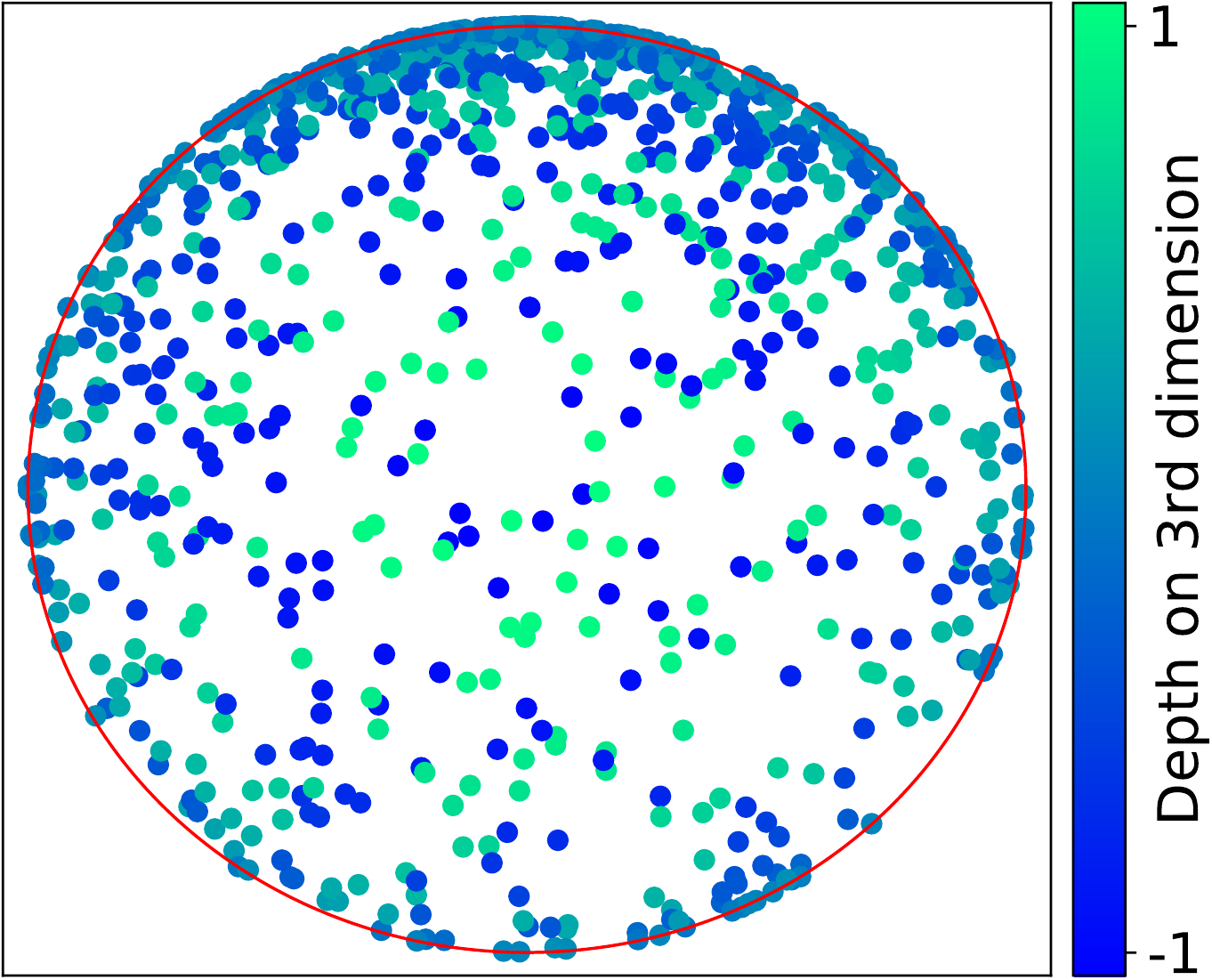} }}

  \caption[Visualization of the bias incurred when sampling through the exponential map]{%
    Visualization of the bias incurred when sampling through the exponential map. For the
    $3$-dimensional Poincar\'e ball (left), uniform samples (from a geodesic ball of radius $5$) are
    more concentrated towards the boundary. The bias is even more apparent for the $2$-dimensional
    sphere (right), where sampling through the exponential map at the south pole, in a ball of
    radius $\pi$, yields too many samples around the north pole. That is because the curvature is
    not taken into account.}
  \label{fig:sampling-bias-hyp-sph}
\end{figure}

%%%
\paragraph{A Word on Uniform Sampling.}
Ideally, to follow~\cite{krioukov2010hyperbolic} and to recover their results, we would have to
sample uniformly from geodesic balls of some fixed radius. However, this is non-trivial for
arbitrary Riemannian manifolds and, in particular, to the best of our knowledge, for the SPD
manifold. One would have to sample from the corresponding Riemannian measure while enforcing the
maximum distance constraint given by the geodesic ball. More precisely, using the following formula
for the measure $\mathrm{d} \mu_g$ in terms of the Riemannian metric $g(x)$, expressed in a normal
coordinates system $x$~\citep[see, e.g.,][]{pennec2006intrinsic}
\begin{equation}
  \mathrm{d} \mu_g = \sqrt{\det g(x)}\ \mathrm{d}^n x,
\end{equation}
one can sample uniformly with, for instance, a rejection sampling algorithm, as long as the right
hand side can be computed \emph{and} the parametrization allows enforcing the support constraint.
This is possible for hyperbolic and elliptical spaces through their polar parametrizations.  In
spite of our efforts, we have not managed to devise a similar procedure for the SPD manifold.
Instead, we have chosen a non-uniform sampling approach, applied consistently, which first generates
uniform tangent vectors from a ball in the tangent space at some point and then maps them onto the
manifold via the exponential map. Note that the manifold homogeneity guarantees that we get the same
results irrespective of the chosen base point, so in our experiments we sample around the identity
matrix. The difference from the uniform distribution is visualized
in~\Cref{fig:sampling-bias-hyp-sph} for $\Hyp(3)$ and $\Sph(2)$.

\begin{figure}
  \centering
  \subfloat[node degrees; $n = 3$]{{\includegraphics[scale=0.30]{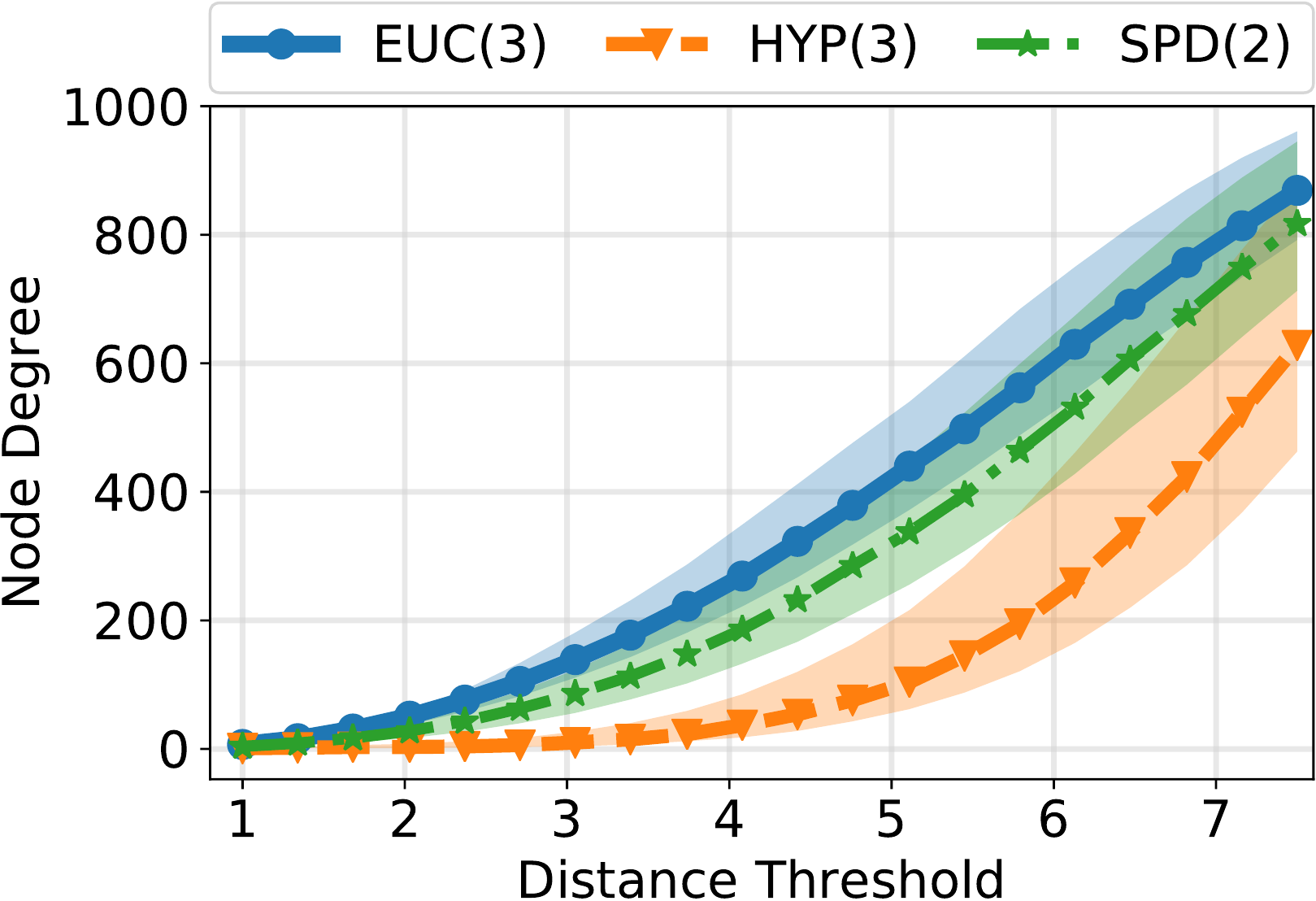} }}
  \subfloat[node degrees; $n = 6$]{{\includegraphics[scale=0.30]{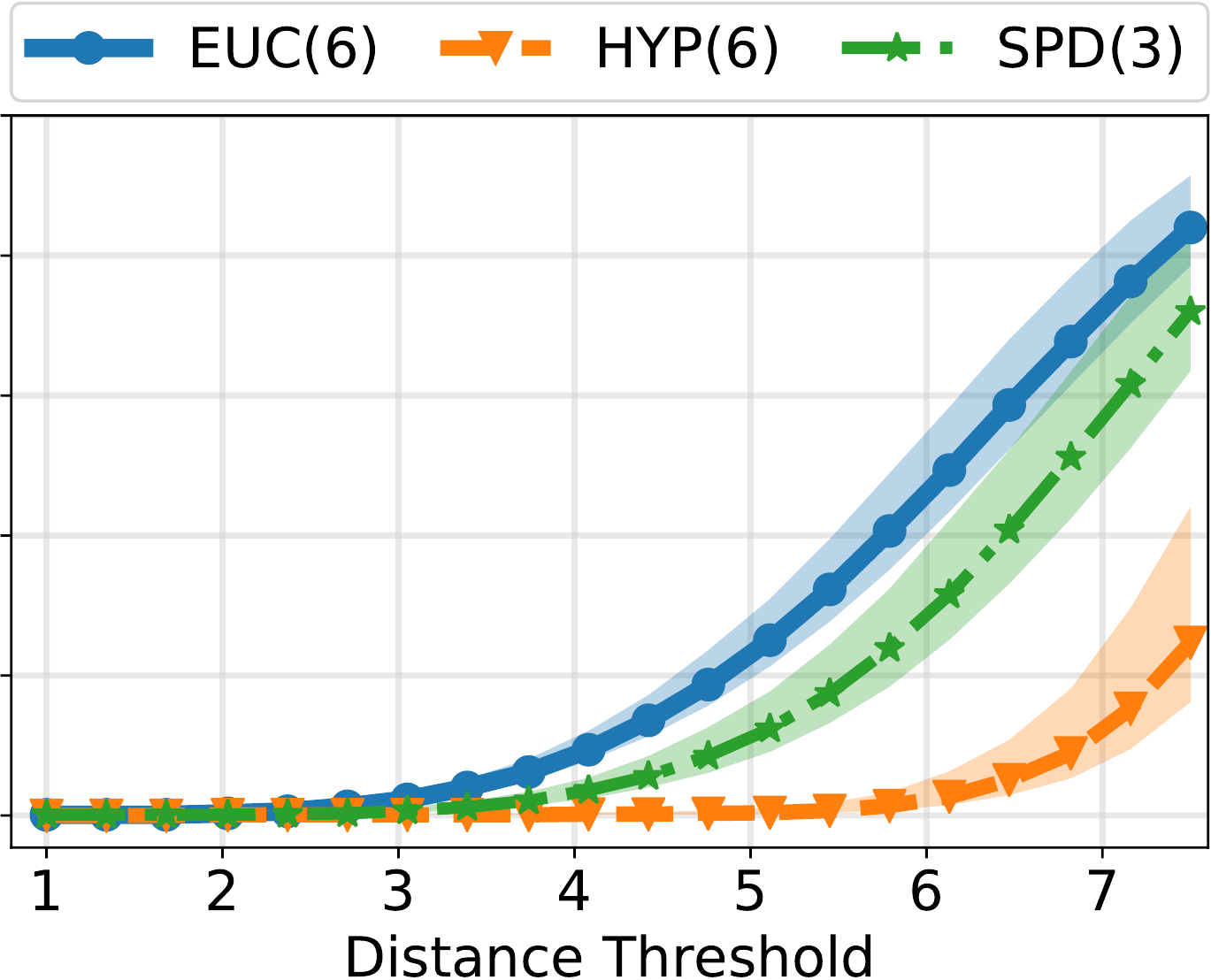} }}
  \subfloat[node degrees; $n = 10$]{{\includegraphics[scale=0.30]{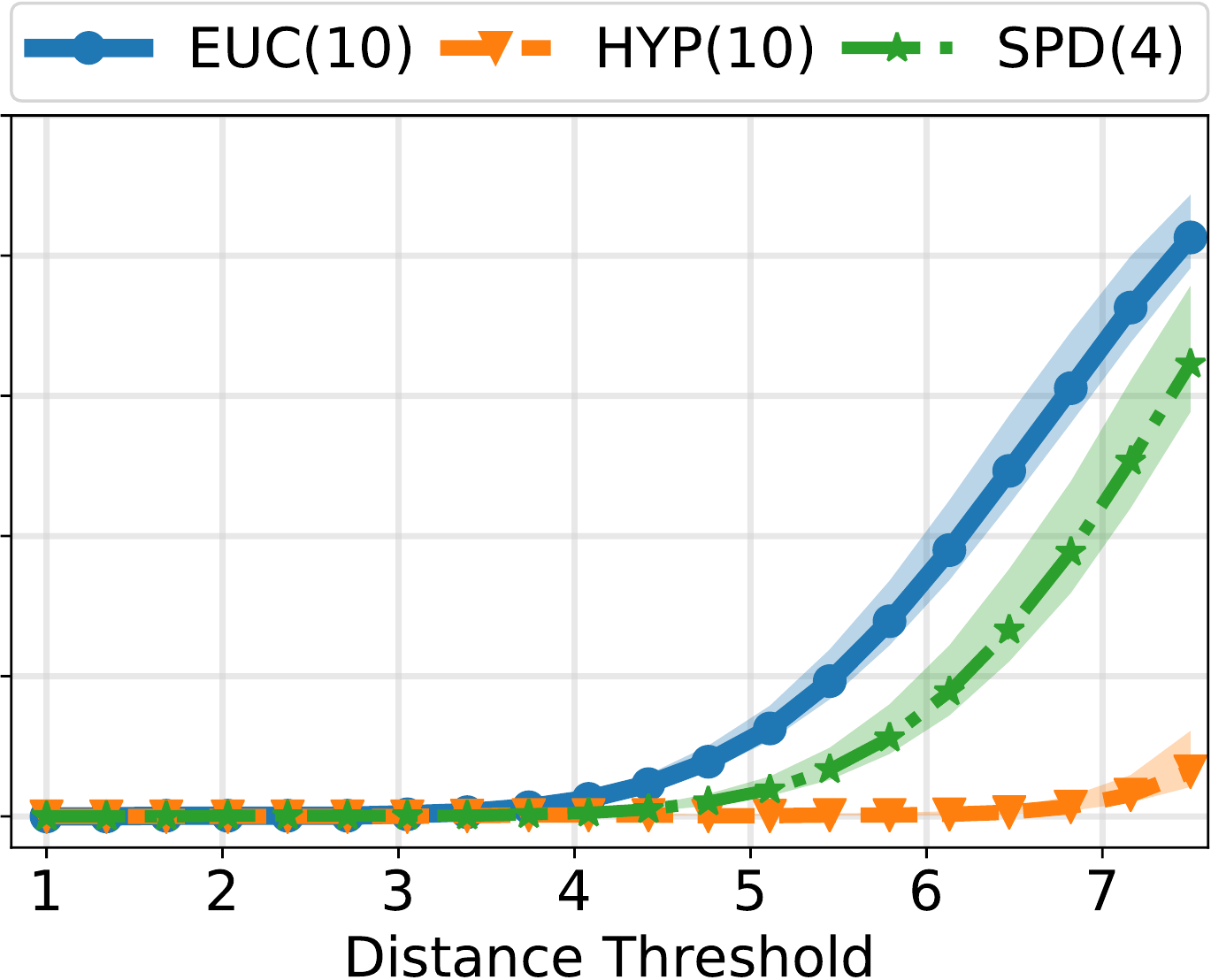} }}

  \subfloat[graph sec.\ curv.; $n = 3$\label{fig:noncomp-curv-sec3}]{{\includegraphics[scale=0.30]{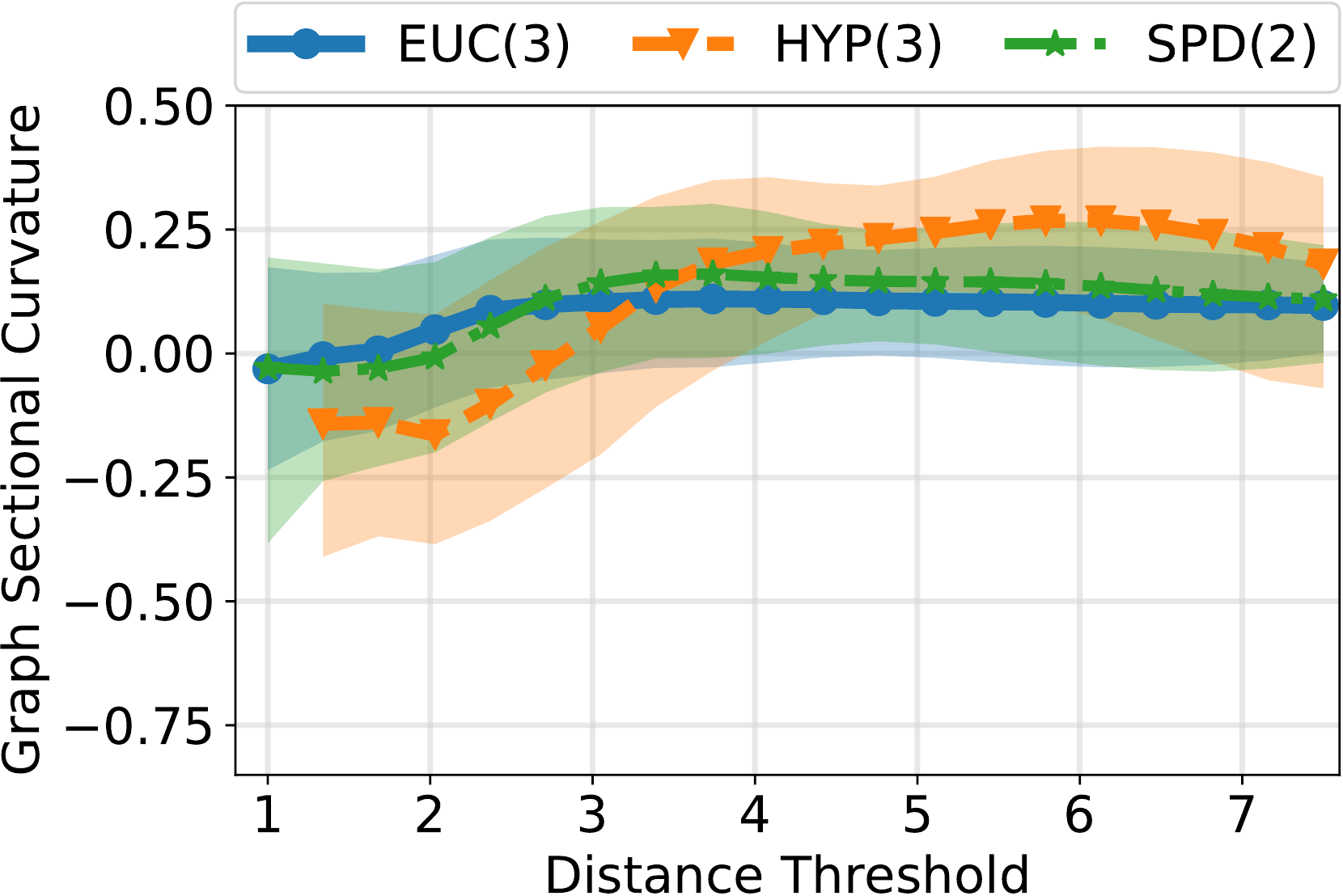} }}
  \subfloat[graph sec.\ curv.; $n = 6$\label{fig:noncomp-curv-sec6}]{{\includegraphics[scale=0.30]{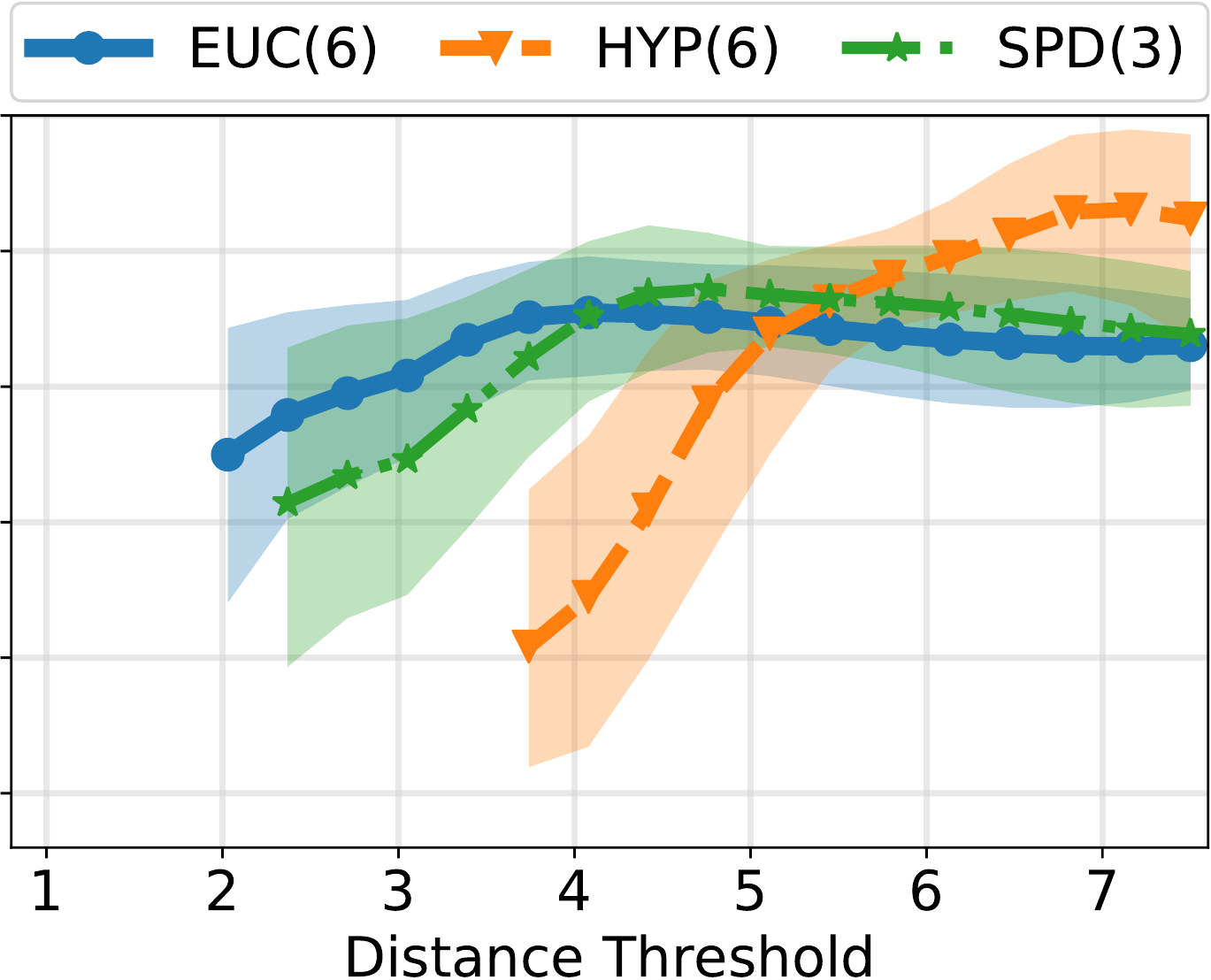} }}
  \subfloat[graph sec.\ curv.; $n = 10$\label{fig:noncomp-curv-sec10}]{{\includegraphics[scale=0.30]{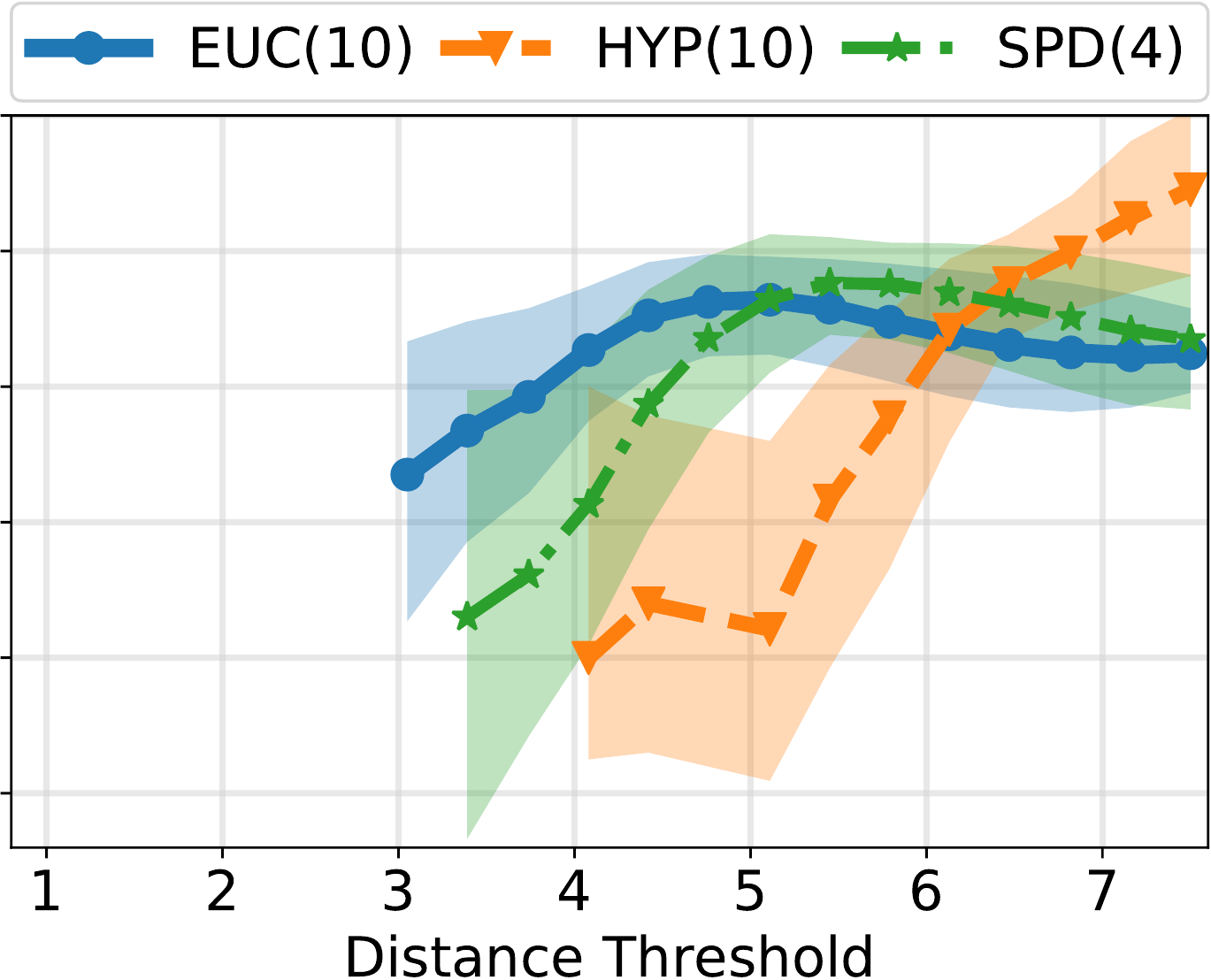} }}

  \subfloat[man.\ angles; $n = 3$]{{\includegraphics[scale=0.30]{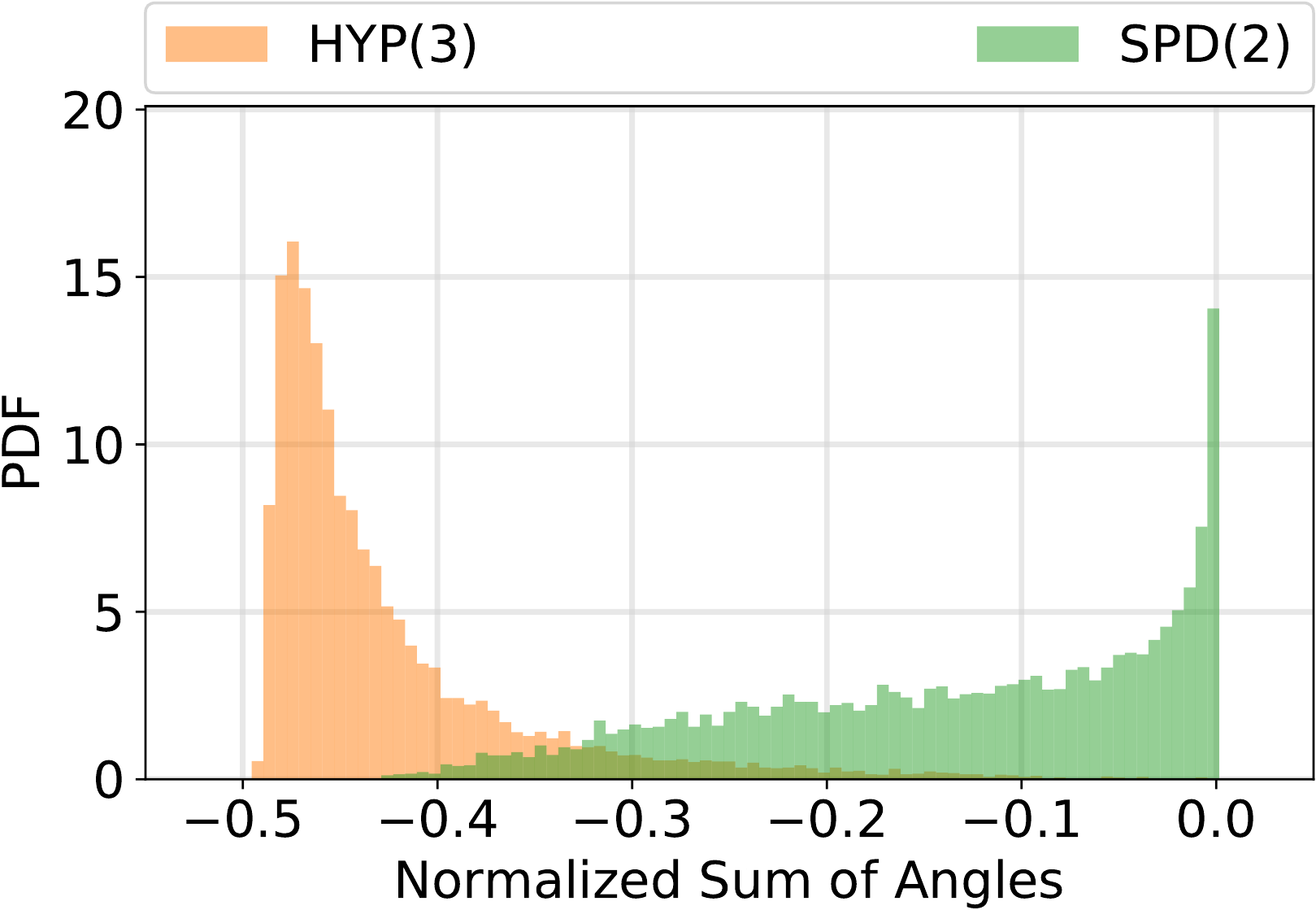} }}
  \subfloat[man.\ angles; $n = 6$]{{\includegraphics[scale=0.30]{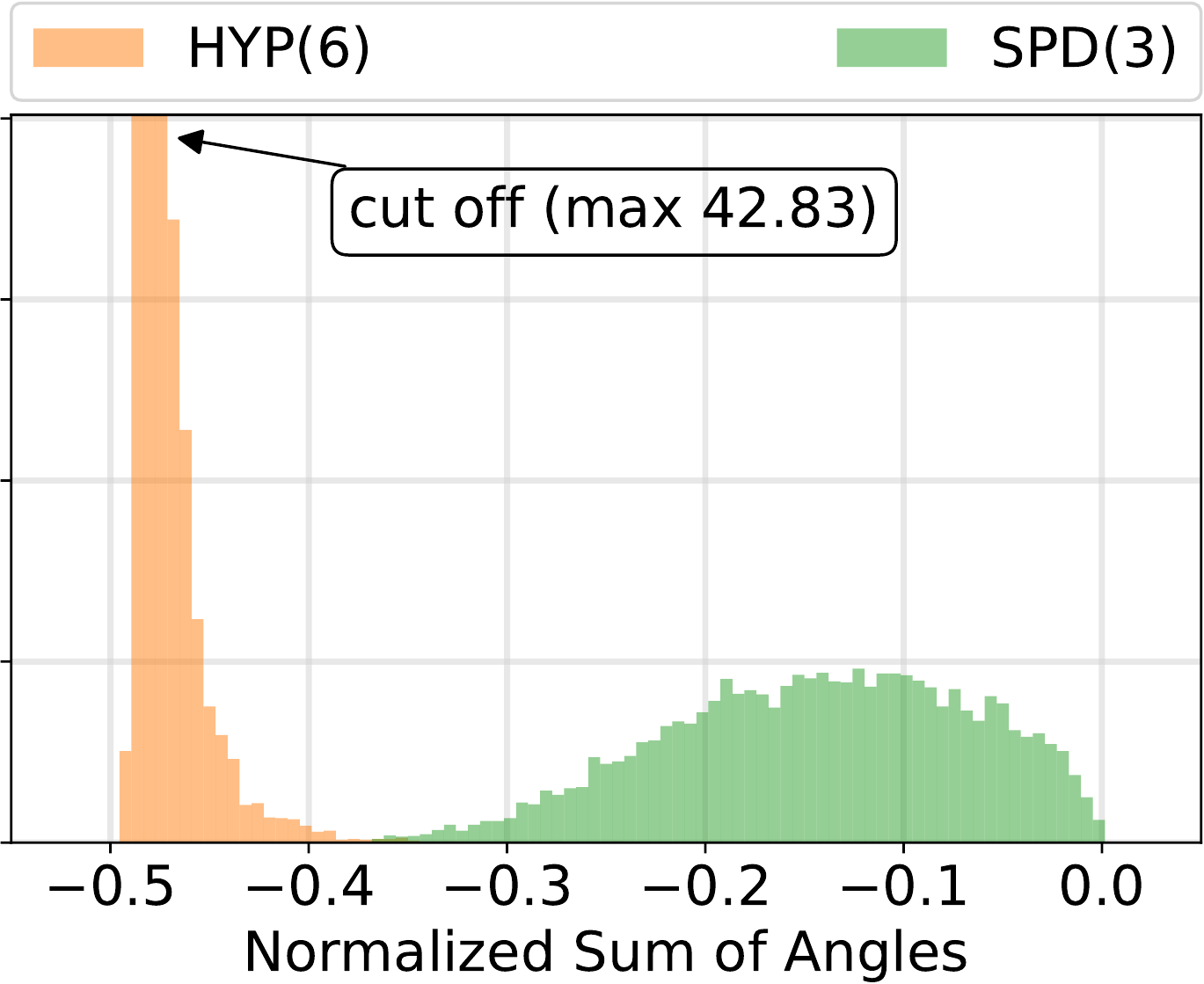} }}
  \subfloat[man.\ angles; $n = 10$]{{\includegraphics[scale=0.30]{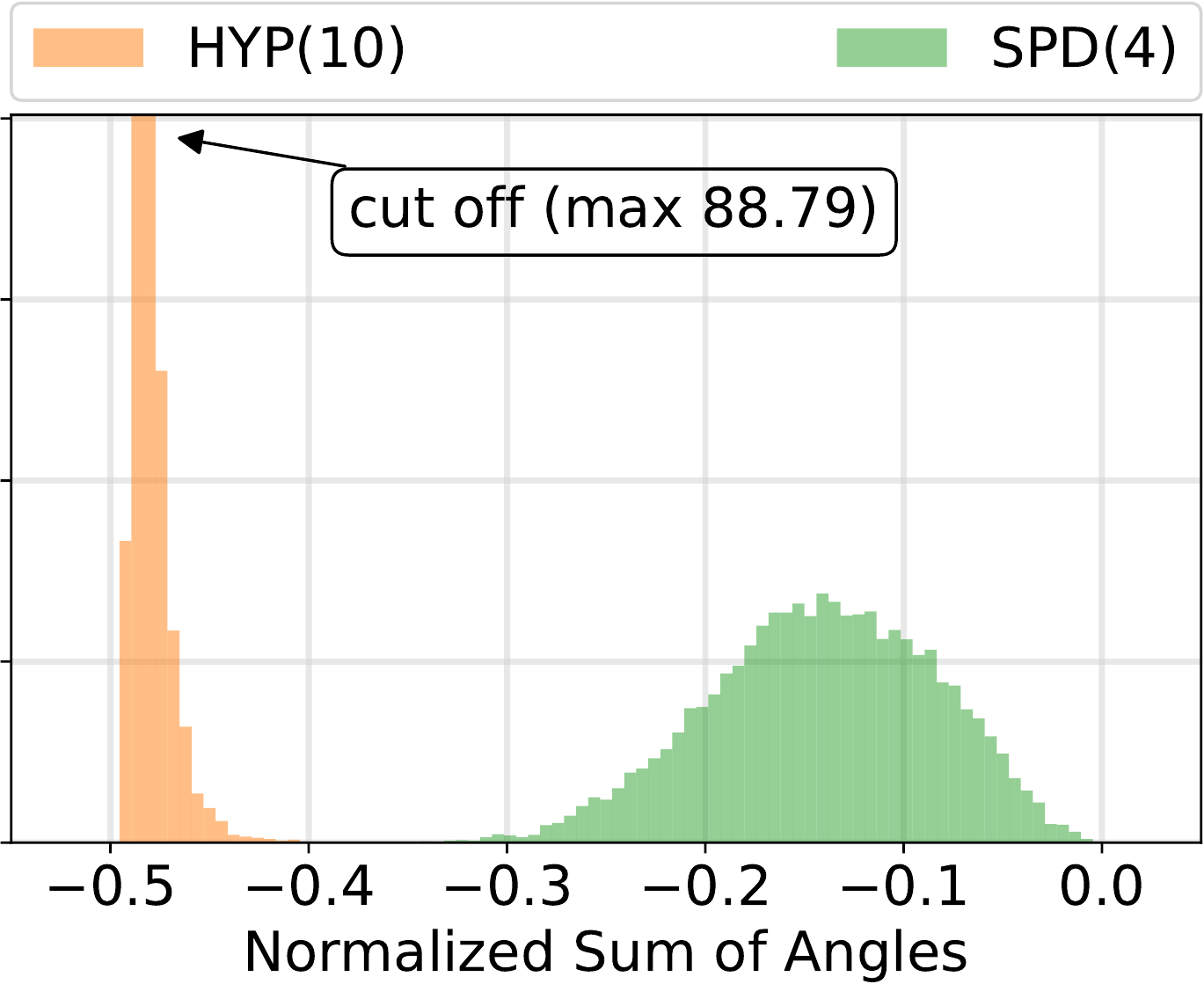} }}

  \caption[Analysis of random graphs sampled from geodesic balls in $\R^n$, $\Hyp(n)$, and
  $\Spd(n)$, respectively.]{Analysis of graphs sampled from geodesic balls in $\R^n$, $\Hyp(n)$, and
  $\Spd(n)$. The plots follow the same organization as in~\Cref{fig:comp-gen-results}. The points
  are obtained by sampling tangent vectors from a ball of radius $5$, uniformly, and mapping them
  onto the manifold through the exponential map. A few points are missing from the second set of
  plots because the largest connected components in the corresponding graphs have too few nodes
  (less than 100).}
  \label{fig:noncomp-gen-results}
\end{figure}

With that, we proceed to discussing the results from~\Cref{fig:noncomp-gen-results}. They are
organized as in the previous subsection. One difference, besides the sampling procedure, is that we
now vary the distance threshold used to link nodes in the graphs from $R / 10$ to $1.5 R$, where $R$
is the radius of the geodesic balls (i.e., half of the maximum distance between any two points). It
is identically $5$ in all scenarios.

%%%
\paragraph{Degree Distributions.} The first obvious characteristic is that the Euclidean space now
bounds from above the two non-flat manifolds (that is, their corresponding median curves).
Contrast this with the compact manifolds (\Cref{fig:comp-gen-results}). It tells us that points are
in general farther away than the others (relative to the Euclidean space), which is expected, given
their (partly) negative curvature. Moreover, recall that the SPD manifold is a higher-rank symmetric
space, which means that there are tangent space subspaces with dimension greater than 1 on which the
sectional curvature is 0. In light of this property, its interpolating behavior (of Euclidean and
Hyperbolic curves), which is already apparent from the evolution of the degree distribution, is not
surprising. Note that in $10$ dimensions, the graphs constructed from the hyperbolic space are
poorly connected even for the higher end of the threshold range. A similar trend characterizes the
Euclidean and SPD spaces, but at a lower rate. In the limit $n \to \infty$, essentially all the mass
is at the boundary of the geodesic ball for all of them, but \emph{how quickly} this happens is what
distinguishes them.

%%%
\paragraph{Graph Sectional Curvatures.} Here, the difference between the three spaces is hardly
noticeable for $n=3$, in~\Cref{fig:noncomp-curv-sec3}, but it becomes clearer in higher dimensions.
We see that as long as the graphs are below a certain median degree, the estimated discrete
curvatures are mostly negative, as one would expect. Furthermore, the order of the empirical
distributions seems to match the intuition built in the previous paragraph: the hyperbolic space is
``more negatively curved'' than the SPD manifold.

%%%
\paragraph{Triangles Thickness.} The last row in~\Cref{fig:noncomp-gen-results} serves as additional
evidence that the SPD manifold is not ``as negatively curved'' as the hyperbolic space. Without any
further comments, we need only emphasize that, as the dimension increases, almost all hyperbolic
triangles are ``ideal triangles'', i.e., the largest possible triangles in hyperbolic geometry, with
the sum-of-angles equal to 0. The analogous histograms for the SPD manifold shift and peak slightly
to the left but at a much slower pace.

\clearpage
\section{Experimental Setup}\label{sec:exp-setup}
The training procedure is the same across all the experiments from~\Cref{sec:experiments}.  The
structure of the input graphs is not altered. We compute all-pairs shortest-paths in each one of the
input graphs and serialize them to disk. They can then be reused throughout our experiments. Since
we are not interested in representing the distances exactly (in absolute value) but only relatively,
we max-scale them. This has the (empirically observed) advantage of making learning less sensitive
to the scaling factors (see below).

\paragraph{The Task.}
We emphasize that we consider the \emph{graph reconstruction} task in~\Cref{sec:experiments}.
Hence, all results correspond to embedding the input graphs containing all information: no edges or
nodes are left out. In other words, we do not evaluate generalization. That is why for each
embedding space and fixed dimension, we report the best performing embedding across all runs and
loss functions. This is consistent with our goal of decoupling learning objectives and evaluation
metrics (see~\Cref{sec:decoupling}). We have chosen to restrict our focus as such in order to have
fewer factors in our ablation studies. As future work, it is natural to extend our work to
downstream tasks and generalization-based scenarios, and study the properties of the introduced
matrix manifolds in those settings.

\paragraph{Training.}
We train for a maximum of 3000 epochs with batches of 512 nodes -- that is, 130816 pairwise
distances. We use the burn-in strategy from~\cite{nickel2017poincare,ganea2018hyperbolic}: training
with a 10 times smaller learning rate for the first 10 epochs. Moreover, if the per-epoch loss does
not improve for more than 50 epochs, we decrease the learning rate by a factor of 10; the training
is ended earlier if this makes the learning rate smaller than $10^{-5}$. This saves time without
affecting performance.

\paragraph{Optimization.}
We repeat the training, as described so far, for three optimization settings. The first one uses
\textsc{Radam}~\cite{becigneul2018riemannian} to learn the embeddings, which we have seen to be the
most consistent across our early experiments. In the other two, we train the embeddings using
\textsc{Rsgd} and \textsc{Radam}, respectively, but we also train a scaling factor of pairwise
distances (with the same optimizer). This is inspired by~\cite{gu2018learning}. The idea is that
scaling the distance function is equivalent to representing the points on a more or less curved
sphere or hyperboloid. In the spirit of Riemannian SNE (\Cref{sec:decoupling}), this can also be seen as
controlling how peaked around the MAP configuration the resulting distribution should be. We have
chosen to optimize without scaling in the first setting because it seemed that even for simple,
synthetic examples, jointly learning the scaling factor is challenging.

We have also experimented with an optimization inspired by deterministic
annealing~\cite{rose1998deterministic}: starting with a high ``temperature'' and progressively
cooling it down. Since we did not see significant improvements, we did not systematically employ
this approach.

\paragraph{Computing Infrastructure.}
We used 4 NVIDIA GeForce GTX 1080 Ti GPUs for the data-parallelizable parts.

\clearpage
\section{Input Graphs -- Properties \& Visualizations}\label{sec:input-graphs}
\begin{figure}[t]
  \subfloat{{\includegraphics[scale=0.3]{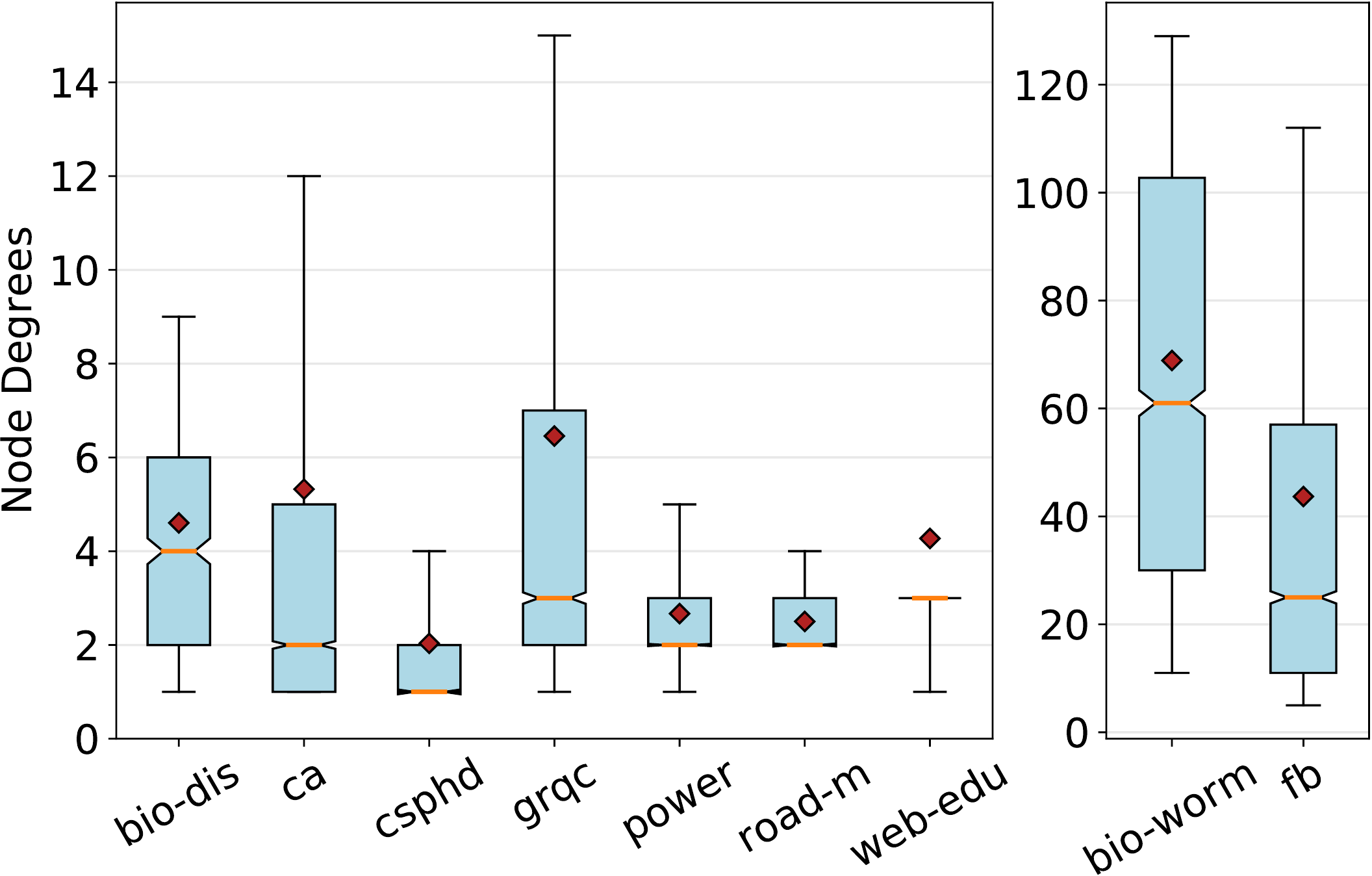} }}
  \hfill
  \subfloat{{\includegraphics[scale=0.3]{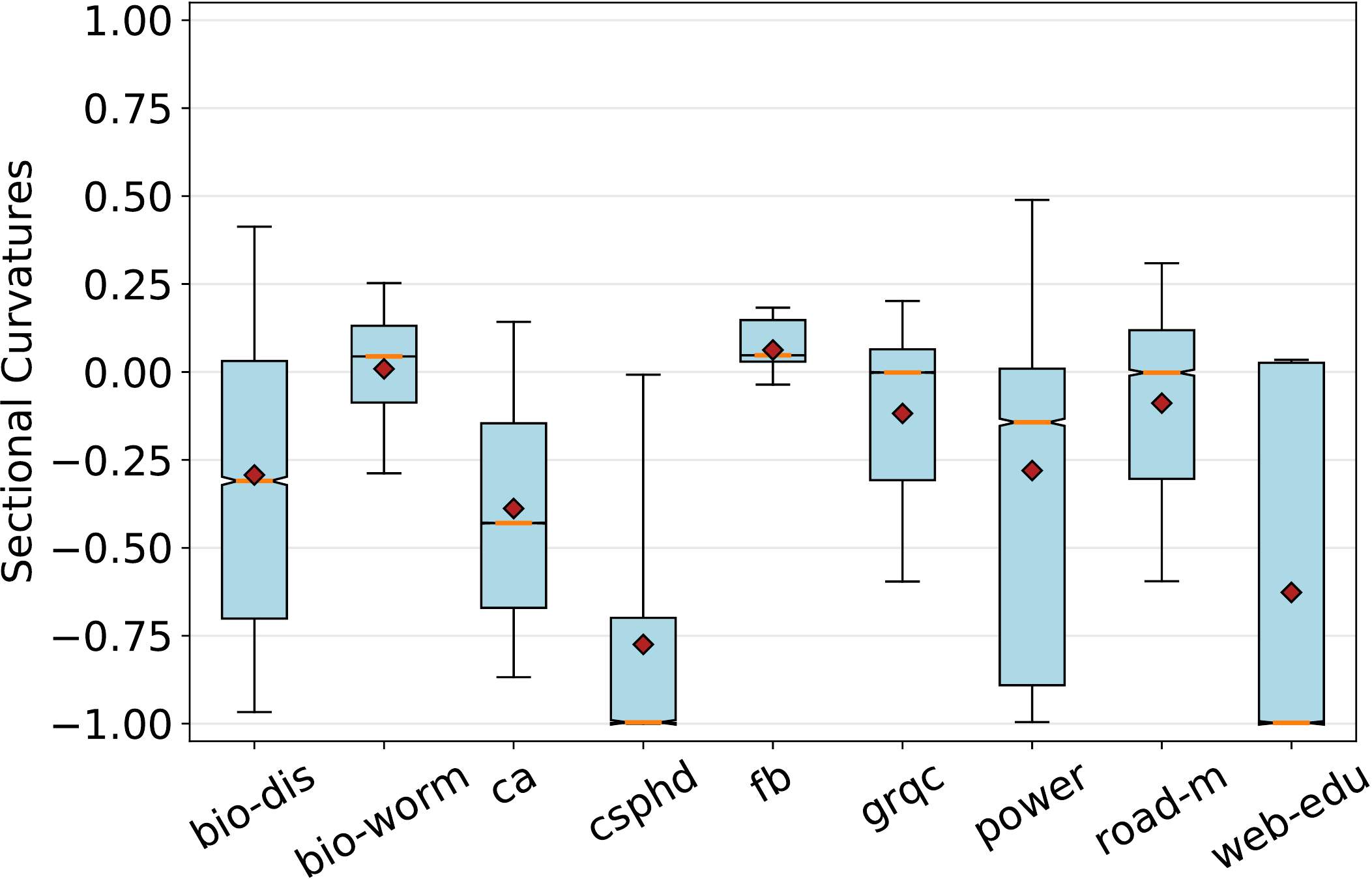} }}
  \vspace*{0.2cm}

  \centering
  \small
  \renewcommand{\arraystretch}{1.2}
  \begin{tabular}{@{}lSSSSSSSSS@{}}
    \toprule
    & {bio-dis} & {bio-worm} & {ca} & {csphd} & {fb} & {grqc} & {power} & {road-m} & {web-edu} \\
    \midrule $\delta$-mean & 0.20 & 0.24 & 0.31 & 0.51 & 0.10 & 0.38 & 0.96 & 3.13 & 0.02 \\
    \midrule $\delta$-max & 2.50 & 2.00 & 2.50 & 6.50 & 1.50 & 3.00 & 10.00 & 25.00 & 1.00 \\

    \bottomrule
  \end{tabular}

  \caption{The node degrees, (graph) sectional curvatures, and $\delta$-hyperbolicities (including
  the mean of the empirical distribution sampled as shown in~\cite{cohen2015computing}) of the input
  graphs used throughout our experiments with non-positively curved manifolds. Shortened dataset
  names are used to save space.}
  \label{fig:graphs-summary}
\end{figure}

\begin{figure}
  \centering

  \subfloat[``bio-diseasome'' ($\abs{V} = 516, \abs{E} = 1188$) -- disorder-gene associations~\cite{goh2007human}]{{
    \includegraphics[scale=0.115]{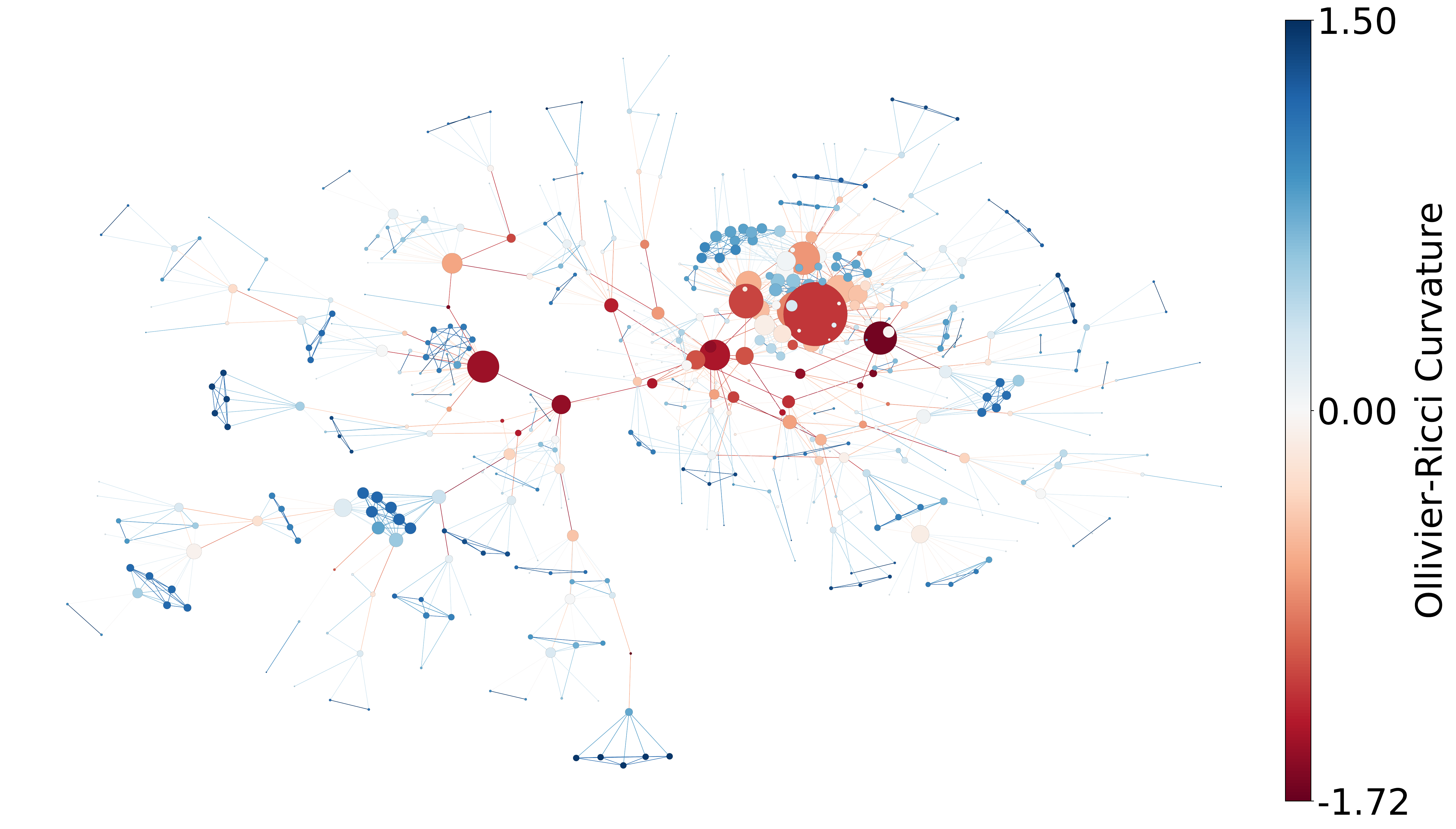} }}
  \hspace*{0.5cm}
  \subfloat[``bio-wormnet'' ($\abs{V} = 2274, \abs{E} = 78328$) -- worms gene network~\cite{cho2014wormnet}]{{
    \includegraphics[scale=0.115]{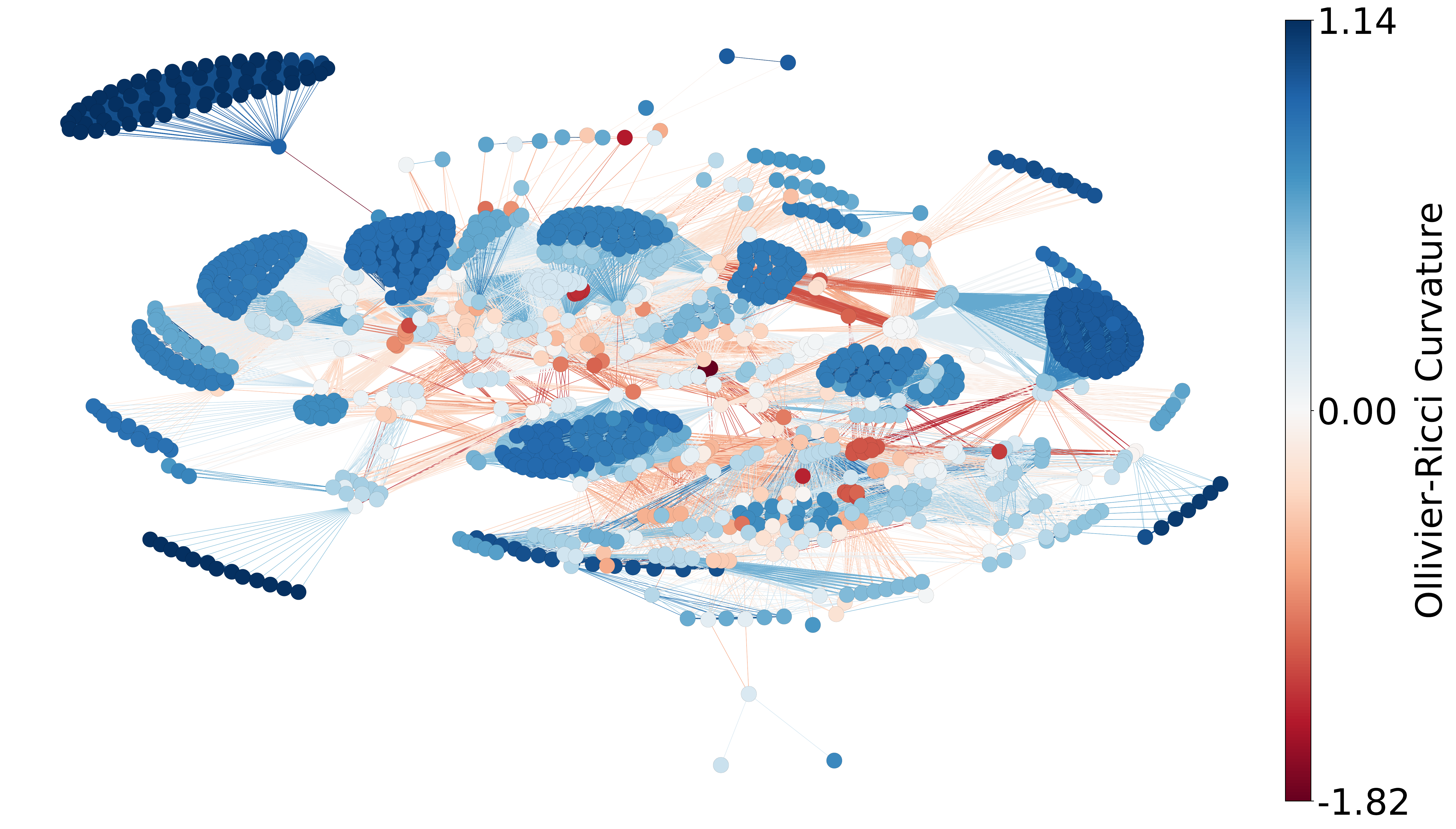} }}

  \subfloat[``california'' ($\abs{V} = 5925, \abs{E} = 15770$) -- \cite{nr}]{{
    \includegraphics[scale=0.115]{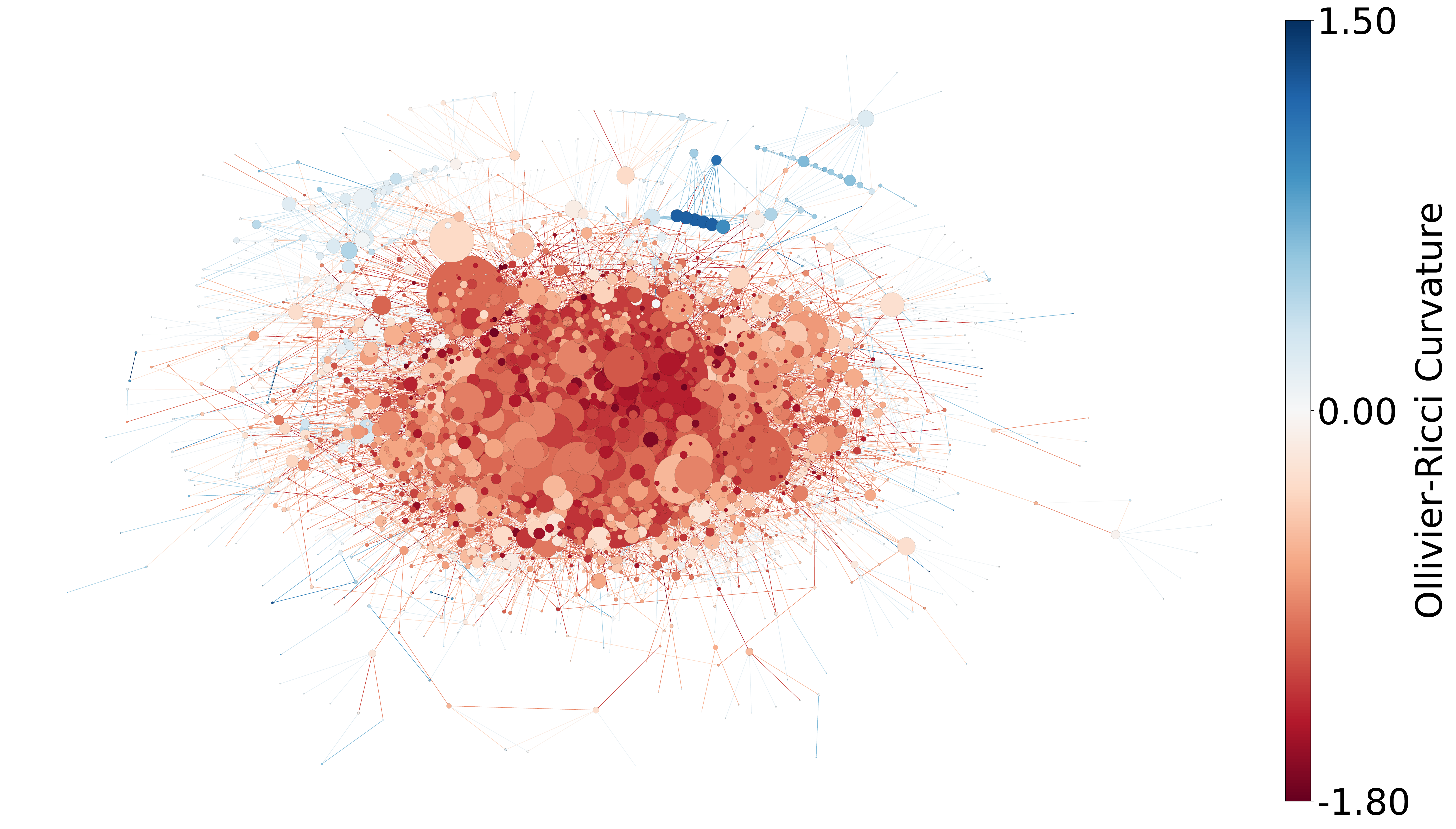} }}
  \hspace*{0.5cm}
  \subfloat[``csphd'' ($\abs{V} = 1025, \abs{E} = 1043$) -- scientific collaboration network~\cite{de2018exploratory}]{{
    \includegraphics[scale=0.115]{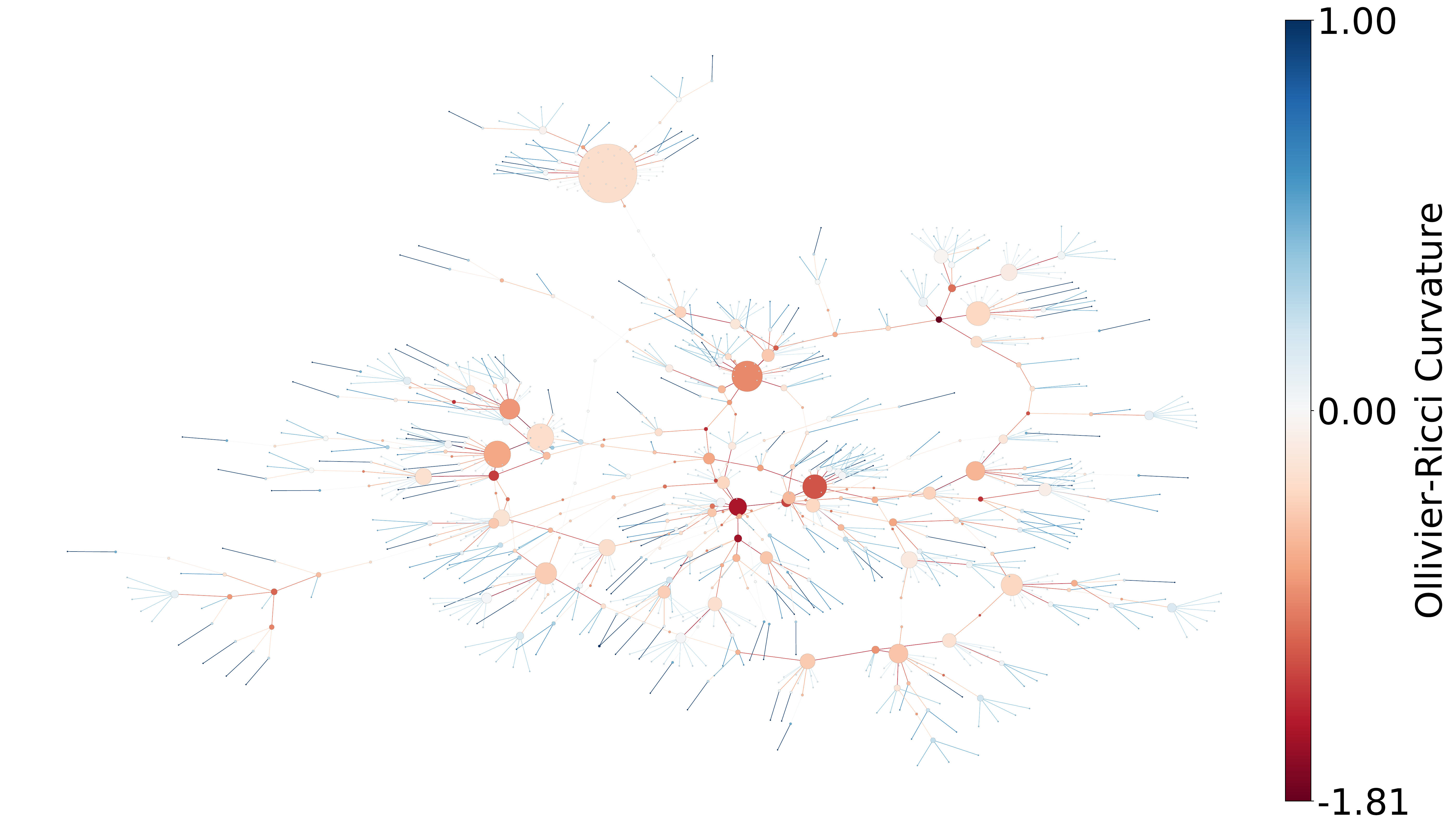} }}

  \subfloat[``facebook'' ($\abs{V} = 4039, \abs{E} = 88234$) -- dense social network from Facebook~\cite{leskovec2012learning}]{{
    \includegraphics[scale=0.115]{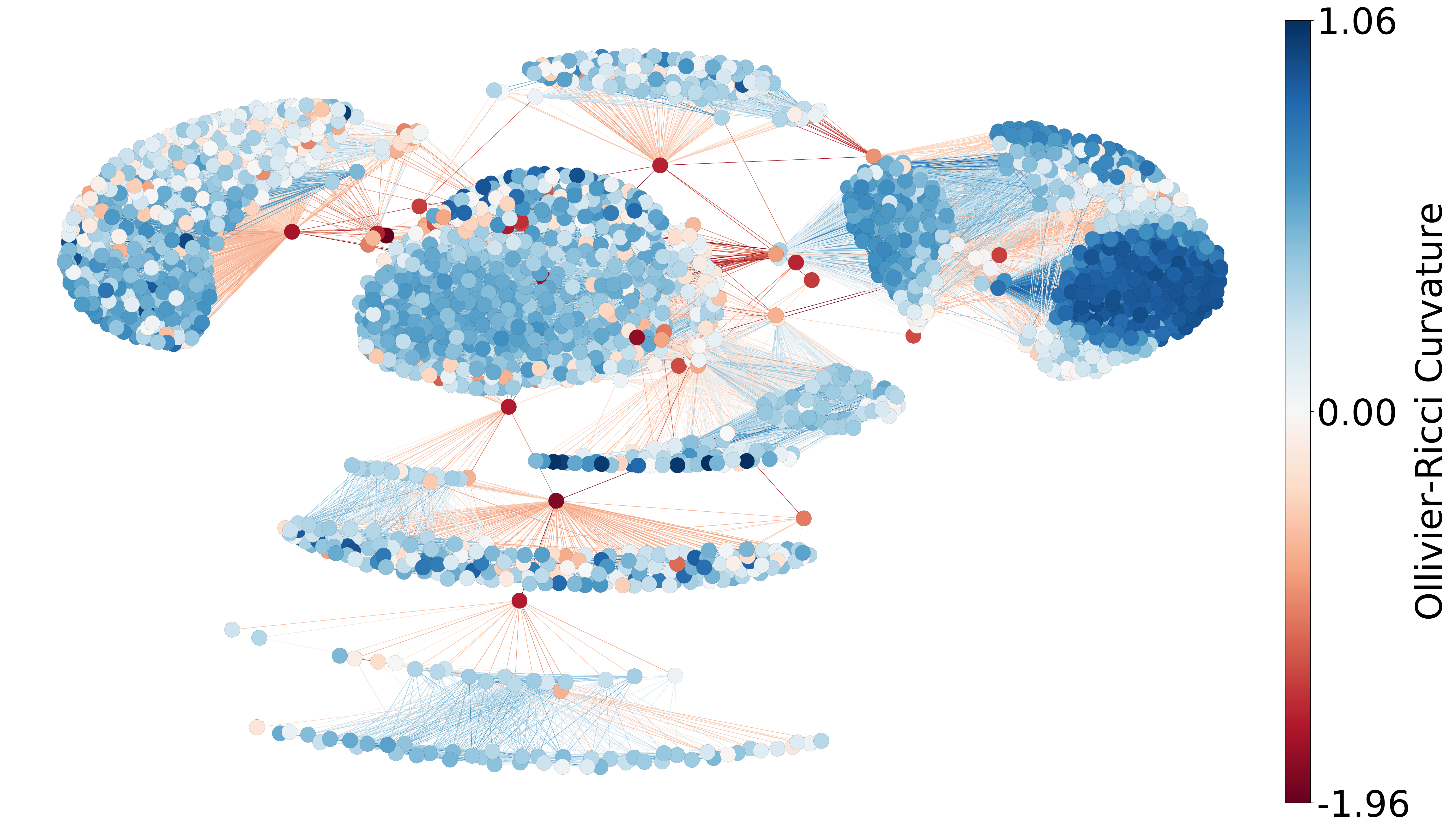} }}
  \hspace*{0.5cm}
  \subfloat[``grqc'' ($\abs{V} = 4158, \abs{E} = 13422$) -- collaboration network on ``General Relativity''~\cite{snapnets}]{{
    \includegraphics[scale=0.115]{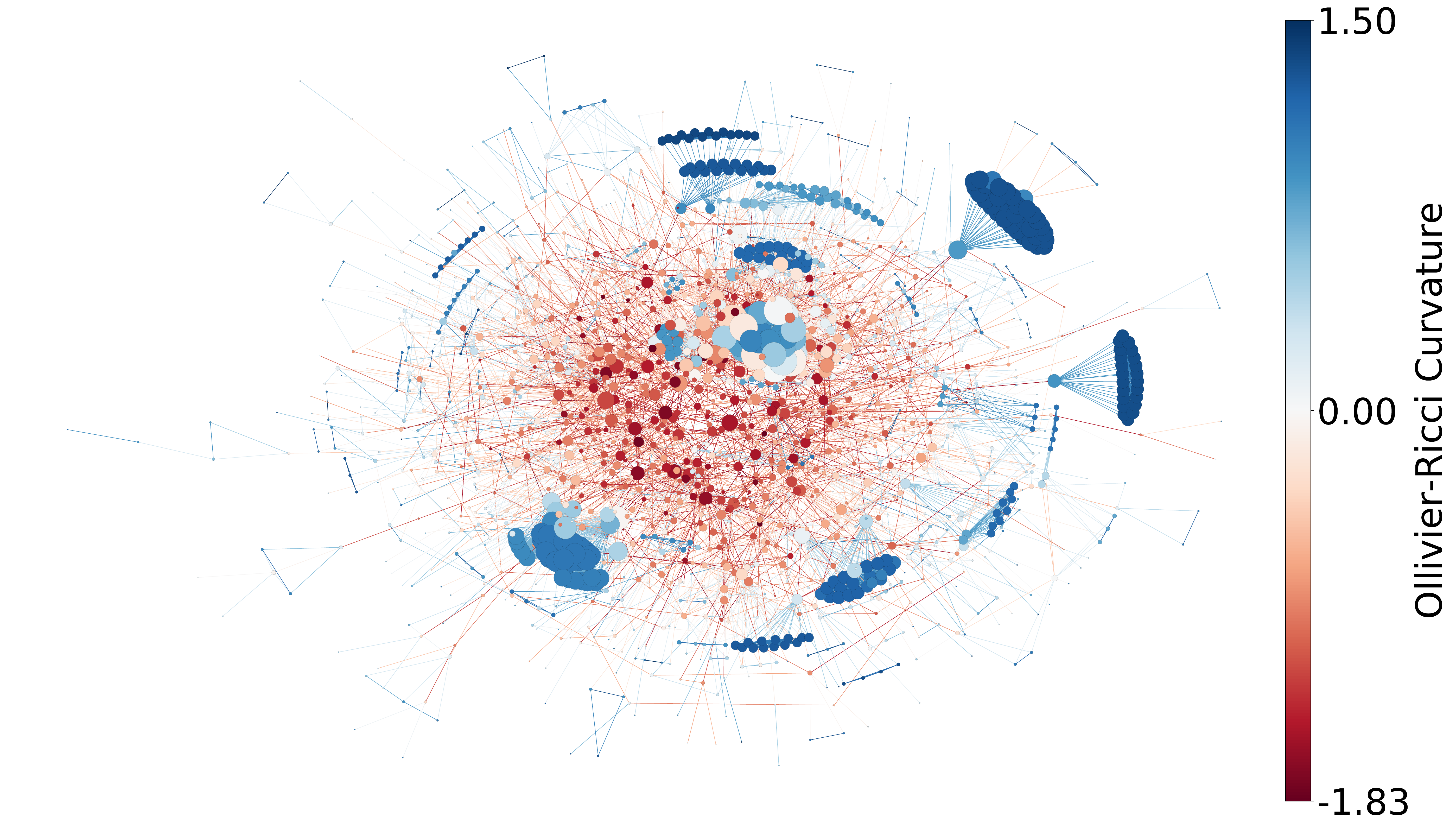} }}

  \subfloat[``power'' ($\abs{V} = 4941, \abs{E} = 6594$) -- grid distribution network~\cite{watts1998collective}]{{
    \includegraphics[scale=0.115]{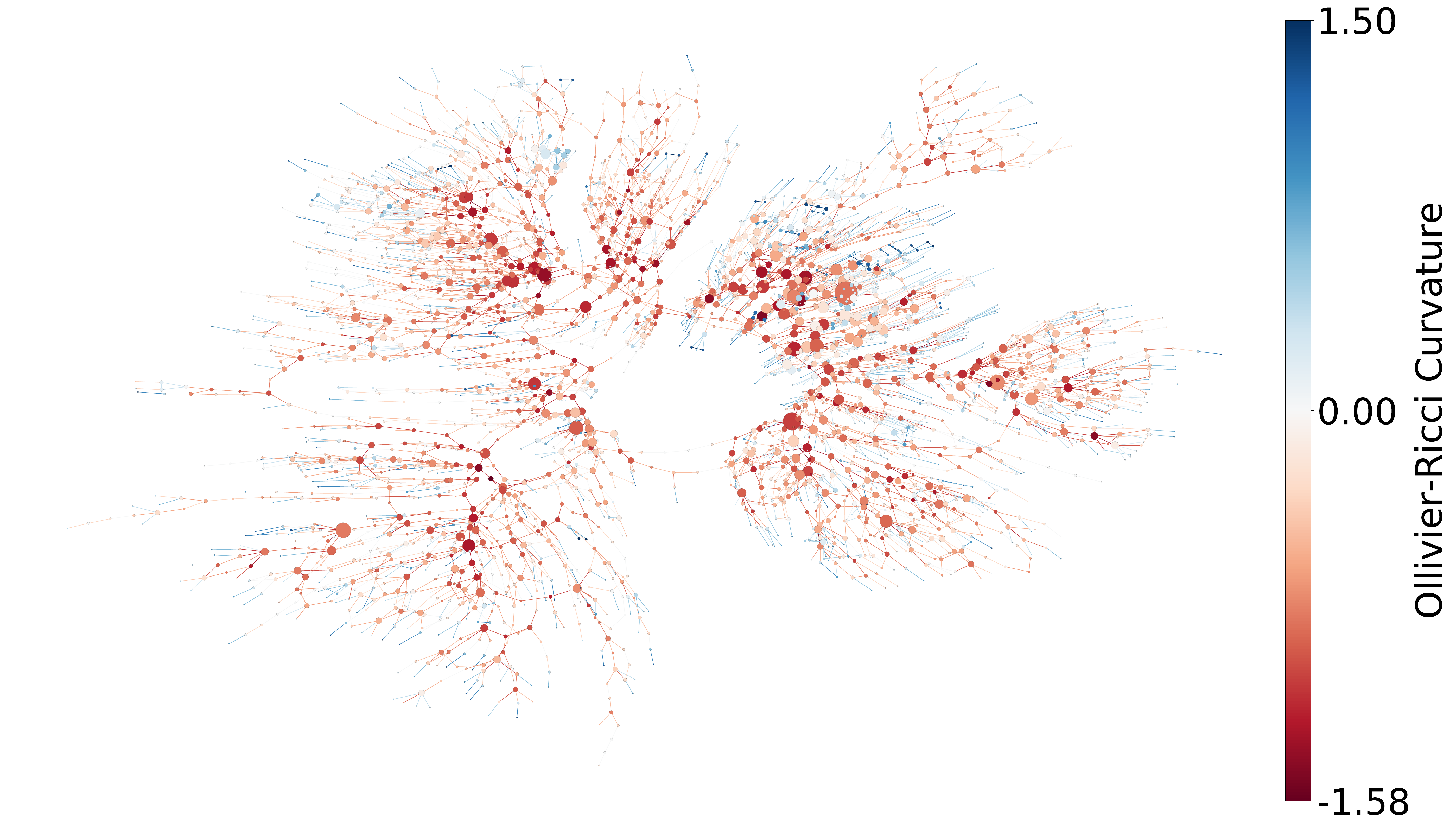} }}
  % \hspace*{0.5cm}
  % \subfloat[``web-edu'' ($\abs{V} = 3031, \abs{E} = 6474$) -- web network from the .edu domain~\cite{gleich2004fast}]{{
  %   \includegraphics[scale=0.115]{assets/img/ricci-web-edu} }}

  \caption{Visualizations of embedded graphs. The edge color depicts the corresponding
  Ollivier-Ricci curvature, as described in~\Cref{sec:geom-prop-graphs}. Similarly, each node is
  colored according to the average curvature of its adjacent edges. Thus, red nodes behave more like
  \emph{backbone} nodes, while blue nodes are either leaves or are part of a clique. In each plot,
  the size of each node is proportional to its degree.}
  \label{fig:graph-vis}
\end{figure}

First of all, we include a visualization of each one of the graph used throughout our graph
reconstruction experiments in~\Cref{fig:graph-vis}. In the rest of this section, we look at some of
their geometric properties, summarized in~\Cref{fig:graphs-summary}. We refer the reader
to~\Cref{sec:geom-prop-graphs} for background on the quantities discussed next.

We start with node degree distributions. Judging by the positions of the means (red diamonds)
relative to those of the medians (orange lines), we conclude that most degree distributions are
long-tailed, a feature of \emph{complex networks}. This is particularly severe for ``web-edu'',
where the mean degree is larger than the 90th percentile. Looking at~\Cref{fig:web-edu}, it becomes
clear what makes it so: there are a few nodes (i.e., web pages) to which almost all the others are
connected.

Next, we turn to the estimated sectional curvatures. They indicate a preference for the negative
half of the range, reinforcing the complex network resemblance. We see that many of the
distributions resemble those of the random graphs sampled from hyperbolic and SPD manifolds
(\Cref{fig:noncomp-curv-sec6,fig:noncomp-curv-sec10}).

Finally, with one exception, the graphs are close to $0$ hyperbolicity, which is an additional
indicator of a preference for negative curvature; but how much negative curvature is beneficial
remains, a-priori, unclear. The exception is ``road-minnesota'' which, as its name implies, is a
road network and, unsurprisingly, has a different geometry than the others. Its
$\delta$-hyperbolicity values are mostly large, an indicator of thick triangles and, hence, positive
curvature, as discussed in~\Cref{sec:geom-prop-graphs}.

\clearpage
% \makeatletter
% \setlength{\@fptop}{0pt}
% \makeatother
\section{Extended Graph Reconstruction Results}\label{sec:extended-results}

\paragraph{$\Spd$ vs.\ $\Hyp$.}
All graph reconstruction results comparing the SPD and hyperbolic manifolds are shown
in~\Cref{tab:all-noncomp-recon-results}.  Compared to~\Cref{tab:noncomp-recon-results} from the main
text this table includes, on one hand, more graphs and, at the same time, the performance results on
10-dimensional manifolds for most datasets. Notice, in particular, that the results for two much
larger graphs are included~\cite{snapnets}: ``cit-dblp'' and ``condmat''. They are both citation
networks with about 12000 and 23000 nodes, respectively. The columns are the same as
in~\Cref{tab:noncomp-recon-results}.

\paragraph{$\Gr$ vs.\ $\Sph$.}
All graph reconstruction results comparing the Grassmann and spherical manifolds are shown
in~\Cref{tab:all-compact-recon-results}. It includes two 3D models from the Stanford 3D Scanning
Repository~\cite{levoy2005stanford}: the notorious ``Stanford bunny'' and a ``drill shaft'' (the
mesh of a drill bit).

\paragraph{Cartesian Products.}
Graph reconstruction results comparing several product manifolds are shown
in~\Cref{tab:cartesian-recon-results}. The three datasets were chosen as some of the most
challenging based on the previous results. The SPD embeddings are trained using the Stein
divergence. All six product manifolds have 12 free parameters.

\begin{figure}
  \captionof{table}{All graph reconstruction results for ``$\Spd$ vs.\ $\Hyp$''. See text for more details.}
  \label{tab:all-noncomp-recon-results}
  \vspace*{0.1in}

  \begin{minipage}[t]{0.5\linewidth}
  \small
  \centering
  \renewcommand{\arraystretch}{1.0}

  \begin{tabularx}{0.85\textwidth}[t]{@{}cclccc@{}}
    \toprule
    $G$ & $n$ & $\mathcal{M}$ & \textbf{F1@1} & \textbf{AUC} & \textbf{AD} \\

    \midrule
    \multirow{8}{*}{\normalsize\bfseries \rotatebox[origin=c]{90}{bio-diseasome}} &
    \multirow{4}{*}{3}
    & Euc & 83.78 & 91.21 & 0.145\\
    && Hyp & \cellcolor{col1}86.21 & \cellcolor{col1}95.72 & 0.137\\
    && SPD & 83.99 & 91.32 & 0.140\\
    && Stein & \cellcolor{col1}86.70 & \cellcolor{col2}94.54 & \cellcolor{col1}0.105\\
    \cmidrule{2-6}
    & \multirow{4}{*}{6}
    & Euc & 93.48 & 95.84 & 0.073\\
    && Hyp & \cellcolor{col1}96.50 & \cellcolor{col1}98.42 & 0.071\\
    && SPD & 93.83 & 95.93 & 0.072\\
    && Stein & \cellcolor{col2}94.86 & \cellcolor{col2}97.64 & \cellcolor{col1}0.066\\

    \midrule
    \multirow{8}{*}{\normalsize\bfseries \rotatebox[origin=c]{90}{bio-wormnet}} &
    \multirow{4}{*}{3}
    & Euc & 89.36 & 93.84 & 0.157 \\
    && Hyp & \cellcolor{col1}91.26 & \cellcolor{col1}97.01 & 0.157 \\
    && SPD & 88.91 & 94.36 & 0.159 \\
    && Stein & \cellcolor{col1}90.92 & \cellcolor{col2}95.80 & \cellcolor{col1}0.120 \\
    \cmidrule{2-6}
    & \multirow{4}{*}{6}
    & Euc & 98.14 & 97.89 & 0.090 \\
    && Hyp & 98.55 & \cellcolor{col1}99.00 & 0.089 \\
    && SPD & 98.12 & 97.90 & 0.090 \\
    && Stein & 98.29 & \cellcolor{col2}98.63 & \cellcolor{col1}0.085 \\

    \midrule
    \multirow{12}{*}{\normalsize\bfseries \rotatebox[origin=c]{90}{california}} &
    \multirow{4}{*}{3}
    & Euc & 15.97 & 77.99 & 0.2297\\
    && Hyp & \cellcolor{col1}29.72 & \cellcolor{col1}85.78 & \cellcolor{col1}0.109\\
    && SPD & 15.61 & 82.82 & 0.230\\
    && Stein & \cellcolor{col2}24.66 & \cellcolor{col2}84.04 & \cellcolor{col1}0.118\\
    \cmidrule{2-6}
    & \multirow{4}{*}{6}
    & Euc & 29.29 & 84.59 & 0.143\\
    && Hyp & \cellcolor{col1}43.04 & \cellcolor{col1}88.64 & \cellcolor{col1}0.098\\
    && SPD & 29.20 & 86.30 & 0.122\\
    && Stein & \cellcolor{col2}34.85 & \cellcolor{col2}87.95 & \cellcolor{col1}0.101\\
    \cmidrule{2-6}
    & \multirow{4}{*}{10}
    & Euc & 41.39 & 87.77 & 0.105\\
    && Hyp & \cellcolor{col1}51.11 & \cellcolor{col2}90.68 & 0.094\\
    && SPD & 41.81 & \cellcolor{col2}90.10 & 0.098\\
    && Stein & \cellcolor{col2}46.03 & \cellcolor{col2}90.11 & 0.091\\

    \midrule
    \multirow{6}{*}{\normalsize\bfseries \rotatebox[origin=c]{90}{condmat}} &
    \multirow{4}{*}{3}
    & Euc & 16.01 & 74.62 & 0.303\\
    && Hyp & \cellcolor{col1}21.67 & \cellcolor{col1}84.14 & \cellcolor{col1}0.186\\
    && SPD & 16.63 & \cellcolor{col2}80.66 & 0.303\\
    \cmidrule{2-6}
    & \multirow{4}{*}{6}
    & Euc & 31.53 & 81.01 & 0.242\\
    && Hyp & \cellcolor{col1}47.53 & \cellcolor{col1}87.58 & \cellcolor{col1}0.180\\
    && SPD & 32.31 & \cellcolor{col2}83.04 & \cellcolor{col2}0.200\\

    \midrule
    \multirow{12}{*}{\normalsize\bfseries \rotatebox[origin=c]{90}{power}} &
    \multirow{4}{*}{3}
    & Euc & 49.34 & 87.84 & 0.119 \\
    && Hyp & \cellcolor{col1}60.18 & \cellcolor{col1}91.28 & \cellcolor{col1}0.068 \\
    && SPD & 52.48 & \cellcolor{col2}90.17 & 0.121 \\
    && Stein & \cellcolor{col2}54.06 & \cellcolor{col2}90.16 & \cellcolor{col2}0.076 \\
    \cmidrule{2-6}
    & \multirow{4}{*}{6}
    & Euc & 63.62 & 92.09 & 0.061 \\
    && Hyp & \cellcolor{col1}75.02 & \cellcolor{col1}94.34 & 0.060 \\
    && SPD & 67.69 & 91.76 & 0.062 \\
    && Stein & \cellcolor{col2}70.70 & \cellcolor{col2}93.32 & \cellcolor{col1}0.049 \\
    \cmidrule{2-6}
    & \multirow{4}{*}{10}
    & Euc & 74.14 & 94.35 & 0.042\\
    && Hyp & \cellcolor{col1}84.77 & \cellcolor{col1}96.25 & 0.038\\
    && SPD & \cellcolor{col2}79.36 & \cellcolor{col2}95.59 & \cellcolor{col2}0.033\\
    && Stein & 78.13 & 94.67 & 0.040\\

    \midrule
    \multirow{8}{*}{\normalsize\bfseries \rotatebox[origin=c]{90}{road-minnesota}} &
    \multirow{4}{*}{3}
    & Euc & \cellcolor{col2}90.35 & 95.94 & 0.058\\
    && Hyp & 90.02 & 95.82 & 0.058\\
    && SPD & 89.71 & 95.94 & 0.058\\
    && Stein & \cellcolor{col1}91.96 & 96.02 & 0.060\\
    \cmidrule{2-6}
    & \multirow{4}{*}{6}
    & Euc & 96.68 & 97.21 & 0.045\\
    && Hyp & 96.59 & 97.17 & 0.046\\
    && SPD & 96.81 & 97.20 & 0.045\\
    && Stein & 96.21 & 97.20 & 0.046\\

    \bottomrule
  \end{tabularx}
  \end{minipage}\hfill
  \begin{minipage}[t]{0.5\linewidth}
  \small
  \centering
  \renewcommand{\arraystretch}{1.0}

  \begin{tabularx}{0.85\textwidth}[t]{@{}cclccc@{}}
    \toprule
    $G$ & $n$ & $\mathcal{M}$ & \textbf{F1@1} & \textbf{AUC} & \textbf{AD} \\

    \midrule
    \multirow{6}{*}{\normalsize\bfseries \rotatebox[origin=c]{90}{cit-dblp}} &
    \multirow{4}{*}{3}
    & Euc & 11.43 & 79.63 & 0.311\\
    && Hyp & \cellcolor{col1}20.05 & \cellcolor{col1}86.97 & \cellcolor{col1}0.199\\
    && SPD & 11.54 & \cellcolor{col2}83.10 & 0.311\\
    \cmidrule{2-6}
    & \multirow{4}{*}{6}
    & Euc & 24.38 & 85.32 & 0.253\\
    && Hyp & \cellcolor{col1}34.79 & \cellcolor{col1}89.80 & \cellcolor{col1}0.205\\
    && SPD & 24.80 & \cellcolor{col2}87.27 & 0.253\\

    \midrule
    \multirow{12}{*}{\normalsize\bfseries \rotatebox[origin=c]{90}{csphd}} &
    \multirow{4}{*}{3}
    & Euc & 52.12 & 89.36 & 0.123\\
    && Hyp & \cellcolor{col1}55.55 & \cellcolor{col1}92.46 & 0.124\\
    && SPD & 52.34 & 89.71 & 0.127\\
    && Stein & \cellcolor{col2}54.45 & \cellcolor{col2}91.61 & \cellcolor{col1}0.098\\
    \cmidrule{2-6}
    & \multirow{4}{*}{6}
    & Euc & 60.63 & 93.91 & 0.065\\
    && Hyp & \cellcolor{col1}64.89 & \cellcolor{col2}94.73 & 0.065\\
    && SPD & 60.59 & \cellcolor{col2}94.12 & 0.066\\
    && Stein & \cellcolor{col2}62.73 & \cellcolor{col2}94.95 & \cellcolor{col2}0.062\\
    \cmidrule{2-6}
    & \multirow{4}{*}{10}
    & Euc & 66.66 & 96.06 & 0.050 \\
    && Hyp & \cellcolor{col1}74.76 & 96.38 & 0.045 \\
    && SPD & 67.46 & 96.34 & 0.048 \\
    && Stein & \cellcolor{col2}70.53 & 96.22 & 0.050 \\

    \midrule
    \multirow{12}{*}{\normalsize\bfseries \rotatebox[origin=c]{90}{facebook}} &
    \multirow{4}{*}{3}
    & Euc & 70.28 & 95.27 & 0.193\\
    && Hyp & \cellcolor{col2}71.08 & 95.46 & 0.173\\
    && SPD & \cellcolor{col2}71.09 & 95.26 & 0.170\\
    && Stein & \cellcolor{col1}75.91 & 95.59 & \cellcolor{col1}0.114 \\
    \cmidrule{2-6}
    & \multirow{4}{*}{6}
    & Euc & 79.60 & 96.41 & 0.090\\
    && Hyp & 81.83 & 96.53 & 0.089\\
    && SPD & 79.52 & 96.37 & 0.090\\
    && Stein & \cellcolor{col1}83.95 & 96.74 & \cellcolor{col1}0.061 \\
    \cmidrule{2-6}
    & \multirow{4}{*}{10}
    & Euc & 85.03 & 97.99 & 0.054\\
    && Hyp & \cellcolor{col2}86.93 & 97.28 & 0.053\\
    && SPD & 85.25 & 97.98 & 0.049\\
    && Stein & \cellcolor{col1}89.25 & 97.82 & \cellcolor{col2}0.044\\

    \midrule
    \multirow{12}{*}{\normalsize\bfseries \rotatebox[origin=c]{90}{grqc}} &
    \multirow{4}{*}{3}
    & Euc & 49.61 & 79.99 & 0.212\\
    && Hyp & \cellcolor{col1}66.54 & \cellcolor{col1}87.34 & \cellcolor{col1}0.108\\
    && SPD & 50.41 & 80.48 & 0.208\\
    && Stein & \cellcolor{col2}57.26 & \cellcolor{col2}85.20 & \cellcolor{col1}0.115\\
    \cmidrule{2-6}
    & \multirow{4}{*}{6}
    & Euc & 71.71 & 86.89 & 0.125\\
    && Hyp & \cellcolor{col1}82.43 & \cellcolor{col1}91.53 & \cellcolor{col1}0.091\\
    && SPD & 72.06 & 88.60 & 0.125\\
    && Stein & \cellcolor{col2}78.00 & \cellcolor{col2}90.20 & \cellcolor{col1}0.094\\
    \cmidrule{2-6}
    & \multirow{4}{*}{10}
    & Euc & 83.97 & 91.28 & 0.090\\
    && Hyp & \cellcolor{col1}89.33 & \cellcolor{col1}94.19 & \cellcolor{col2}0.077\\
    && SPD & 84.74 & \cellcolor{col2}93.48 & \cellcolor{col2}0.081\\
    && Stein & \cellcolor{col2}87.62 & \cellcolor{col2}93.19 & \cellcolor{col2}0.081\\

    \midrule
    \multirow{12}{*}{\normalsize\bfseries \rotatebox[origin=c]{90}{web-edu}} &
    \multirow{4}{*}{3}
    & Euc & 29.18 & 87.14 & 0.245\\
    && Hyp & \cellcolor{col1}55.60 & \cellcolor{col1}92.10 & 0.245\\
    && SPD & 29.02 & 88.54 & 0.246\\
    && Stein & \cellcolor{col2}48.28 & \cellcolor{col2}90.87 & \cellcolor{col1}0.084\\
    \cmidrule{2-6}
    & \multirow{4}{*}{6}
    & Euc & 49.31 & 91.19 & 0.143\\
    && Hyp & \cellcolor{col1}66.23 & \cellcolor{col2}95.78 & 0.143\\
    && SPD & 42.16 & 91.90 & 0.142\\
    && Stein & \cellcolor{col2}62.81 & \cellcolor{col1}96.51 & \cellcolor{col1}0.043\\
    \cmidrule{2-6}
    & \multirow{4}{*}{10}
    & Euc & 42.47 & 93.31 & 0.082\\
    && Hyp & \cellcolor{col1}98.43 & \cellcolor{col1}98.18 & 0.073\\
    && SPD & 88.30 & \cellcolor{col2}96.86 & \cellcolor{col2}0.045\\
    && Stein & \cellcolor{col2}91.02 & \cellcolor{col1}98.24 & \cellcolor{col1}0.037\\

    \bottomrule
  \end{tabularx}
\end{minipage}
\end{figure}

\begin{figure}
  \begin{minipage}[t]{0.45\linewidth}
  \captionof{table}{All graph reconstruction results for ``$\Gr$ vs.\ $\Sph$''. See text for more details.}
  \label{tab:all-compact-recon-results}
  \vspace*{0.1in}

  \small
  \renewcommand{\arraystretch}{1.0}

  \begin{tabularx}{\textwidth}[t]{@{}cclcccc@{}}
    \toprule
    $G$ & $n$ & $\mathcal{M}$ & \textbf{F1@1} & \textbf{AUC} & \textbf{AD} \\

    \midrule
    \multirow{8}{*}{\normalsize\bfseries \rotatebox[origin=c]{90}{sphere-mesh}} &
    \multirow{2}{*}{2}
    & Sphere & \cellcolor{col1}99.99 & \cellcolor{col1}98.31 & \cellcolor{col1}0.051\\
    && $\Gr(1,3)$ & 97.20 & 90.66 & 0.148\\
    \cmidrule{2-6}
    & \multirow{2}{*}{3}
    & Sphere & 100.00 & 98.76 & \cellcolor{col1}0.042\\
    && $\Gr(1,4)$ & 100.00 & 98.38 & 0.060\\
    \cmidrule{2-6}
    & \multirow{3}{*}{4}
    & Sphere & 100.00 & 98.84 & \cellcolor{col1}0.041\\
    && $\Gr(1,5)$ & 100.00 & 98.69 & 0.060\\
    && $\Gr(2,4)$ & 100.00 & 98.94 & \cellcolor{col1}0.040\\

    \midrule
    \multirow{8}{*}{\normalsize\bfseries \rotatebox[origin=c]{90}{bunny}} &
    \multirow{2}{*}{2}
    & Sphere & \cellcolor{col1}88.12 & \cellcolor{col1}89.61 & 0.146\\
    && $\Gr(1,3)$ & 85.06 & 85.95 & 0.146\\
    \cmidrule{2-6}
    & \multirow{2}{*}{3}
    & Sphere & 94.96 & 96.86 & 0.062\\
    && $\Gr(1,4)$ & \cellcolor{col2}95.41 & \cellcolor{col2}97.22 & 0.062\\
    \cmidrule{2-6}
    & \multirow{3}{*}{4}
    & Sphere & 95.91 & 97.53 & 0.055\\
    && $\Gr(1,5)$ & 96.03 & 97.63 & 0.057\\
    && $\Gr(2,4)$ & 95.86 & 97.62 & 0.058\\

    \midrule
    \multirow{8}{*}{\normalsize\bfseries \rotatebox[origin=c]{90}{drill-shaft}} &
    \multirow{2}{*}{2}
    & Sphere & \cellcolor{col1}84.67 & 96.27 & 0.073\\
    && $\Gr(1,3)$ & 83.40 & 96.34 & 0.074\\
    \cmidrule{2-6}
    & \multirow{2}{*}{3}
    & Sphere & \cellcolor{col2}89.14 & 97.85 & 0.052\\
    && $\Gr(1,4)$ & 88.88 & 97.81 & 0.052\\
    \cmidrule{2-6}
    & \multirow{3}{*}{4}
    & Sphere & 92.45 & 98.51 & 0.043\\
    && $\Gr(1,5)$ & 92.44 & 98.51 & 0.043\\
    && $\Gr(2,4)$ & 92.77 & 98.54 & 0.043\\

    \midrule
    \multirow{8}{*}{\normalsize\bfseries \rotatebox[origin=c]{90}{road-minnesota}} &
    \multirow{2}{*}{2}
    & Sphere & \cellcolor{col1}82.19 & 94.02 & 0.085\\
    && $\Gr(1,3)$ & 78.91 & 94.02 & 0.085\\
    \cmidrule{2-6}
    & \multirow{2}{*}{3}
    & Sphere & 89.55 & 95.89 & 0.059\\
    && $\Gr(1,4)$ & \cellcolor{col2}90.02 & 95.88 & 0.058\\
    \cmidrule{2-6}
    & \multirow{3}{*}{4}
    & Sphere & 93.65 & 96.66 & 0.049\\
    && $\Gr(1,5)$ & 93.89 & 96.67 & 0.049\\
    && $\Gr(2,4)$ & \cellcolor{col2}94.01 & 96.66 & 0.049\\

    \midrule
    \multirow{8}{*}{\normalsize\bfseries \rotatebox[origin=c]{90}{cat-cortex}} &
    \multirow{2}{*}{2}
    & Sphere & {-} & {-} & 0.255\\
    && $\Gr(1,3)$ & {-} & {-} & \cellcolor{col1}0.234\\
    \cmidrule{2-6}
    & \multirow{2}{*}{3}
    & Sphere & {-} & {-} & 0.195\\
    && $\Gr(1,4)$ & {-} & {-} & \cellcolor{col1}0.168\\
    \cmidrule{2-6}
    & \multirow{3}{*}{4}
    & Sphere & {-} & {-} & 0.156\\
    && $\Gr(1,5)$ & {-} & {-} & \cellcolor{col2}0.139\\
    && $\Gr(2,4)$ & {-} & {-} & \cellcolor{col1}0.129\\

    \bottomrule
  \end{tabularx}
  \end{minipage}\hfill
  \begin{minipage}[t]{0.485\linewidth}
  \captionof{table}{Graph reconstruction results for six Cartesian products of Riemannian manifolds. See text for more details.}
  \label{tab:cartesian-recon-results}
  \vspace*{0.1in}

  \small
  \renewcommand{\arraystretch}{1.0}

  \begin{tabularx}{\textwidth}[t]{@{}clcccc@{}}
    \toprule
    $G$ & $\mathcal{M}$ & \textbf{F1@1} & \textbf{AUC} & \textbf{AD} \\

    \midrule
    \multirow{6}{*}{\normalsize\bfseries \rotatebox[origin=c]{90}{california}} &
    $\Hyp(3)^4$ & \cellcolor{col2}55.15 & \cellcolor{col2}91.35 & 0.096\\
    & $\Hyp(6)^2$ & \cellcolor{col1}56.93 & \cellcolor{col2}91.55 & 0.096\\
    & $\Hyp(6) \times \Sph(6)$ & \cellcolor{col2}55.12 & \cellcolor{col2}91.15 & 0.096\\
    & $\Spd(2)^4$ & 49.49 & 90.77 & \cellcolor{col2}0.087\\
    & $\Spd(3)^2$ & 50.78 & 90.82 & \cellcolor{col2}0.086\\
    & $\Spd(3) \times \Gr(3,5)$ & 47.21 & 90.53 & \cellcolor{col2}0.089\\

    \midrule
    \multirow{6}{*}{\normalsize\bfseries \rotatebox[origin=c]{90}{grqc}} &
    $\Hyp(3)^4$ & \cellcolor{col1}92.03 & \cellcolor{col2}95.06 & 0.081\\
    & $\Hyp(6)^2$ & \cellcolor{col2}91.14 & \cellcolor{col2}94.97 & 0.080\\
    & $\Hyp(6) \times \Sph(6)$ & \cellcolor{col2}91.82 & 94.44 & 0.081\\
    & $\Spd(2)^4$ & 89.39 & 94.25 & 0.080\\
    & $\Spd(3)^2$ & 89.29 & 94.20 & \cellcolor{col2}0.076\\
    & $\Spd(3) \times \Gr(3,5)$ & 89.54 & 93.79 & 0.081\\

    \midrule
    \multirow{6}{*}{\normalsize\bfseries \rotatebox[origin=c]{90}{web-edu}} &
    $\Hyp(3)^4$ & \cellcolor{col1}99.26 & \cellcolor{col2}98.54 & 0.070\\
    & $\Hyp(6)^2$ & \cellcolor{col1}98.62 & \cellcolor{col1}99.24 & 0.071\\
    & $\Hyp(6) \times \Sph(6)$ & \cellcolor{col1}99.14 & \cellcolor{col2}98.39 & 0.071\\
    & $\Spd(2)^4$ & 71.90 & 96.78 & \cellcolor{col1}0.027\\
    & $\Spd(3)^2$ & \cellcolor{col2}78.47 & 97.12 & \cellcolor{col1}0.027\\
    & $\Spd(3) \times \Gr(3,5)$ & 69.76 & 96.24 & 0.073\\

    \bottomrule
  \end{tabularx}
  \end{minipage}
\end{figure}

\end{document}